\documentclass[11pt]{article}

\date{}

\usepackage[utf8]{inputenc} 
\usepackage[T1]{fontenc}    
\usepackage{hyperref}       
\usepackage{url}            
\usepackage{booktabs}       
\usepackage{amsfonts}       
\usepackage{nicefrac}       
\usepackage{microtype}      
\usepackage{xspace}
\usepackage{natbib}

\usepackage{comment}
\usepackage{amsmath}
\usepackage{amsthm}
\usepackage{graphicx}
\usepackage{rotating}
\usepackage{float}
\usepackage{mathrsfs}
\usepackage{amssymb}
\usepackage{autobreak}
\usepackage{mathtools}
\usepackage{wrapfig}
\usepackage{xcolor}
\usepackage{bbm}
\usepackage[ruled,vlined]{algorithm2e}
\usepackage{algorithmic}

\usepackage{support_caption}
\usepackage{subcaption}
\usepackage{lipsum}

\usepackage{tikz}
\usetikzlibrary{shapes,backgrounds}

\DeclareMathOperator*{\argmax}{arg\,max}
\DeclareMathOperator*{\argmin}{arg\,min}
\newcommand\numberthis{\addtocounter{equation}{1}\tag{\theequation}}
\newcommand{\CMDP}{\textsc{\small{CoMDP}}\xspace}
\newcommand{\PG}{\textsc{\small{PG}}\xspace}
\newcommand{\RL}{\textsc{\small{RL}}\xspace}
\newcommand{\gdapg}{\textsc{\small{GDA-PG}}\xspace}
\newcommand{\gda}{\textsc{\small{GDA}}\xspace}
\newcommand{\cgd}{\textsc{\small{CGD}}\xspace}
\newcommand{\cpo}{\textsc{\small{CoPO}}\xspace}
\newcommand{\cpg}{\textsc{\small{CoPG}}\xspace}
\newcommand{\cpgsp}{\textsc{\small{CoPG-SP}}\xspace}

\newcommand{\trcpo}{\textsc{\small{TRCoPO}}\xspace}
\newcommand{\trpogda}{\textsc{\small{TRGDA}}\xspace}
\newcommand{\maddpg}{\textsc{\small{MADDPG}}\xspace}
\newcommand{\lola}{\textsc{\small{LOLA}}\xspace}

\newcommand{\stermination}{s_{T}}
\newcommand{\GAE}{GAE}

\newcommand*{\eg}{e.g.\@\xspace}
\newcommand*{\eq}{Eq.}
\newcommand*{\fig}{Fig.}
\newcommand*{\lem}{Lem.}
\newcommand*{\tab}{Tab.}
\newcommand*{\appx}{Apx.}
\newcommand*{\sect}{Sec.}
\newcommand*{\thm}{Thm.}
\newcommand*{\alg}{Alg.}
\newcommand*{\prop}{Prop.}

\newcommand{\MDP}{\textsc{\small{MDP}}\xspace}
\newcommand{\TRPO}{\textsc{\small{TRPO}}\xspace}

\newcommand{\A}{\mathcal A}

\newcommand{\T}{\mathcal T}

\renewcommand{\S}{\mathcal S}

\newcommand{\R}{\mathcal{R}}

\newcommand{\wh}{\widehat}

\newtheorem{lemma}{Lemma}

\newtheorem{theorem*}{Theorem}

\newtheorem{proposition}{Proposition}
\newtheorem{theorem}{Theorem}

\makeatletter
\def\thanks#1{\protected@xdef\@thanks{\@thanks
        \protect\footnotetext{#1}}}
\makeatother

\title{Competitive Policy Optimization}

\author{%
  Manish Prajapat$^{1}$, Kamyar Azizzadenesheli$^2$, Alexander Liniger$^1$,\\ Yisong Yue$^2$, Anima Anandkumar$^2$\\
  $^1$ ETH Z\"urich\\
  $^2$ California Institute of Technology
}
\begin{document}

\maketitle

\begin{abstract}
A core challenge in policy optimization in competitive Markov decision processes is the design of efficient optimization methods with desirable convergence and stability properties. To tackle this, we propose competitive policy optimization (\cpo), a novel policy gradient approach that exploits the game-theoretic nature of competitive games to derive policy updates. Motivated by the competitive gradient optimization method,
we derive a bilinear approximation of the game objective. In contrast, off-the-shelf policy gradient methods utilize only linear approximations, and hence do not capture interactions among the players. 
We instantiate \cpo in two ways: $(i)$ competitive policy gradient, and $(ii)$ trust-region competitive policy optimization. We theoretically study these methods, and empirically investigate their behavior on a set of comprehensive, yet challenging, competitive games. We observe that they provide stable optimization, convergence to sophisticated strategies, and higher scores when played against baseline policy gradient methods. 
\end{abstract}

\section{Introduction}
\label{sec:introduction}
	Reinforcement learning (\RL) in competitive Markov decision process {\CMDP}~\citep{filar2012competitive}
is the study of competitive players, sequentially making decisions in an environment~\citep{puterman2014markov,Sutton2018}. In \CMDP{}s, the competing agents (players) interact with each other within the environment, and through their interactions, learn how to develop their behavior and improve their notion of reward. In this paper, we 
considered the rich and fundamental class of zero-sum two-player games. Applications of such games abound from hypothesis testing~\citep{wald1945sequential}, 
generative adversarial networks~\citep{goodfellow2014generative}, 
Lagrangian optimization~\citep{dantzig1998linear}, online learning~\citep{cesa2006prediction}, algorithmic game theory~\citep{roughgarden2010algorithmic},  robust control~\citep{zhou1996robust}, to board games~\citep{tesauro1995temporal,silver2016mastering}.

In these applications, two players compete against each other, potentially in unknown environments. One goal is to learn policies/strategies for agents by making them play against each other.
%
Due to the competitive nature of games, \RL methods that do not account for the interactions are often susceptible to poor performance and divergence even in simple scenarios. One of the core challenges in such settings is to design optimization paradigms with desirable convergence and stability properties.


Policy gradient (\PG) is a prominent \RL approach that directly optimizes for policies, and enjoys simplicity in implementation and deployment~\citep{robbins1951stochastic,Aleksandrov68,sutton2000policy}. 
\PG updates are derived by optimizing the first order (linear) approximation of the objective function while regularizing the parameter deviations. 
In two-player zero-sum \CMDP{}s, policy updates of the maximizing (minimizing) agent are derived through maximizing (resp. minimizing) the linear approximation of the game objective with an additive (subtractive) regularization, resulting in the gradient descent ascent (\gda) \PG algorithm. 
%
This approximation is linear in agents' parameters and does not take their interaction into account. Therefore, \gda directly optimizes the policy of each agent by assuming the policy/strategy of the opponent is not updated, which can be a poor approach for competitive optimization.

We propose a new paradigm known as competitive policy optimization (\cpo), which exploits the game-theoretic and competitive nature of \CMDP{}s, and, instead, inspired by competitive gradient descent~\citep{schfer2019competitive}, deploys a bilinear approximation of the game objective. The bilinear approximation is separately linear in each agent's parameters and is built on agents' interactions to derive the policy updates. To compute the policy updates, \cpo computes the Nash equilibrium of the bilinear approximation of the game objective with a (e.g., $\ell_2$) regularization of the parameter deviations. In \cpo, each agent derives its update with the full consideration of what the other agent's current move and moves in the future time steps might be. Importantly, each agent considers how the environment, as the results of the agents' current and future moves temporally, evolves in favor of each agent. Therefore, each agent hypothesizes about what the other agent's and the environment's responses would be, consequently resulting in the common \textit{recursive reasoning} in game theory, especially temporal recursion in \CMDP{}s. We further show that \cpo is a generic approach that does not rely on the structure of the problem, at the cost of high sample complexity. 

We instantiate \cpo in two ways to arrive at practical algorithms. We propose competitive policy gradient (\cpg), a novel \PG algorithm that exploit value functions and structure of \CMDP{}s to efficiently obtain policy updates. We also propose trust region competitive policy optimization (\trcpo), a novel trust region based \PG method~\citep{kakade2002approximately,schulman2015trust}. \trcpo optimizes a surrogate game objective in a trust region. \trcpo updates agents' parameters simultaneously by deriving the Nash equilibrium of bilinear (in contrast to linear approximation in off-the-shelf trust region methods) approximation to the surrogate objective within a defined trust region in the parameter space.



We empirically study the performance of \cpg and \trcpo on six representative games, ranging from single-state to general sequential games, and tabular to infinite/continuous states/action games. We observe that \cpg and \trcpo not only provide more stable and faster optimization, but also converge to more sophisticated, opponent aware, and competitive strategies compared with their conventional counterparts, \gda and \trpogda. When trained agents play against each other, we observe the superiority of \cpg and \trcpo agents, e.g., in the Markov soccer game, \cpg agent beats \gda in 74\%, and \trcpo agent beats \trpogda in 85\% of the matches.
We observe a similar trend with other methods, such as multi-agent deep deterministic \PG(\maddpg)~\citep{lowe2017multi}. 


A related approach that uses a second-level reasoning (as opposed to \cpo with infinite-level reasoning) is \lola~\citep{FoersterLOLA}, which shows positive results on three games, matching pennies, prisoners' dilemma and a coin game. We compare \lola against \cpg and \trcpo and observe a similar trend in the superiority of \cpo based methods. We conclude the paper by arguing the generality of \cpo paradigm, and how it can be used to generalize single-agent \PG methods to \CMDP{}s.

\section{Preliminaries}
\label{sec:priliminaries}
    A two player \CMDP is a tuple of $\langle\S,\A^1,\A^2,\R,\T,\mathcal{P},\gamma\rangle$, where $\S$ is the state space, $s \in \S$ is a state, for player $i\in\lbrace 1,2\rbrace$, $\A^i$ is the player $i$'s action space with $a^i\in\A^i$. $\R$ is the reward kernel with probability distribution $R(\cdot|s,a^1,a^2)$ and mean function $r(s,a^1,a^2)$ on $\mathbb{R}$. 
For a probability measure $\mathcal{P}$, $p$ denotes the probability distribution of initial state, and for the transition kernel $\T$, $T(s'|s,a^1,a^2)$ is the distribution of successive state $s'$ after taking actions  $a^1,a^2$ simultaneously at state $s$, with discount factor $\gamma \in [0,1]$. We consider episodic environments with reachable absorbing state $\stermination$ almost surely in finite time. 
An episode starts at $s_0\sim p$, and at each time step $k\geq 0$ at state $s_k$, each player $i$ draws its action $a_{k}^i$ according to policy $\pi(a_{k}^i|s_k;\theta^i)$ parameterized by $\theta^i\in \Theta^i$, where $\Theta^i\subset\mathbb{R}^{l}$ is a compact metric space. Players $1,2$ receive $(r_{k},-r_{k})$ with $r_{k}\sim R(s_k,a^1_k,a^2_k)$, and the environment evolves to a new state $s_{k+1}$. A realization of this stochastic process is a trajectory $\tau = \big( (s_k,a^1_{k},a^2_{k},r_{k})_{k=0}^{|\tau|-1} ,s_{|\tau|}\big)$, an ordered sequence 
with random length $|\tau|$, where $|\tau|$ is determined by episode termination time and state $s_{|\tau|}=\stermination$. Let $f(\tau;\theta^1,\theta^2)$ denote the probability distribution of the trajectory $\tau$ following players' policies $\pi(\theta^i)$,
\vspace{-1.00mm}
\begin{align} \label{eq: trajectory_distribution}
f(\tau; \theta^1, \theta^2)= p(s_0) \prod_{k=0}^{|\tau|-1} \pi(a^1_{k}|s_k;\theta^1) \pi(a^2_{k}|s_k;\theta^2)R(r_k|s_k,a^1_k,a^2_k) T(s_{k+1}|s_k,a^1_{k},a^2_{k}). \numberthis
\vspace{-3.00mm}
\end{align}
For $R(\tau) = \sum_{k=0}^{|\tau|}\gamma^kr(s_k,a^1_k,a^2_k)$, the $Q$-function, $V$-functions, and game objective are defined,
\begin{align}\label{eqn: val_rew}
    &\quad\quad\quad\quad Q(s_k,a^1_{k},a^2_{k};\theta^1, \theta^2)= \mathbb{E}_{\tau\sim f(\cdot; \theta^1, \theta^2)} \Big[ \sum\nolimits_{j=k}^{|\tau|-1}\! \gamma^{j-k} r(s_j,a^1_{j},a^2_{j})| s_k, a^1_k, a^2_k \Big],\\
    &\!\!V(s_k;\theta^1, \theta^2) = \mathbb{E}_{\tau\sim f(\cdot; \theta^1, \theta^2)} \Big[ \sum\nolimits_{j=k}^{|\tau|-1} \!\gamma^{j-k} r(s_j,a^1_{j},a^2_{j})| s_k\, \Big],
    ~\eta (\theta^1,\theta^2)=\!\! \int_{\tau} \!f(\tau;\theta^1,\theta^2) R(\tau) d\tau\nonumber 
    \vspace{-3.00mm}
\end{align}
We assume $V$, $Q$, and $\eta$ are differentiable and bounded in $(\Theta^1,\Theta^2)$ and for $f$ on $(\Theta^1$ $\Theta^2)$, $D_{\theta^i}f= \frac{\partial}{\partial {\theta'}^i}f({\theta'}^1\!\!,\!{\theta'}^2)\big|_{({\theta'}^1,{\theta'}^2)=({\theta}^1,{\theta}^2)}$, and $D_{\theta^i\theta^j} f = \frac{\partial}{\partial {\theta'}^i} \big( \frac{\partial }{\partial {\theta'}^j}f({\theta'}^1\!\!,\!{\theta'}^2) \big)\big|_{({\theta'}^1,{\theta'}^2)=({\theta}^1\!,{\theta}^2)}$, for $i,j\in\lbrace 1,2\rbrace$.


\section{Competitive Policy Optimization}
\label{sec:cpo_theory}
	Player 1 aims to  maximize the game objective $\eta$ \eq~\eqref{eqn: val_rew}, while player 2 aims to minimize it, i.e., simultaneously solving for $\max_{\theta^1} \eta (\theta^1,\theta^2) $ and $\min_{\theta^2} \eta (\theta^1,\theta^2)$ respectively with,
\begin{align}\label{eq: game_goal}
\theta^{1*} \in \argmax\nolimits_{\theta^1\in\Theta^1} \eta (\theta^1,\theta^2) ,~\text{and}~~ \theta^{2*} \in \argmin\nolimits_{\theta^2\in\Theta^2} \eta (\theta^1,\theta^2).
\end{align}
As discussed in the introduction, a straightforward generalization of \PG methods to \CMDP, results in \gda (\alg\ref{alg:gda_short}). Given players' parameters $(\theta^1,\theta^2)$, \gda prescribes to optimize a linear approximation of the game objective in the presence of a regularization for the policy updates,
\begin{align}\label{eq:GDAlinearised_game}
\!\!\!\!\theta^1\!\leftarrow\!\theta^1\!+\!\!\!\argmax_{{\Delta\theta}:{\Delta\theta}+\theta\in \Theta^1}\!\!\! {\Delta\theta}^{\top}\! D_{{\theta}^1} \eta - \frac{1}{2\alpha} ||{\Delta\theta}||^2\!,~\text{and}
\ \theta^2\leftarrow\theta^2\!\!+\!\!\!\argmin_{{\Delta\theta}:{\Delta\theta}+\theta\in \Theta^2}\!\!\! {\Delta\theta}^{\top} \!D_{{\theta}^2} \eta + \frac{1}{2\alpha} ||{\Delta\theta}||^2\!,\!\!
\end{align}
where $\alpha$ represent the step size. 
The parameter updates in \eq~\ref{eq:GDAlinearised_game} result in greedy updates along the directions of maximum change, assuming the other player stays constant. These updates are myopic, and ignore the agents' interactions. In other words, player 1 does not take player's 2 potential move into consideration and vice versa. While \gda might be an approach of interest in decentralized \CMDP{}, it mainly falls short in the problem of competitive and centralized optimization in unknown \CMDP{}s, i.e., the focus of this work. 
In fact, this behaviour is far from optimal and is shown to diverge in many simple cases \eg plain bilinear or linear quadratic games ~\citep{schfer2019competitive,mazumdar2019policygradient}.

We propose competitive policy optimization \cpo, a policy gradient approach for optimization in unknown \CMDP{}s. In contrast to standard \PG methods, such as \gda, \cpo considers a bilinear approximation of the game objective, and takes the interaction between players into account. \cpo incorporates the game theoretic nature of the \CMDP{} optimization and derives parameter updates through finding the Nash equilibrium of the bilinear approximation of 
the game objective,
\begin{align}\label{eq:CGDBilinearised_game}
&\theta^1\leftarrow\theta^1+\argmax\nolimits_{{\Delta\theta^1}:{\Delta\theta^1}+\theta^1\in \Theta^1} {\Delta\theta^1}^{\top} D_{{\theta}^1} \eta  + {\Delta\theta^1}^{\top} D_{\theta^1\theta^2} \eta {\Delta\theta^2} - \frac{1}{2\alpha} ||{\Delta\theta}^1||^2,\nonumber\\
&\theta^2\leftarrow\theta^2+\argmin\nolimits_{{\Delta\theta^2}:{\Delta\theta^2}+\theta^2\in \Theta^2} {\Delta\theta^2}^{\top} D_{{\theta}^2} \eta + {\Delta\theta^2}^{\top} D_{\theta^2\theta^1} \eta {\Delta\theta^1} + \frac{1}{2\alpha} ||{\Delta\theta}^2||^2,
\end{align}
%
which has an extra term, the interaction term, in contrast to \eq~\ref{eq:GDAlinearised_game}, and has the following closed-form solution,
%
\begin{align}\label{eqn: CGDSolution}
\theta^1 &\leftarrow \theta^1 + \alpha \bigl( I + \alpha^2 D_{\theta^1\theta^2} \eta D_{\theta^2\theta^1}\eta \bigr)^{-1} \bigl( D_{\theta^1} \eta - \alpha D_{\theta^1\theta^2}\eta D_{\theta^2}\eta \bigr) , \nonumber \\
\theta^2 &\leftarrow \theta^2 - \alpha \bigl( I + \alpha^2 D_{\theta^2\theta^1}\eta D_{\theta^1\theta^2}\eta \bigr)^{-1} \bigl( D_{\theta^2}\eta + \alpha D_{\theta^2\theta^1}\eta D_{\theta^1} \eta \bigr),
\end{align}
%
\hspace{1.0cm}
\begin{wrapfigure}{r}{0.45\textwidth}
\vspace*{-1.7 em}
\hspace{-2em}
    \begin{subfigure}[t]{0.30\columnwidth}
  	\includegraphics[width=\linewidth]{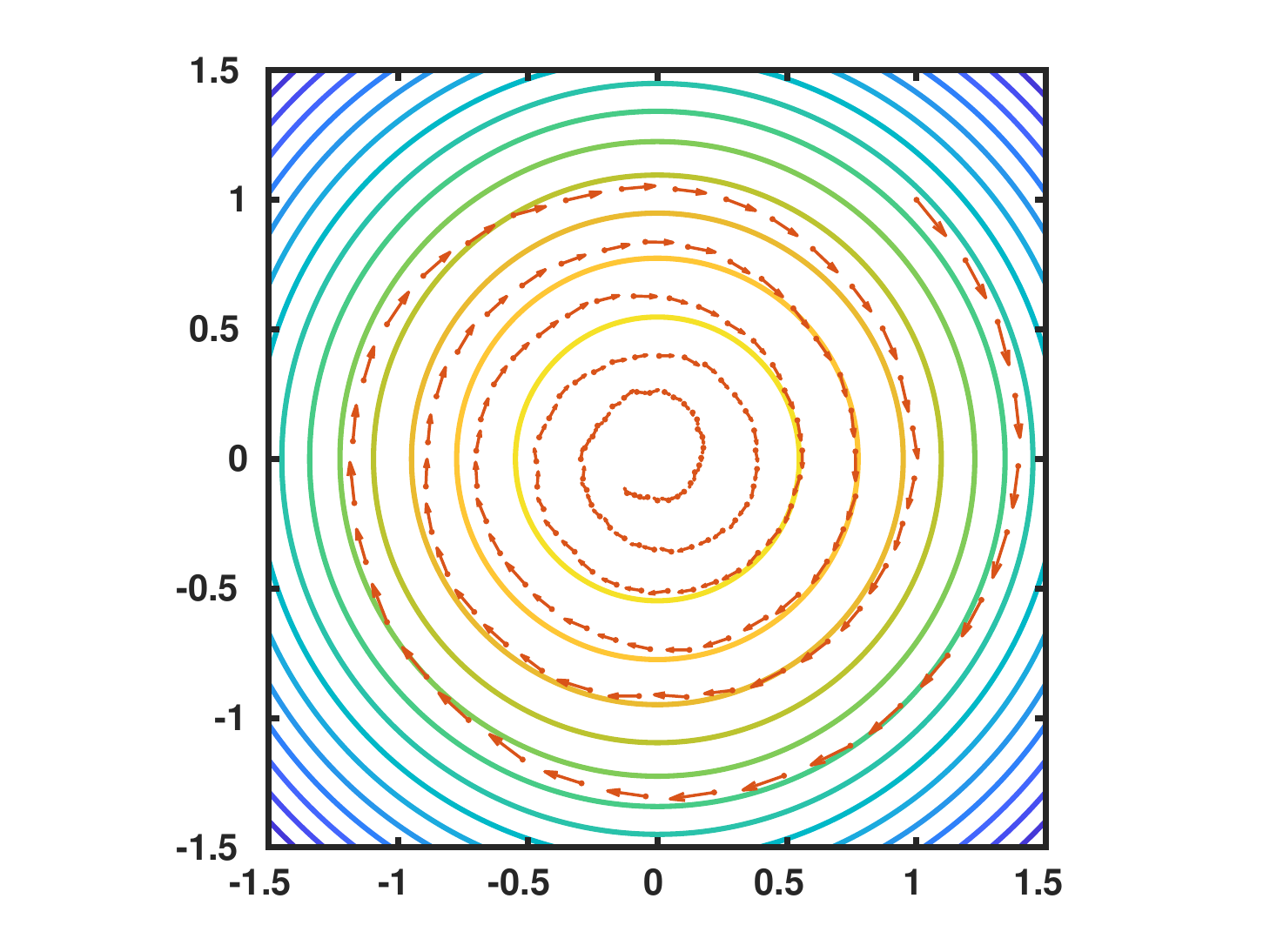}
\hspace*{4.em}(a) \cpo
    \end{subfigure}
    \hspace{-2cm}
  \begin{subfigure}[t]{0.30\columnwidth}
  \hspace*{3em}
  	\includegraphics[width=\linewidth]{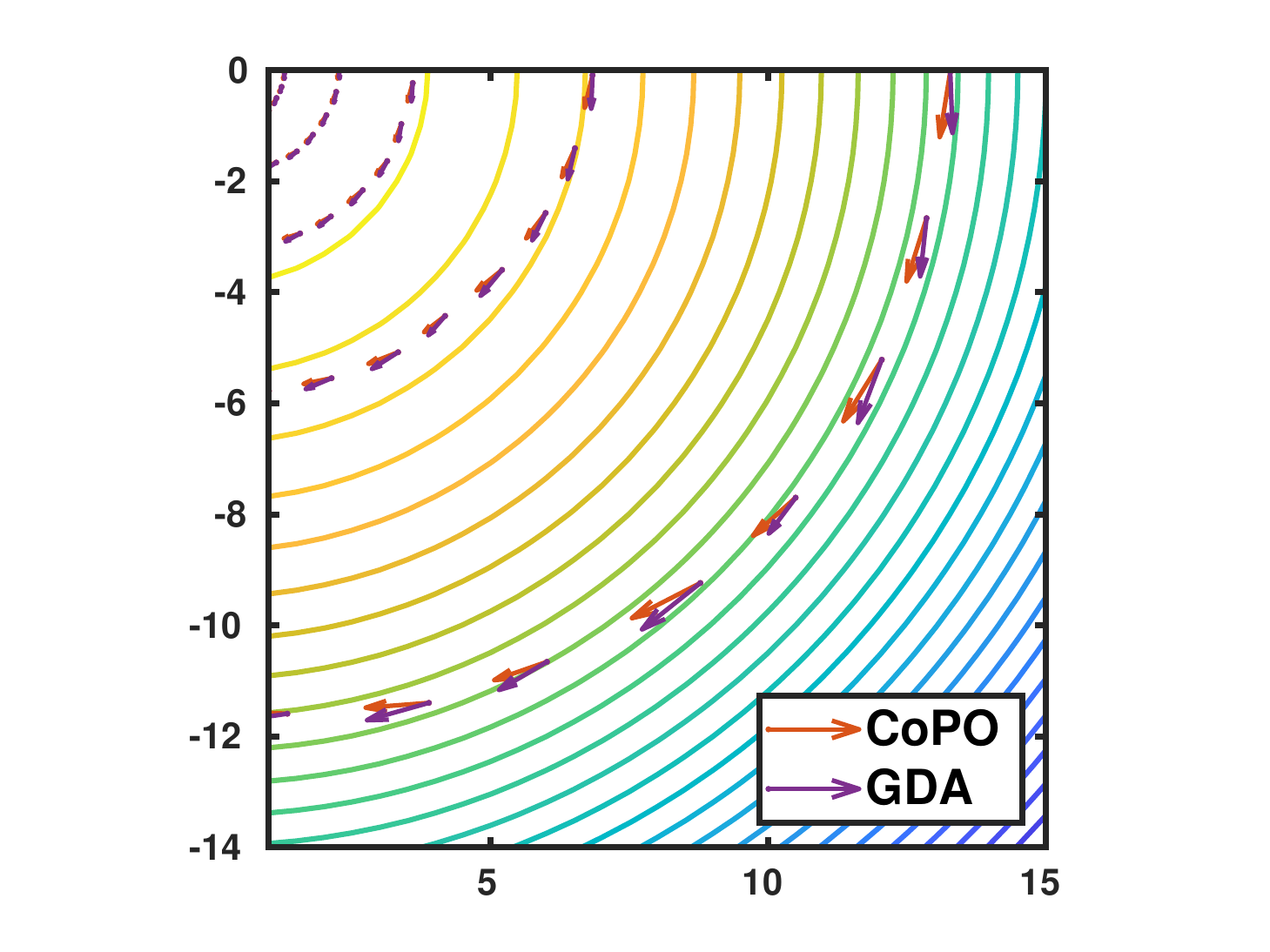}  
\hspace*{4.965em}(b) \cpo vs \gda\!\!\!\!\!\!\!\!\!\!\!\!\!\!\!\!\!\!\!\!\!\!\!\!\!\!\!\!\!\!\!\!\!\!\!\!
  \end{subfigure}
  \vspace*{-.5em}
  \caption{Bilinear games: (a) \cpo updates towards Nash equilibrium. (b) \gda updates point outward, leading to cycling/divergence.}
  \label{Fig: GradDirection}
  \vspace{-1em}
\end{wrapfigure}
%
where $I$ is an identity matrix of appropriate size. Note that, the bilinear approximation in \eq~\ref{eq:CGDBilinearised_game} is still linear in each player's action/parameters. Including any other terms, e.g., block diagonal Hessian terms from the Taylor expansion of the game objective, results in nonlinear terms in at least one player's parameters. %
As such, \cpo can be viewed as a natural linear generalization of \PG.
It is known that \gda-style co-learning approaches can often diverge or cycle indefinitely and never converge~\citep{mertikopoulos2018cycles}. \fig~\ref{Fig: GradDirection} shows that for a bilinear game, the gradient flow of \gda cycles or has gradient flow outward, while the \cpg flow, considering players' future moves, has gradient flow toward the Nash equilibrium and converges.
In \appx\ref{sec:Neuman}, we deploy the Neumann series expansion of the inverses in \eq\ref{eq:CGDBilinearised_game}, and show that \cpo recovers the infinite recursion reasoning between players and the environment, while \gda correspond to the first term, and LOLA corresponds to the first two terms in the series. Next, we compute terms in \eq~\ref{eqn: CGDSolution} using the score function $g(\tau, \theta^i):= D_{\theta^i}(\log \prod_{k=0}^{|\tau|-1} \pi(a_{k}|s_k;\theta^i))$,
\begin{proposition}\label{prop: CPOGradHessianTrajDist}
Given a \CMDP, players $i,j\in\!\lbrace 1,2\rbrace$, $i \neq j$ and the policy parameters $\theta^i, \theta^j$,
\begin{align*}
D_{{\theta}^i}\eta = \int_{\tau} f(\tau;\theta^1,\theta^2) g(\tau, \theta^i) R(\tau) d\tau,\textit{and}~D_{\theta^{i}\theta^{j}} \eta\! = \int_{\tau} f(\tau;\theta^1,\theta^2)  g(\tau, \theta^{i}) g(\tau, \theta^{j})^{\top} R(\tau) d\tau.
\end{align*}
\end{proposition}
Proof in \appx~\ref{appx: CPOGradHessianTrajDist}. In practice, we use conjugate gradient and Hessian vector product to efficiently compute the updates in \eq\ref{eqn: CGDSolution}, as explained in later sections. A closer look at \cpo shows that this paradigm does not require the knowledge of the environment if sampled trajectories are available. It neither requires full observability of the sates, nor any structural assumption on \CMDP, but the Monte Carlo estimation suffer from high variance. In the following, we explicitly take the \CMDP structure into account to develop efficient algorithms.

\subsection{Competitive Policy Gradient}
\label{sec:cpg_theory}

We propose competitive policy gradient (\cpg), an efficient algorithm that exploits the structure of \CMDP{}s to compute the parameter updates. The following is the \CMDP generalizing of the single agent \PG theorem ~\citep{sutton2000policy}. For $\tau_{l:l'}= (s_k,a^1_{k},a^2_{k},r_{k})_{k=l}^{l'}$, the events from time step $l$ to $l'$, we have:

\begin{theorem} \label{thm: CPGGradHessianVal}
Given a \CMDP, players $i,j\in\!\lbrace 1,2\rbrace$, $i \neq j$, and the policy parameters $\theta^i,\theta^j$,
\begin{align}
\!\!\!\!\!\!\!\!\!\!\!\!\!\!\!\!\!\!\!\!\!\!\!\!\!\!\!D_{\theta^i} \eta &=\int_{\tau}\!\! \sum_{k=0}^{|\tau|-1} \gamma^k f(\tau_{0:k};\theta^1,\theta^2) D_{\theta^i} (\log\pi(a_k^i|s_k;\theta^i)) Q(s_k,a_k^1,a_k^2;\theta^1\!\!,\theta^2) d\tau,\!\!\!\!\!\label{eq:CPGGradHessianVal1}\\
\!\! D_{\theta^i \theta^j} \eta &=\!(\wh 1)\!\!:\!\!\!\int_{\tau}\!\! \!\sum_{k=0}^{|\tau|-1}\!\!\! \gamma^k \! f(\!\tau_{0:k};\theta^1\!\!,\theta^2)D_{\theta^i}( \log\pi(\!a_k^i|s_k;\theta^i)) D_{\theta^j}( \log\pi(\!a_k^j|s_k;\theta^j))^{\!\top} \! Q(\!s_k,a_k^1,a_k^2;\theta^1\!\!,\theta^2) d\tau \nonumber \\
&\!\!\!\!\!\!\!\!\!\!\!\!\!\!\!\!\!+\!(\wh 2)\!:\!\!\! \int_{\tau}\!\!\sum_{k=1}^{|\tau|-1}\!\!\!\gamma^k \!f(\!\tau_{0:k};\theta^1\!,\!\theta^2) D_{\theta^i} (\log\!\! \prod_{l=0}^{k-1}\!\!\pi(\!a_{l}^i|s_l;\theta^i)) D_{\theta^j}(\log \pi(\!a_{k}^j|s_{k};\theta^j))\!^\top \! Q(\!s_k,a_{k}^1,a_{k}^2;\theta^1\!\!,\theta^2) d\tau \ \nonumber\\
&\!\!\!\!\!\!\!\!\!\!\!\!\!\!\!\!\!+\!(\wh 3)\!\!:\!\!\! \int_{\tau}\!\! \sum_{k=1}^{|\tau|-1}\!\!\!  \gamma^k  \!f(\!\tau_{0:k};\theta^1\!\!,\!\theta^2) D_{\theta^i} (\log \pi(\!a_{k}^i|s_{k};\theta^i)) D_{\theta^j} (\log\!\! \prod_{l=0}^{k-1}\!\!\pi(\!a_{l}^j|s_l;\theta^j))\!^\top \! Q(\!s_k,a_{k}^1,a_{k}^2;\theta^1\!\!,\theta^2) d\tau. \!\! \!\!\!\label{eq:CPGGradHessianVal}\!\!\!\!\!
\end{align}
\end{theorem}
Proof in \appx~\ref{appx: CPGGradHessianVal}. This theorem indicates that the terms in \eq~\ref{eqn: CGDSolution} can be computed using $Q$ function. In \eq~\ref{eq:CPGGradHessianVal}, ($\wh 1$) is the immediate interaction between players, ($\wh 2$) is the interaction of player $i$'s behavior up to time step $k$ with player $j$'s reaction at time step $k$, and the environment. ($\wh 3$) is the interaction of player $j$'s behavior upto time step $k$ with player $i$'s reaction at time step $k$, and the environment.
%
%

\cpg operates in epochs. At each epoch, \cpg deploys $\pi(\theta^1)$, $\pi(\theta^2)$ to collect trajectories, exploits them to estimate the $Q$,
$D_{\theta^i} \eta$, and $D_{\theta^i \theta^j} \eta$. 
Then \cpg follows the parameter updates in Eq.~\ref{eqn: CGDSolution} and updates $(\theta^1\!\!,\theta^2)$, and this process goes on to the next epoch (\alg\ref{alg:cpg_short}). Many variants of \PG approach uses baselines, and replace $Q$ with, the advantage function $A(s,a^1,a^2; \theta^1,\theta^2) = Q(s,a^1,a^2; \theta^1,\theta^2) - V(s; \theta^1,\theta^2)$,
Monte Carlo estimate of Q-V~\citep{baird1993}, empirical TD error or generalized advantage function (\GAE)~\citep{schulman2015highdimensional}. Our algorithm can be extended to this set up and in \appx~\ref{secappx: advantagebilinear} we show how \cpg can be accompanied with all the mentioned variants.

\subsection{Trust Region Competitive Policy Optimization}
\label{sec:trcpo}
	Trust region based policy optimization methods exploit the local Riemannian geometry of the parameter space to derive more efficient policy updates~\citep{kakade2002approximately,kakade2002natural,schulman2015trust}. In this section, we propose trust region competitive policy optimization (\trcpo), the \cpo generalization of \TRPO\citep{schulman2015trust}, that exploits the local geometry of the competitive objective to derive more efficient parameter updates. 

%
\begin{lemma}\label{lemma: trcpo_advantage}
Given the advantage function of policies $\pi(\theta^1),\pi(\theta^2)$,  $\forall({\theta'}^1,{\theta'}^2)\in\Theta^1\times\Theta^2$ we have,
\begin{align}\label{eqn: TRCPOAdvantage}
\eta(\theta'^1,\theta'^2) = \eta(\theta_{}^1,\theta_{}^2) +
\mathbb{E}_{\tau \sim  f(\cdot;\theta'^1,\theta'^2)}\sum\nolimits_{k=0}^{|\tau|-1} \gamma^k A(s,a^1,a^2;\theta_{}^1,\theta_{}^2).
\end{align}
\end{lemma}
Proof in \appx~\ref{appx: trcpo_advantage}. \eq\ref{eqn: TRCPOAdvantage} indicates that, considering our current policies $(\pi(\theta^1),\pi(\theta^1))$, having access to their advantage function, and also samples from $f(\cdot;\theta'^1,\theta'^2)$ of any $({\theta'}^1,{\theta'}^2)$ (without rewards), we can directly compute $\eta(\theta'^1,\theta'^2)$ and optimize for $(\theta'^1,\theta'^2)$. However, in practice, we might not have  access to $f(\cdot;\theta'^1,\theta'^2)$ for all $({\theta'}^1,{\theta'}^2)$ to accomplish the optimization task, therefore, direct use of \eq\ref{eqn: TRCPOAdvantage} is not favorable. Instead, we define a surrogate objective function, $L_{\theta^1,\theta^2}(\theta'^1,\theta'^2)$,
\begin{align}
\label{eqn: SurrogateObjective}
\hspace{-2.00mm}
L_{\theta^1,\theta^2}(\theta'^1,\theta'^2)\!\!= \!\eta(\theta_{}^1,\theta_{}^2)\! +\!
\mathbb{E}_{\tau \sim  f(\cdot;\theta^1,\theta^2)} \!\!\! \sum_{k=0}^{|\tau|-1} \!\! \gamma^k \mathbb{E}_{ \pi(a'^1_k|s_k;\theta'^1),\pi(a'^2_k|s_k;\theta'^2)} A(s_k,a'^1_k,a'^2_k;\theta_{}^1,\theta_{}^2),\!\!
\end{align}
which can be computed using trajectories of our current polices $\pi(\theta^1),\pi(\theta^2)$.
$L_{\theta_{}^1,\theta_{}^2}({\theta'}^1,{\theta'}^2)$ is an object of interest in \PG~\citep{kakade2002approximately,schulman2015trust} since mainly its gradient is equal to that of $\eta(\theta^1,\theta^2)$ at $(\theta^1,\theta^2)$, and as stated in the following theorem, it can carefully be used as a surrogate of the game value.
For KL divergence 
$D_{KL}((\theta_{}^1,\theta_{}^2), (\theta'^1,\theta'^2)) := \int_{\tau} f(\tau,\theta_{}^1,\theta_{}^2)\log\big(f(\tau,\theta_{}^1,\theta_{}^2)/f(\tau,\theta'^1,\theta'^2)\big) d\tau$, we have, 
\begin{theorem}\label{theorem: BoundTRCPO}
$L_{\theta_{}^1,\theta_{}^2}({\theta'}^1,{\theta'}^2)$ approximate $\eta({\theta'}^1,{\theta'}^2)$ up to the following error upper bound
\begin{align}
\label{eqn: BoundTRCPO}
|\eta(\theta'^1,\theta'^2)-L_{\theta_{}^1\theta_{}^2}(\theta'^1,\theta'^2)| \leq \epsilon/\sqrt{2} \sqrt{D_{KL}((\theta'^1,\theta'^2), (\theta_{}^1,\theta_{}^2))},
\end{align}
with constant $\epsilon := \max_{\tau} \sum_{k}^{|\tau|} \gamma^k \mathbb{E}_{ \pi(a'^1_k|s_k;\theta'^1),\pi(a'^2_k|s_k;\theta'^2)} A(s_k,a'^1_k,a'^2_k;\theta_{}^1,\theta_{}^2)$.
\end{theorem}
Proof in \appx~\ref{appx: ProofBoundTRPO}. This theorem states that we can use $L_{\theta_{}^1\theta_{}^2}(\theta'^1,\theta'^2)$ to optimize for $\eta(\theta'^1,\theta'^2)$ as long as we keep the $(\theta'^1,\theta'^2)$ in the vicinity of $\theta^1,\theta^2$.
Similar to single agent \TRPO ~\citep{schulman2015trust}, we optimize for $L_{\theta_{}^1\theta_{}^2}(\theta'^1,\theta'^2)$, while constraining the KL divergence, $D_{KL}((\theta_{}^1,\theta_{}^2), (\theta'^1,\theta'^2))\leq \delta'$, i.e.,
\begin{align}\label{eqn: TRCPOFinalObjective}
    \max\nolimits_{\theta'^1\in\Theta^1} \min\nolimits_{\theta'^2\Theta^1}  L_{\theta^1\theta^2}(\theta'^1,\theta'^2),~    \text{subject to: } D_{KL}((\theta_{}^1,\theta_{}^2),(\theta'^1,\theta'^2)) \leq \delta'.
\end{align}
The \gda  generalizing of \TRPO uses a linear approximation of $L_{\theta^1\theta^2}(\theta'^1,\theta'^2)$ to approach this optimization, which again ignores the players' interactions. In contrast, \trcpo considers the game theoretic nature of this optimization, and uses a bilinear approximation. \trcpo operates in epochs. At each epoch, \trcpo deploys $(\pi(\theta^1),\pi(\theta^2))$ to collect trajectories, exploits them to estimate $A$ (or \GAE), then updates parameters accordingly, \alg\ref{alg: trcpo_short}. 
%
(For more details, please refer to \appx~\ref{appx: trcpoOptimization}.)



\section{Experiments}
\label{sec:exp_cpg_trcpo}
	We empirically study the performance of \cpg and \trcpo and their counterparts \gda and \trpogda, on six games, ranging from single-state repeated games to general sequential games, and tabular games to infinite/continuous states/action games. They are 1) linear-quadratic(LQ) game, 2) bilinear game, 3) matching pennies (MP), 4) rock paper scissors (RPS),  5) Markov soccer, and 6) car racing. These games are representative enough that their study provides insightful conclusions, and challenging enough to highlight the core difficulties and interactions in competitive games. 

We show that \cpg and \trcpo converge to stable points, and learn opponent aware strategies, whereas \gda's and \trpogda's greedy approach shows poor performance and even diverge in bi-linear, MP, and RPS games. For the LQ game, when \gda does not diverge, it almost requires 1.5 times the amount of samples, and is $1.5$ times slower than \cpg.
In highly strategic games, where players' policies are tightly coupled, we show that \cpg and \trcpo learn much better interactive strategies. In the soccer game, \cpg and \trcpo players learn to defend, dodge and score goals, whereas \gda and \trpogda players learn how to score when they are initialized with the ball, and give way to the other player otherwise. In the car racing, while \gda and \trpogda show poor performance, \cpg and \trcpo produce competing players, which learn to block and fake each other. Overall, we observe \cpg and \trcpo considerably outperform their counterparts in terms of convergence, learned strategies, and gradient dynamics. Note that in self-play, one player plays against itself using one policy model (neural network) for both players, is a special case of \cpo and we provide its detailed study in \appx~\ref{appx: selfplay}.

We implemented all algorithms in Pytorch ~\citep{NEURIPS2019_9015}, and made the code and the videos public\footnote{Videos are available here: \href{https://sites.google.com/view/rl-copo}{https://sites.google.com/view/rl-copo}}. 
In our implementation, we deploy Pytorch's autograd package and Hessian vector product to efficiently obtain gradients and Hessian vector products to compute the bilinear terms in the optimizer.  Moreover, we use the conjugate gradient trick to efficiently computed the inverses-matrix vector product in \eq\ref{eqn: CGDSolution} which incurs a minimal computational overhead (see \citep{CG_Jonathan1994} for more details). To improve computation times, we compute inverse-matrix vector product only for one player strategy, and use optimal counter strategy for other player $\Delta \theta^2$ which is computed without an inverse matrix vector product. Also, the last optimal strategy can be used to warm start the conjugate gradient method which improves convergence times. We provide efficient implementation for both \cpo- and \gda-based methods, where \cpo incurs 1.5 times extra computation per batch. 

\begin{figure}[h]
	\hspace{-3.00mm}
	\centering
    \begin{subfigure}[t]{0.26\columnwidth}
  	\centering
  	\includegraphics[width=\linewidth]{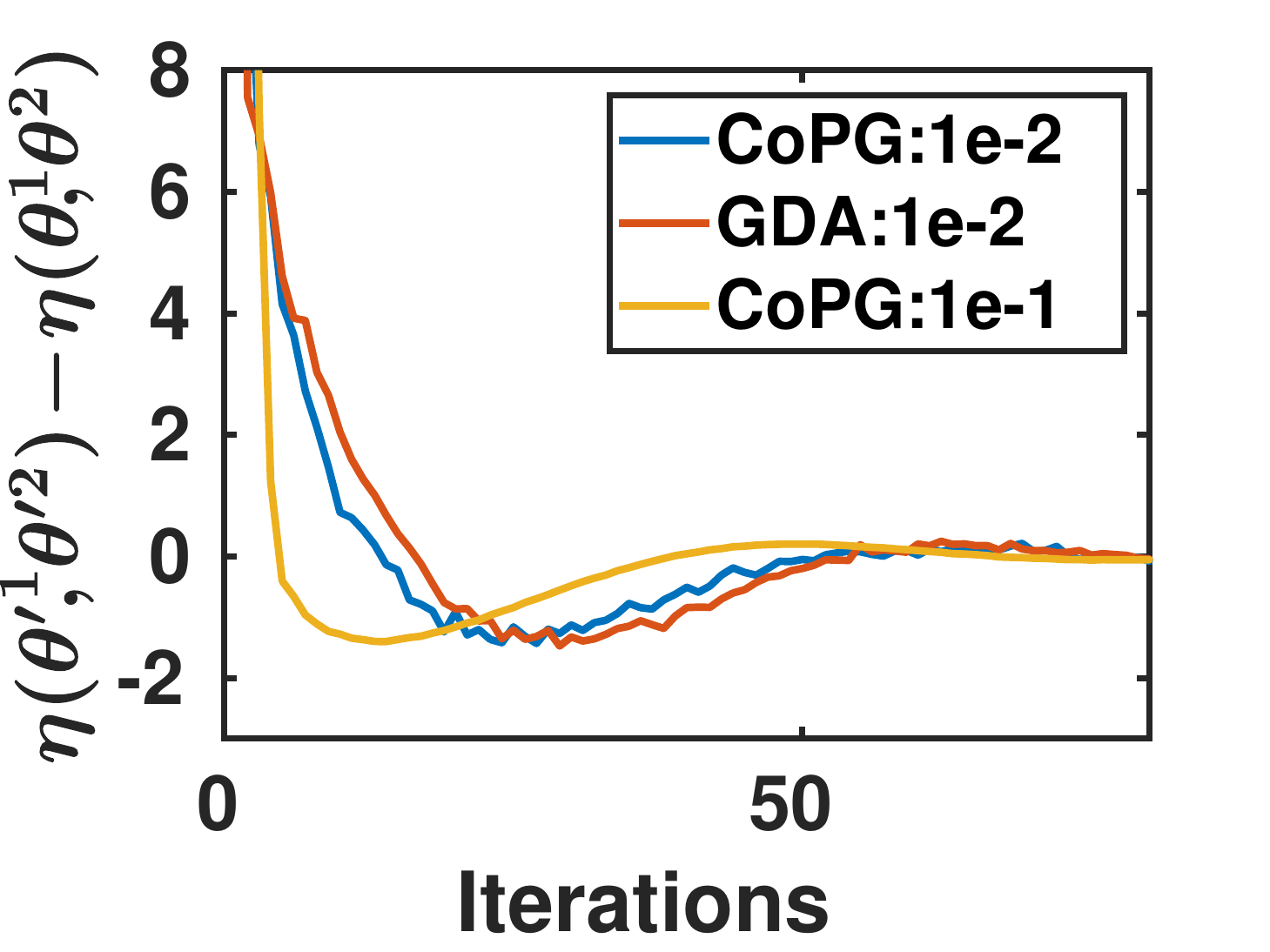}
    \caption{}
    \label{fig: LQlr_1e2}
    \end{subfigure}
    \hspace{-6.00mm}
~
    \begin{subfigure}[t]{0.26\columnwidth}
  	\centering
  	\includegraphics[width=\linewidth]{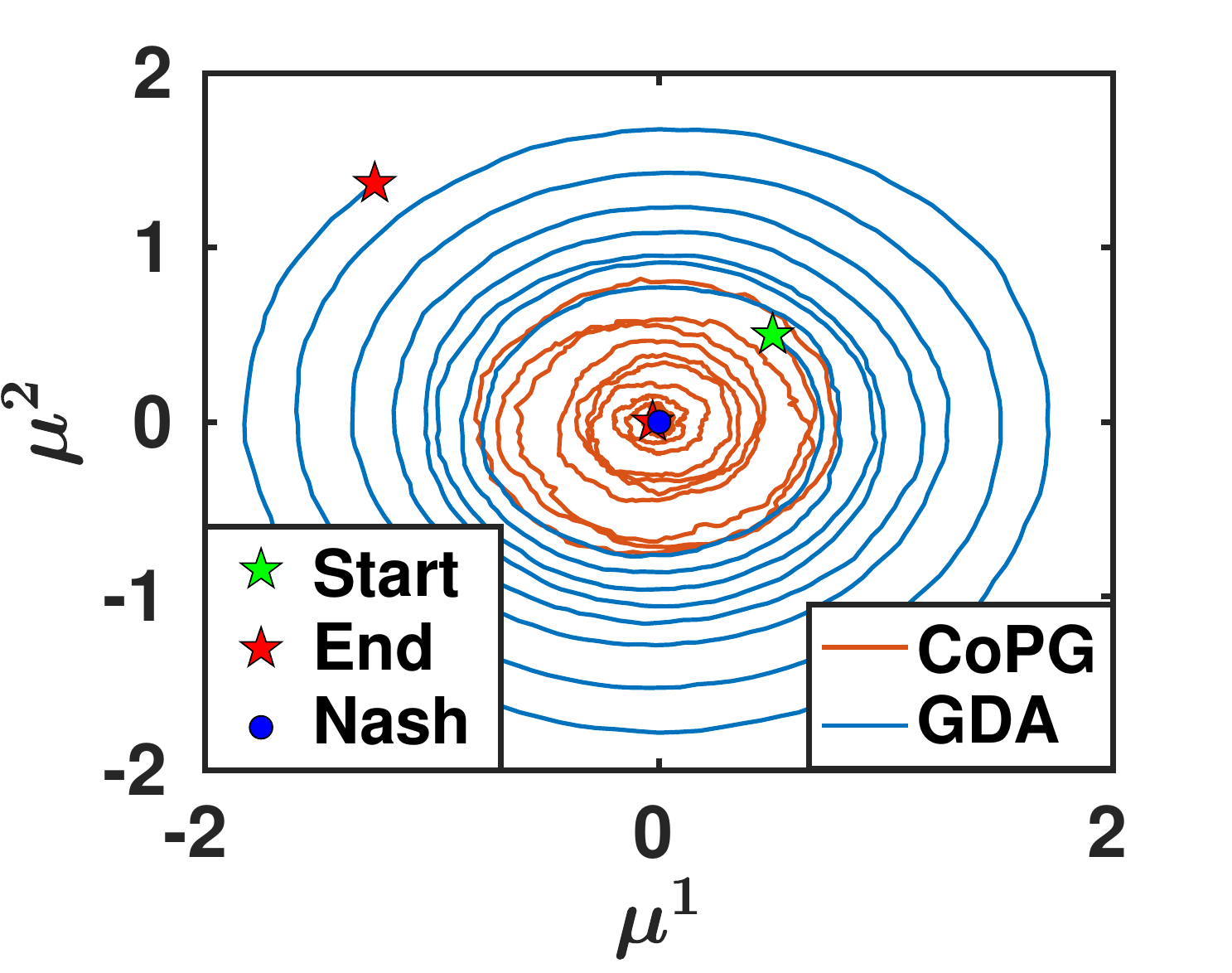}
    \caption{}
    \label{fig: combined_bilinear_mat}
    \end{subfigure}
    \hspace{-6.00mm}
   ~
    \begin{subfigure}[t]{0.26\columnwidth}
  	\centering
  	\includegraphics[width=\linewidth]{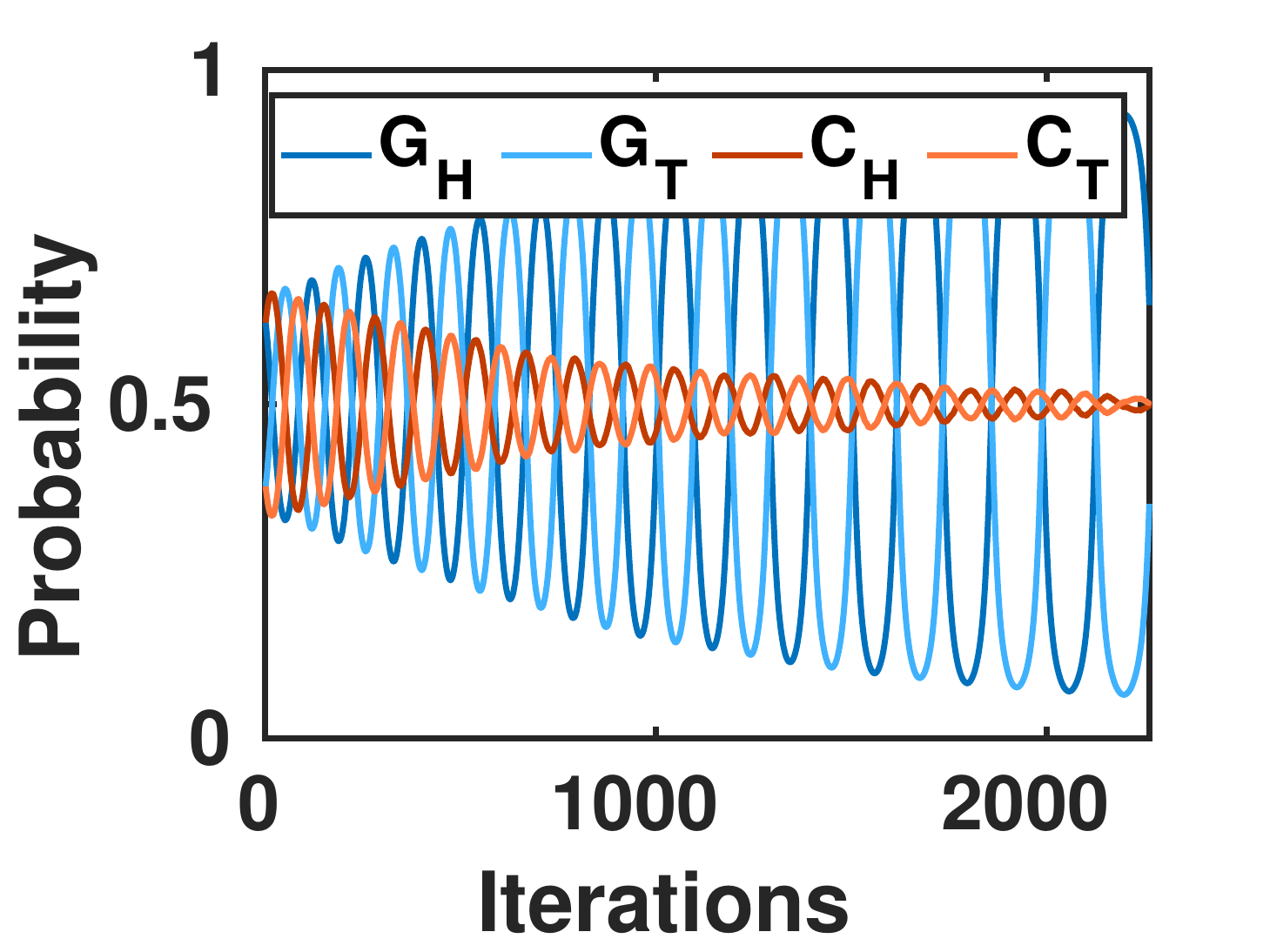}
    \caption{}
    \label{fig: combine_matching_pennies}
    \end{subfigure}
    \hspace{-6.00mm}
~
  \begin{subfigure}[t]{0.26\columnwidth}
  	\centering
  	\includegraphics[width=\linewidth]{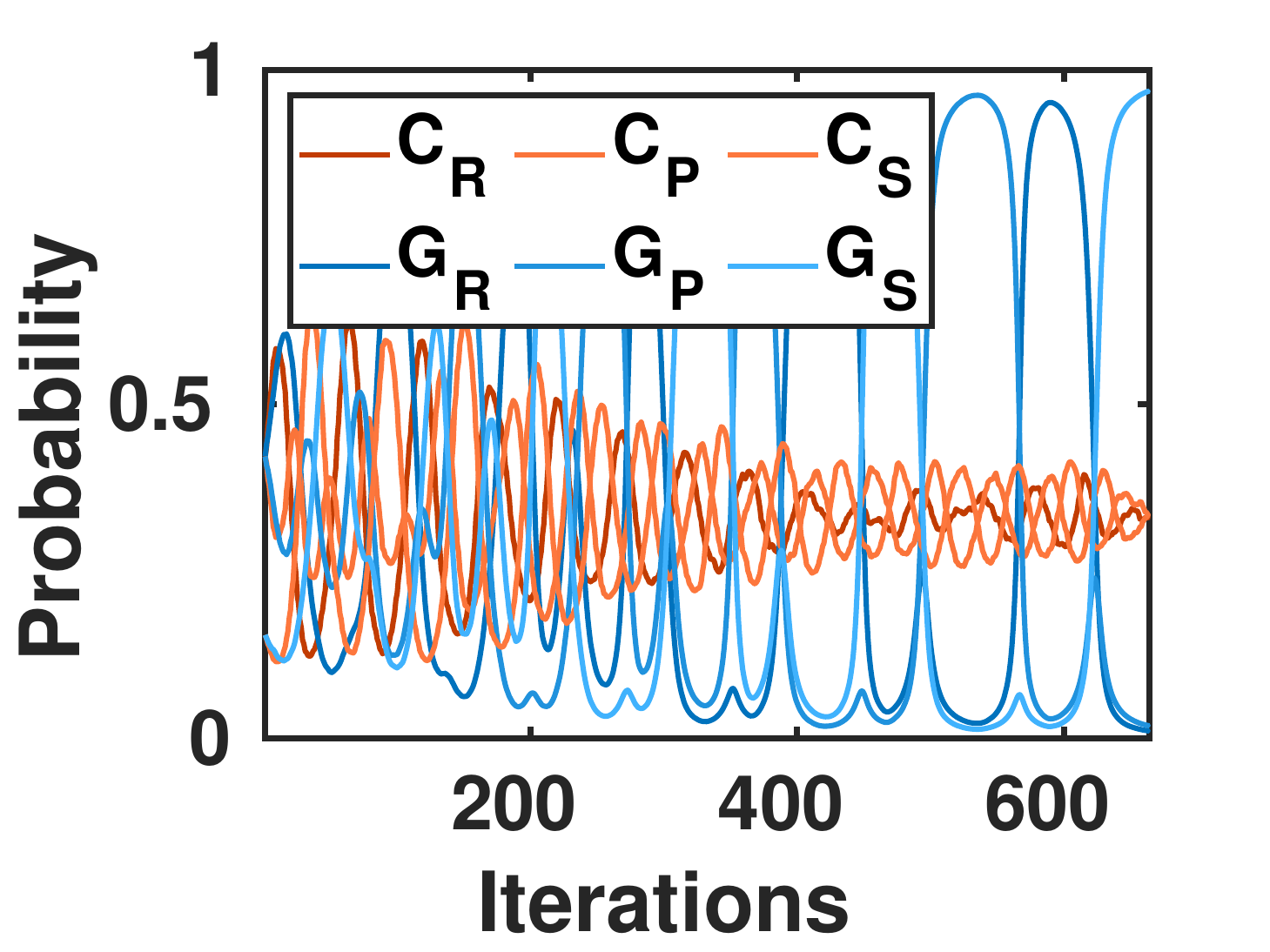}  
    \caption{}
    \label{fig: combined_rps}
  \end{subfigure}
  \hspace{-4.00mm}
  \vspace{-3.00mm}
  \caption{During training \cpg (C) vs \gda (G) on a) LQ game, difference in game objective due to policy update for $\alpha=1e-1,1e-2$ b) Bilinear game, $\mu^1$ vs $\mu^2$ c) MP, probability of selecting Head(H) and Tail(T)  d) RPS, probability of selecting rock(R), paper(P) and scissors(S).} 
  \hspace{-4.00mm}
  \vspace{-6.00mm}
\end{figure}

\textbf{Zero-sum LQ game} is a continuous state-action linear dynamical game between two players, where \gda, with considerably small learning rate, has favorable convergence guarantees~\citep{zhang2019policyZerosum}. This makes the LQ game an ideal platform to study the range of allowable step sizes and convergence rate of \cpg and \gda.
We show that, with increasing learning rate, \gda generates erratic trajectories and policy updates, which cause instability (see ``$\oslash$'' in \tab~\ref{tab: LQgameLR}), whereas \cpg is robust towards this behavior. \fig~\ref{fig: LQlr_1e2} shows that \cpg dynamics are not just faster at the same learning rate but more importantly, \cpg can potentially take even larger steps. 

\textbf{Bilinear game} is a state-less strongly non-cooperative game, where \gda is known to diverge~\citep{BalduzziHamiltonian}. In this game, reward $r(a^1,a^2)=\langle{a^1},a^2\rangle$ where $a^1\sim\mathcal{N}(\mu_1,\sigma_1)$ and $a^2\sim\mathcal{N}(\mu_2,\sigma_2)$, and $(\mu_1,\sigma_1)$, $(\mu_2,\sigma_2)$ are policy parameters. 
We show that \gda diverges for all learning rates, whereas \cpg converges to the unique Nash equilibrium \fig~\ref{fig: combined_bilinear_mat}.

\textbf{Matching pennies} and \textbf{Rock paper scissors}, are challenging matrix games with mixed strategies as Nash equilibria, demand opponent aware optimization.\footnote{Traditionally, fictitious and counterfactual regret minimization approaches have been deployed~\citep{CFR_RPS}.} We show that \cpg and \trcpo both converge to the unique Nash equilibrium of MP \fig~\ref{fig: combine_matching_pennies} and RPS \fig~\ref{fig: combined_rps}, whereas \gda and \trpogda diverges (to a sequence of polices that are exploitable by deterministic strategy).
Detailed empirical study, formulation and explanation of these four games can be found in \appx~\ref{appxsec: NashGame}.
\begin{figure}[!h]
\hspace{-4.00mm}
	\centering
    \begin{subfigure}[t]{0.26\columnwidth}
  	\centering
  	\includegraphics[width=\linewidth]{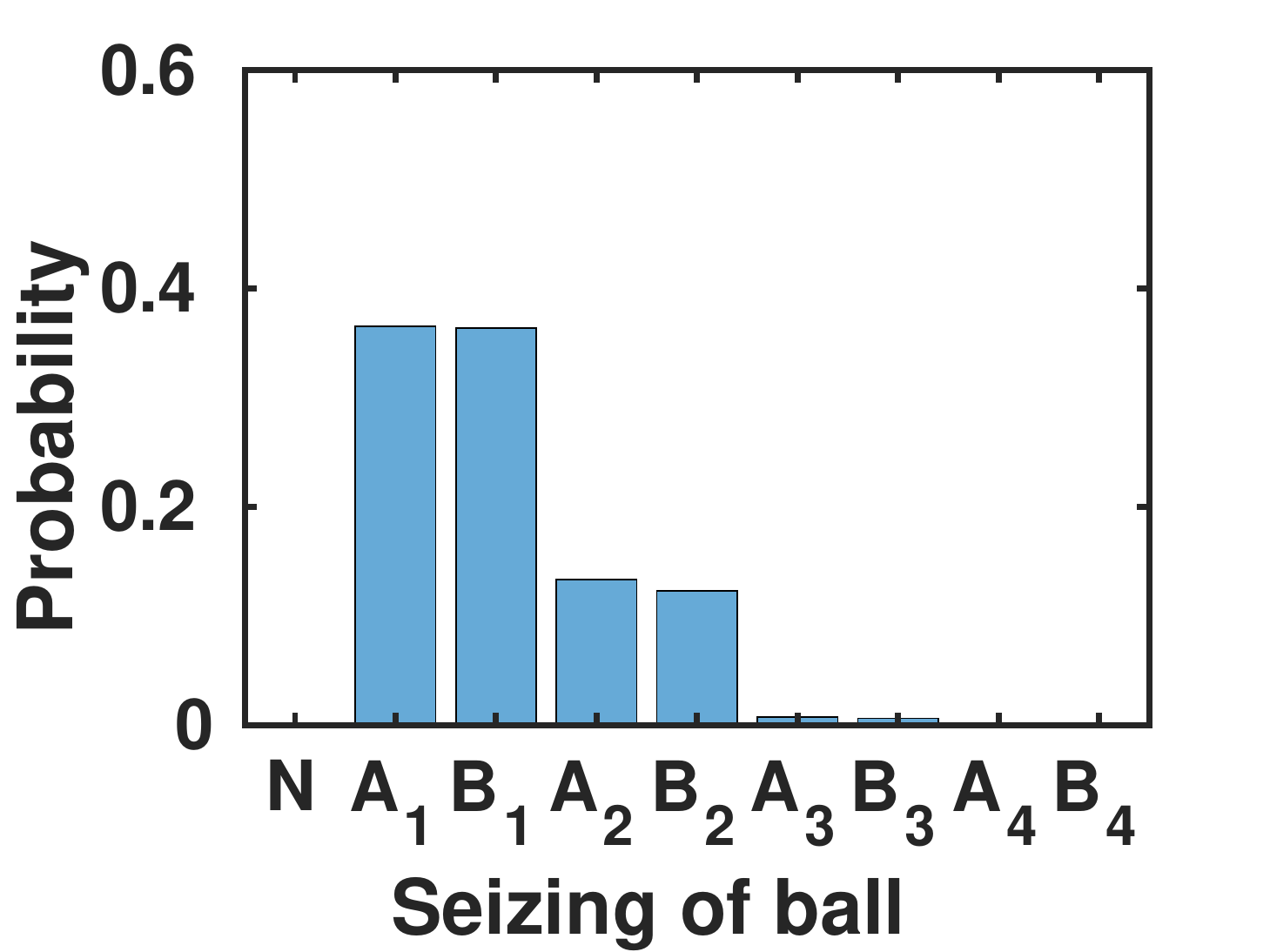}
    \caption{\cpg vs \cpg}
    \label{Fig: cpg_cpg_interaction_plot}
    \end{subfigure}
    \hspace{-5.00mm}
  ~
    \begin{subfigure}[t]{0.26\columnwidth}
  	\centering
  	\includegraphics[width=\linewidth]{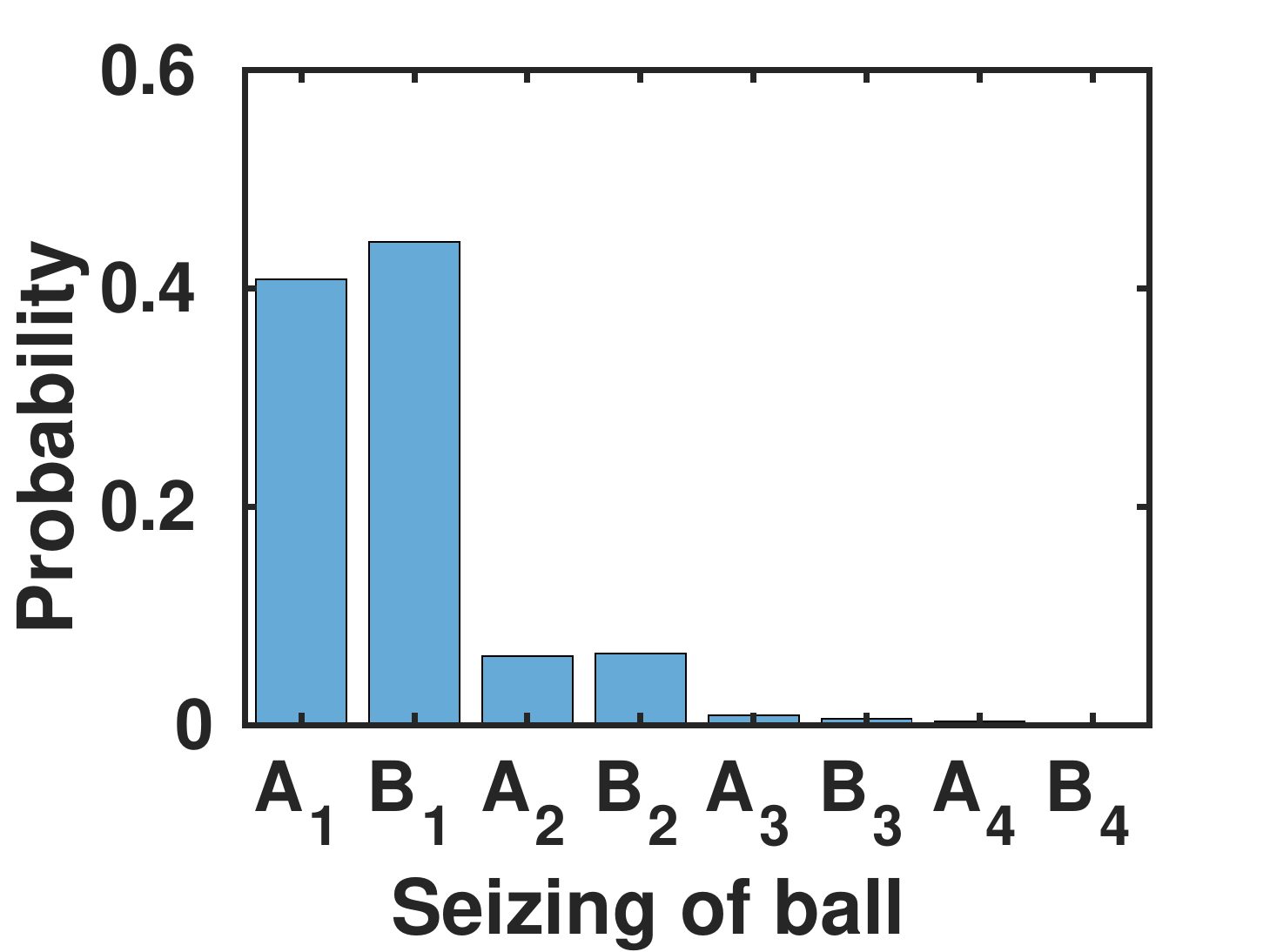}
    \caption{\gda vs \gda}
    \label{Fig: sgd_sgd_interaction_plot}
    \end{subfigure}
    \hspace{-5.00mm}
    ~
    \begin{subfigure}[t]{0.26\columnwidth}
  	\centering
  	\includegraphics[width=\linewidth]{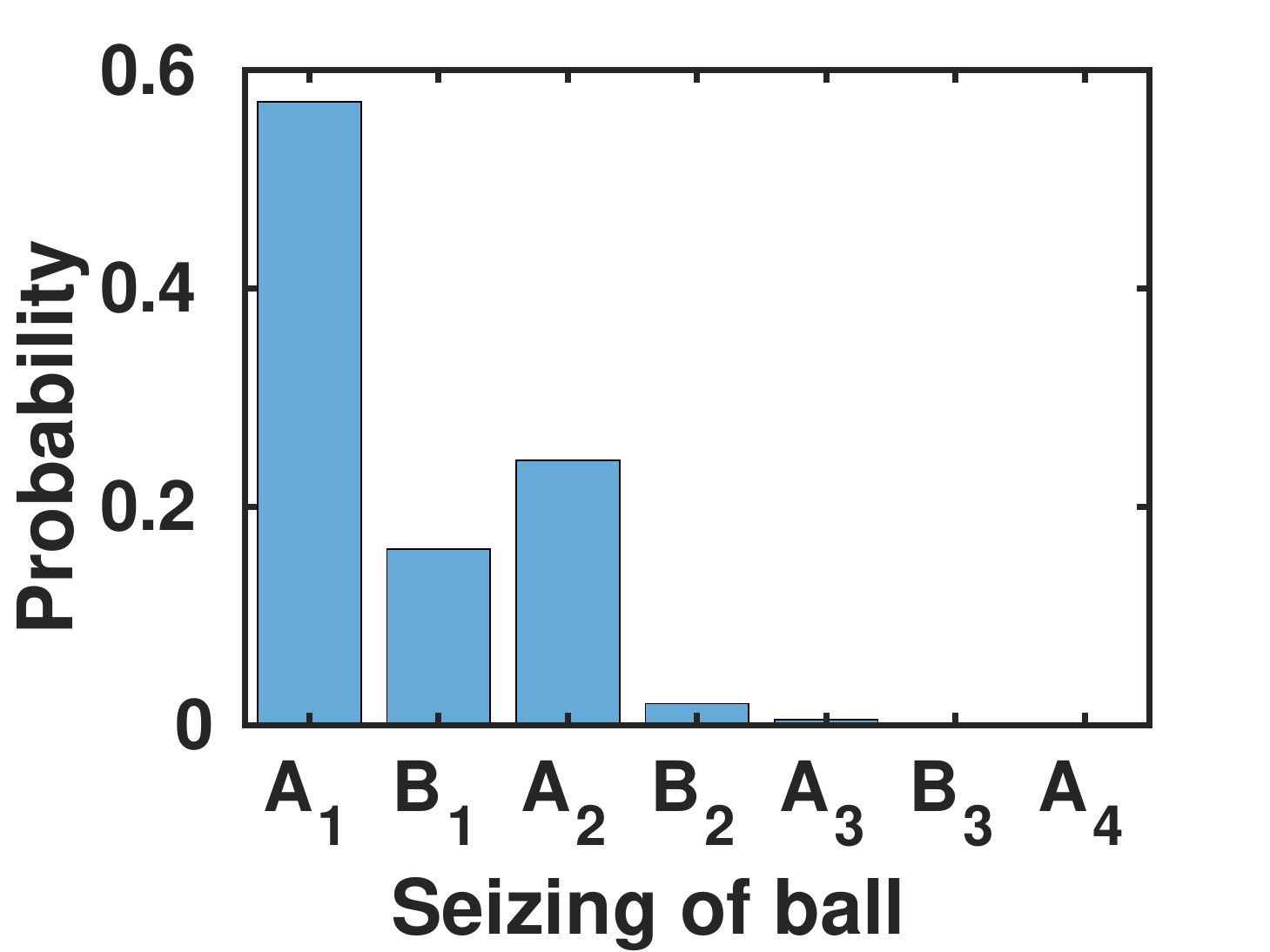}
    \caption{\cpg vs \gda}
    \label{Fig: cpg_sgd_interaction_plot}
    \end{subfigure}
    \hspace{-5.00mm}
~
  \begin{subfigure}[t]{0.26\columnwidth}
  	\centering
  	\includegraphics[width=\linewidth]{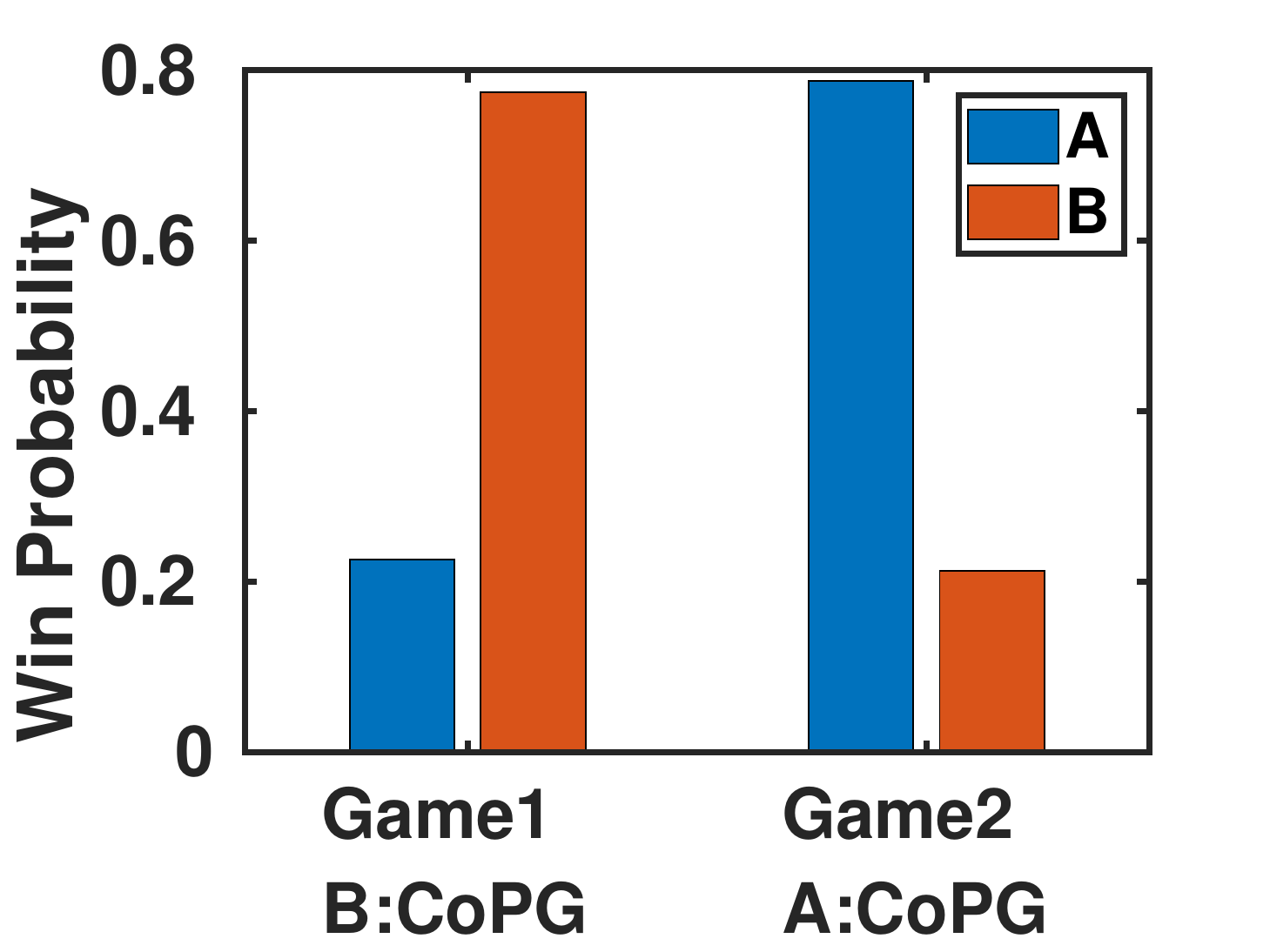}  
    \caption{\gda vs \cpg}
    \label{Fig: sgd_cpg_winning_plot}
  \end{subfigure}
  \hspace{-4.00mm}

\hspace{-4.00mm}
	\centering
    \begin{subfigure}[t]{0.26\columnwidth}
  	\centering
  	\includegraphics[width=\linewidth]{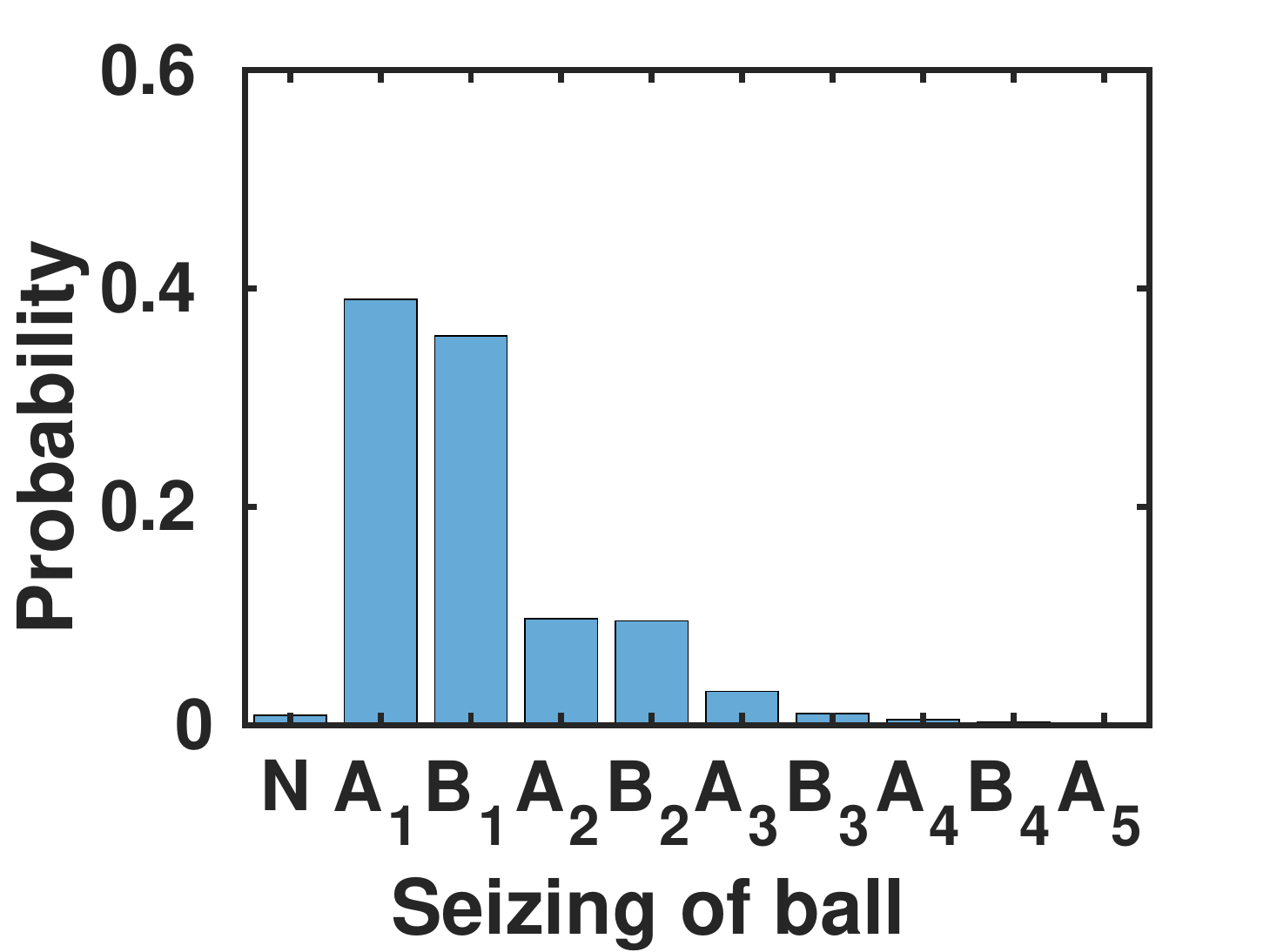}
    \caption{\trcpo vs \trcpo}
    \label{Fig: trcpo_trcpo_gmres_interaction_plot}
    \end{subfigure}
    \hspace{-5.00mm}
    ~
    \begin{subfigure}[t]{0.26\columnwidth}
  	\centering
  	\includegraphics[width=\linewidth]{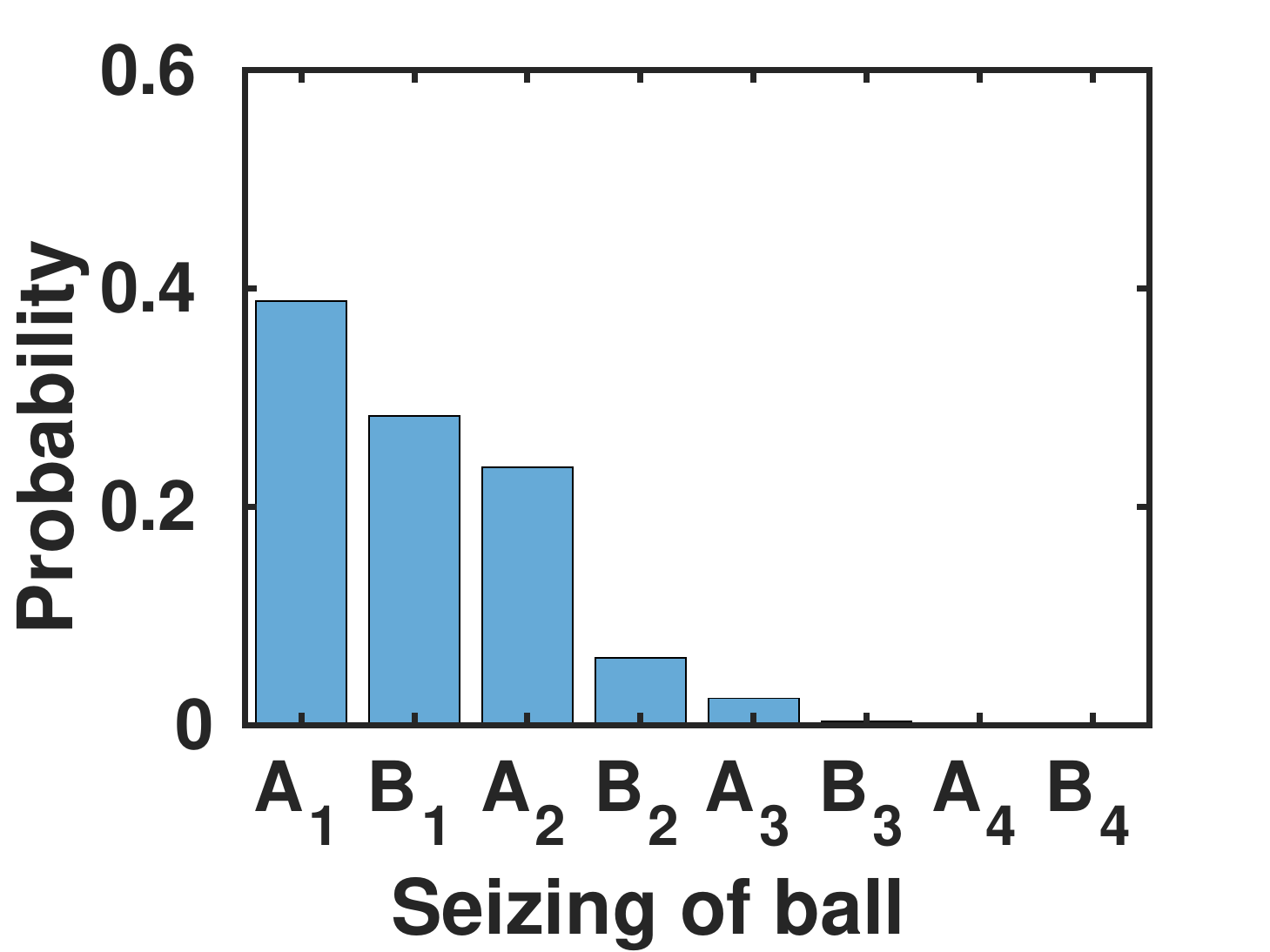}
    \caption{\trpogda vs \trpogda}
    \label{Fig: trpo_trpo_gmres_interaction_plot}
    \end{subfigure}
    \hspace{-5.00mm}
    ~
    \begin{subfigure}[t]{0.26\columnwidth}
  	\centering
    \includegraphics[width=\linewidth]{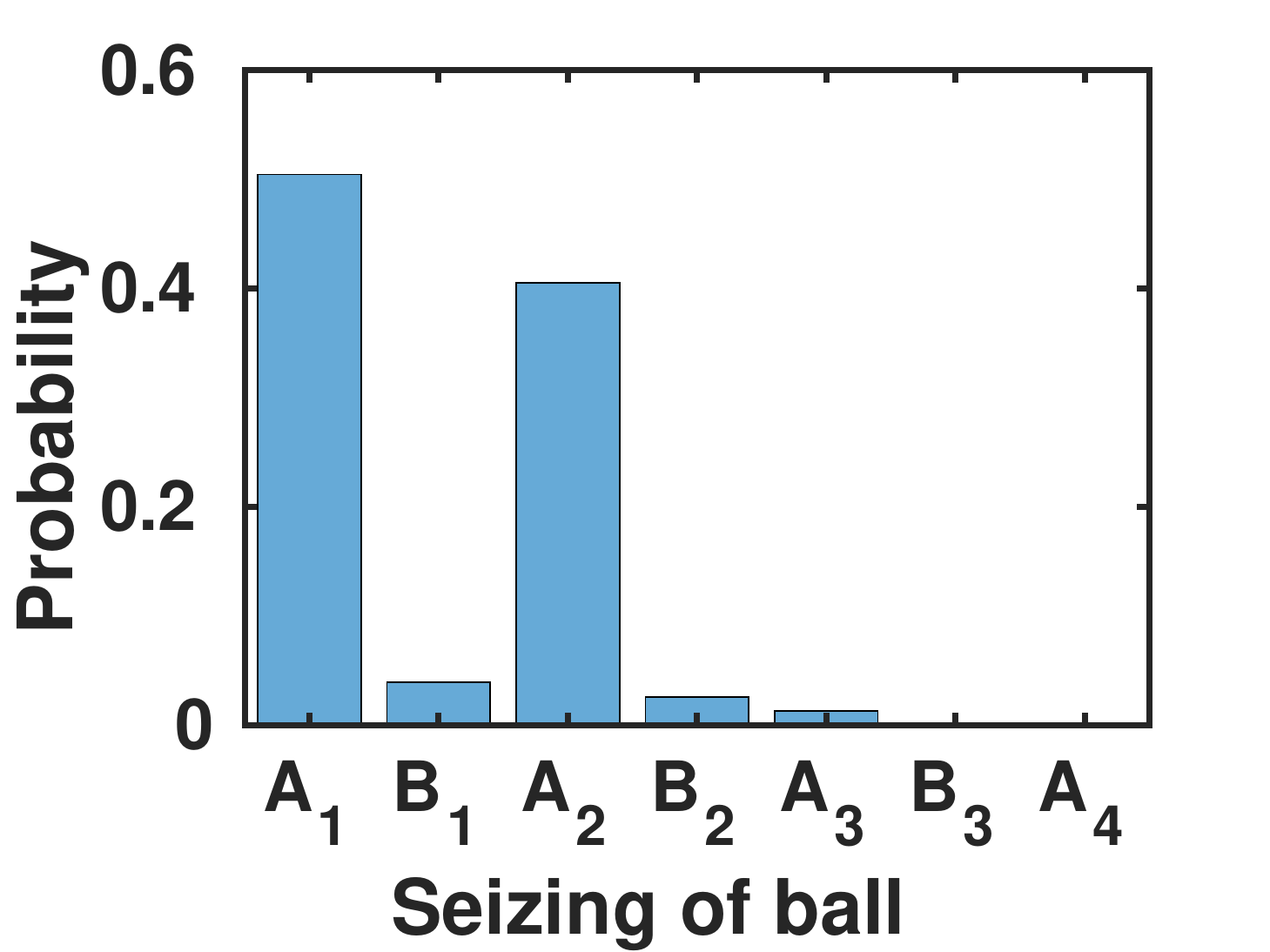}
    \caption{\trcpo vs \trpogda}
    \label{Fig: trpo_trcpo_gmres_interaction_plot}
    \end{subfigure}
    \hspace{-5.00mm}
~
  \begin{subfigure}[t]{0.26\columnwidth}
  	\centering
  	\includegraphics[width=\linewidth]{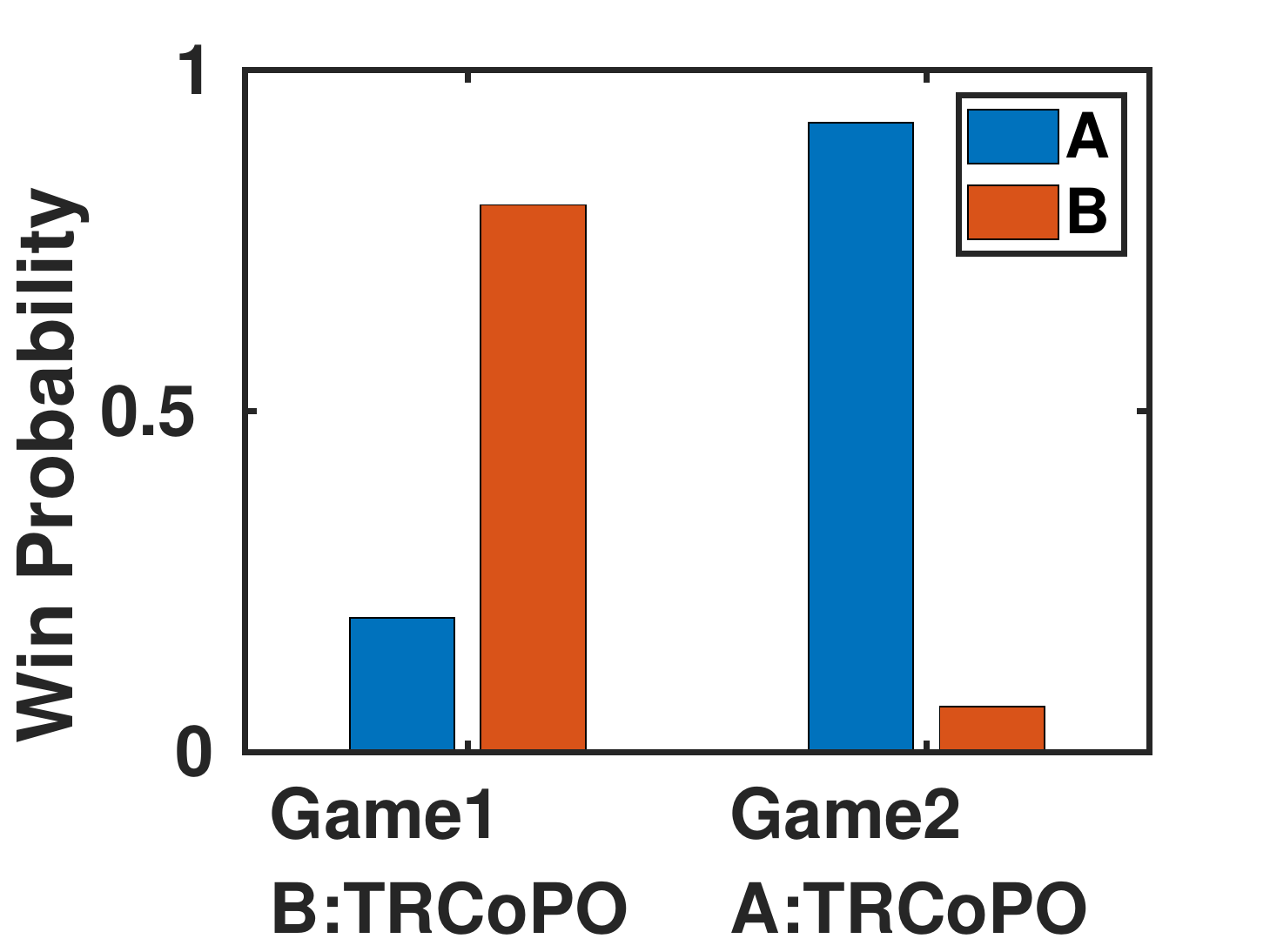}  
    \caption{\trpogda vs \trcpo}
    \label{Fig: trpo_trcpo_winning_plot}
  \end{subfigure}
  \hspace{-5.00mm}
  \vspace{-2.00mm}
  \caption{ a-c), e-g) Interaction plots, representing probability of seizing ball in the game between $A$ vs $B$. X-axis convention, for player $A$. $A_1$: $A$ scored goal, $A_2$: $A$ scored goal after seizing ball from $B$, $A_3$: $A$ scored goal by seizing ball from $B$ which took the ball from $A$ and so forth. Vice versa for player $B$. N: No one scored goal both kept on seizing ball. d),h) probability of games won. 
  }
  \vspace{-1.0em}
  \label{Fig: InteractionBilinear}
\end{figure}

\label{sec: CPGMarkovSoccer}
\begin{wrapfigure}{r}{0.3\textwidth}
    \begin{subfigure}[t]{0.30\columnwidth}
  	\includegraphics[width=\linewidth]{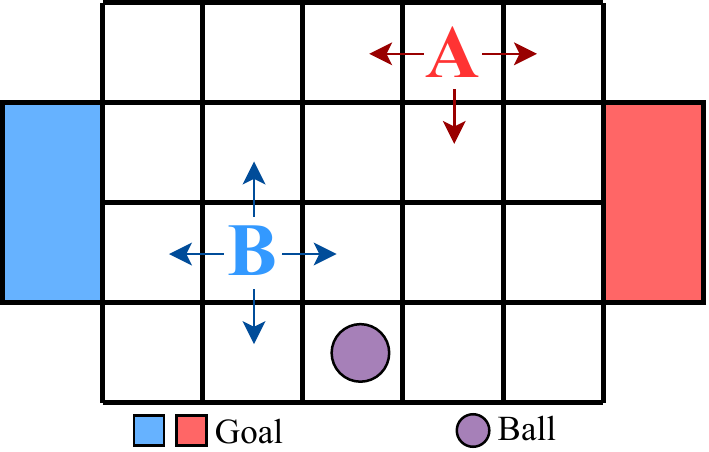}
    \end{subfigure}
    \hspace{-2.2cm}
  \vspace*{-.5em}
  \caption{Markov Soccer}
  \label{fig: Soccer}
  \vspace{-1em}
\end{wrapfigure}
\textbf{Markov soccer} game,~\fig~\ref{fig: Soccer}, consists of players A and B that are randomly initialised in the field, that are supposed to pick up the ball and put it in the opponent's goal~\citep{littmansoccer,HeHeOppo}. 
The winner is awarded with +1 and the loser with -1 (see \appx~\ref{appxsec: soccer} for more details). Since all methods converge in this game, it is a suitable game to compare learned strategies.
In this game, we expect a reasonable player to learn sophisticated strategies of defending, dodging and scoring. 
For each method playing against their counterparts, e.g., \cpg against \cpg and \gda, \fig~\ref{Fig: InteractionBilinear} shows the number of times the ball is seized between the players before one player finally scores a goal, or time-out.
In (\ref{Fig: cpg_cpg_interaction_plot}), \cpg vs \cpg, we see agents seize ball, twice the times as compared to (\ref{Fig: sgd_sgd_interaction_plot}),\gda vs \gda (see $A_2$ and $B_2$ in (\ref{Fig: cpg_cpg_interaction_plot}) and (\ref{Fig: sgd_sgd_interaction_plot})).
In the matches \cpg vs \gda (\ref{Fig: cpg_sgd_interaction_plot}), \cpg trained agent could seize the ball from \gda agent ($A_2$) nearly 12 times more due to better seizing and defending strategy, but \gda can hardly take the ball back from \cpg ($B_2$) due to a better dodging strategy of the \cpg agent. Playing \cpg agent against \gda one, we observe that \cpg wins more than 74\% of the games~\fig~\ref{Fig: sgd_cpg_winning_plot}. We observe a similar trend for trust region based methods \trcpo and \trpogda, playing against each other (a slightly stronger results of 85\% wins, $A_2$ column in Figs~\ref{Fig: trpo_trcpo_gmres_interaction_plot}, ~\ref{Fig: cpg_sgd_interaction_plot}).
We also compared \cpg with \maddpg~\citep{lowe2017multi} and \lola~\citep{FoersterLOLA}. We observe that the \maddpg learned policy behaves similar to \gda, and loses 80\% of the games to \cpg's (\fig~\ref{fig: EX_cpg_maddpg_winning_plot}). \lola learned policy, with its second level reasoning, performs better than \gda, but lose to \cpg 72\% of the matches. For completeness, we also compared \gdapg, \cpg, \trpogda, and \trcpo playing against each other. \trcpo performs best,\cpg was runner up, then \trpogda, and lastly \gda (see \appx~\ref{appxsec: soccer}).
\begin{figure}[h!]
\vspace{-0.5em}
~\hspace{-7.00mm}
	\centering
    \begin{subfigure}[t]{0.21\columnwidth}
  	\centering
  	\includegraphics[width=\linewidth]{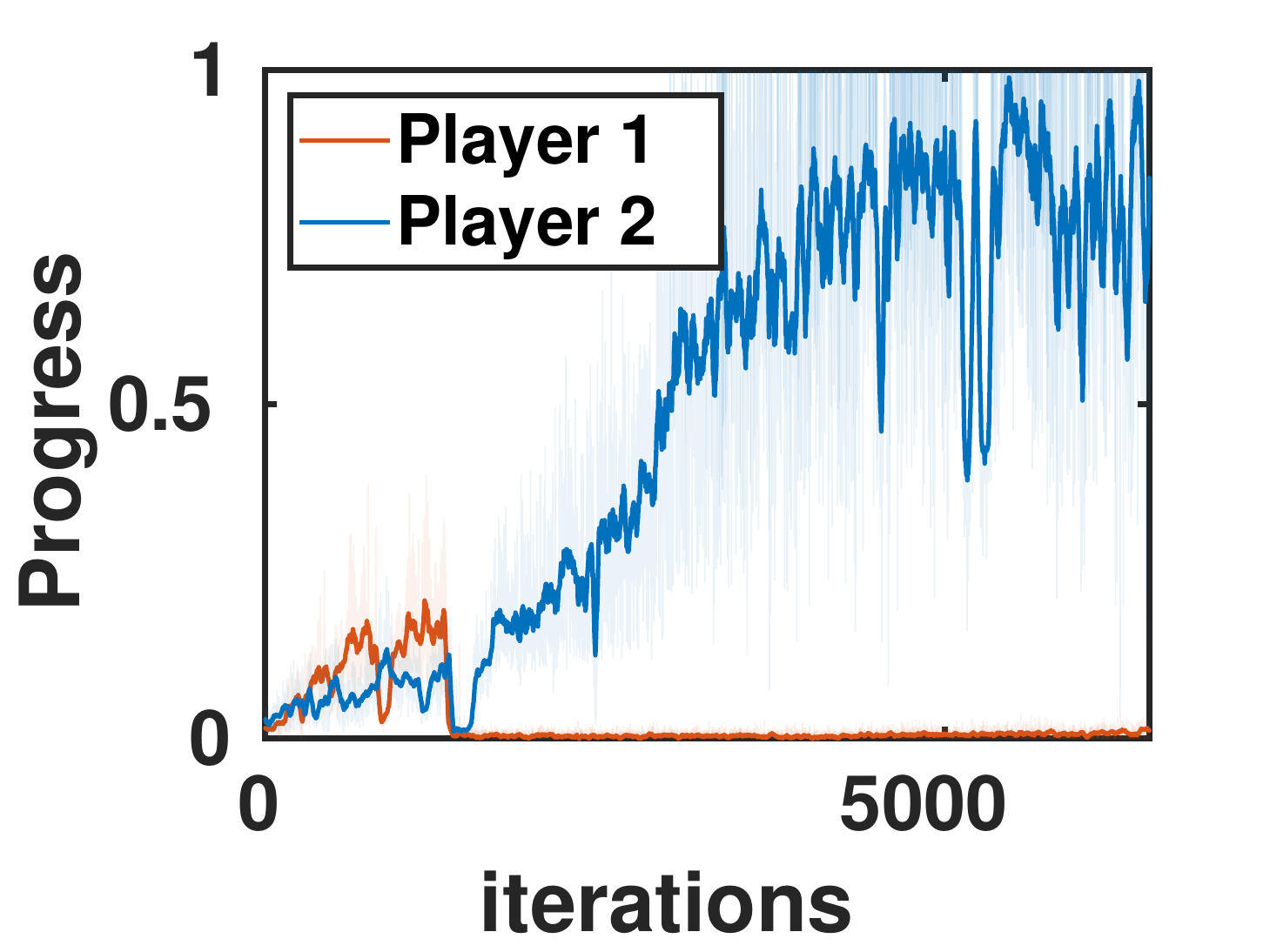}
    \caption{\gda}
    \label{fig: orca_sgd}
    \end{subfigure}
    ~\hspace{-6.20mm}
    ~
    \begin{subfigure}[t]{0.21\columnwidth}
  	\centering
  	\includegraphics[width=\linewidth]{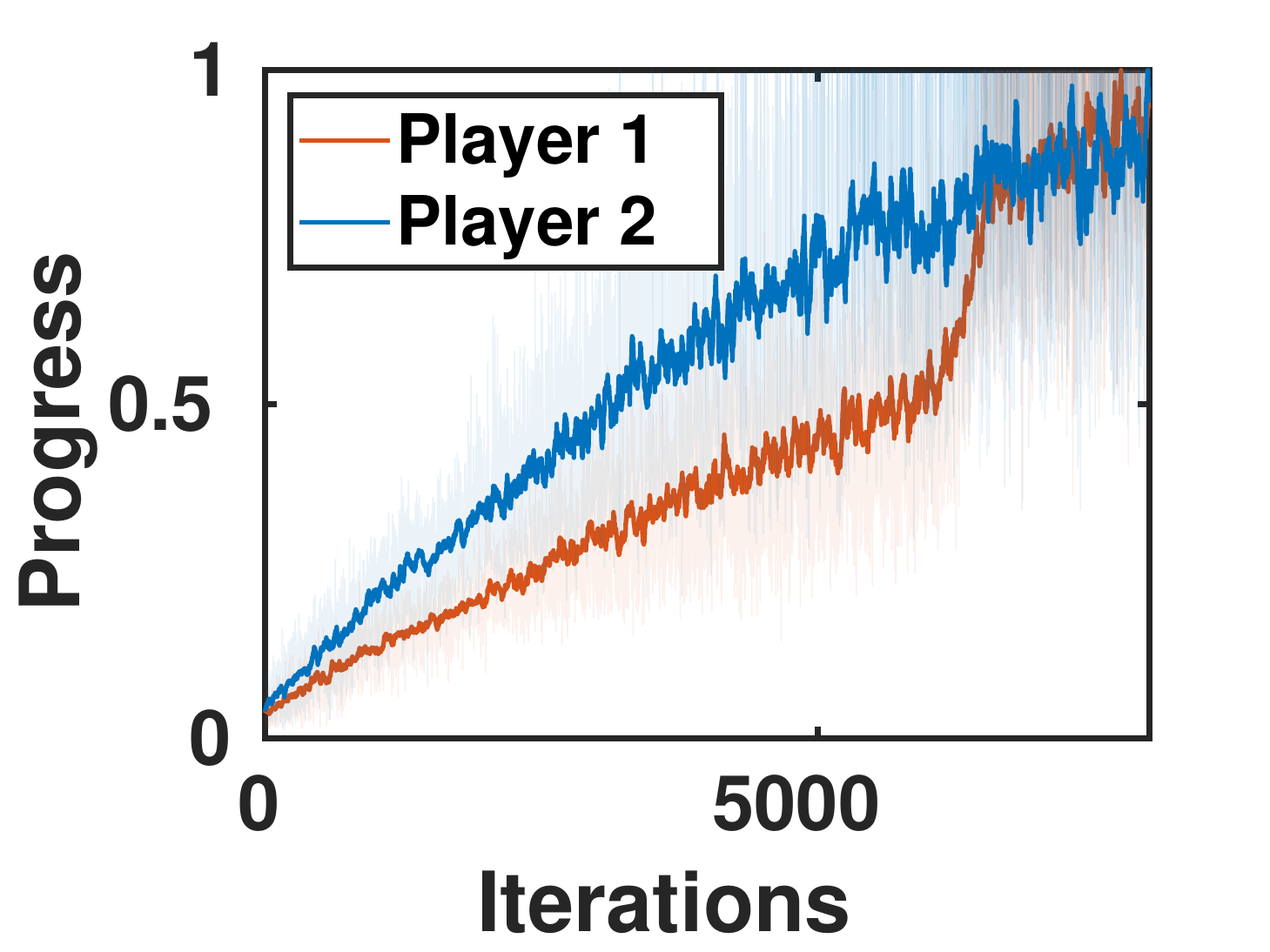}
    \caption{\cpg}
    \label{fig: orca_cpg_progress_plot}
    \end{subfigure}
    ~\hspace{-6.20mm}
    ~
    \begin{subfigure}[t]{0.21\columnwidth}
  	\centering
  	\includegraphics[width=\linewidth]{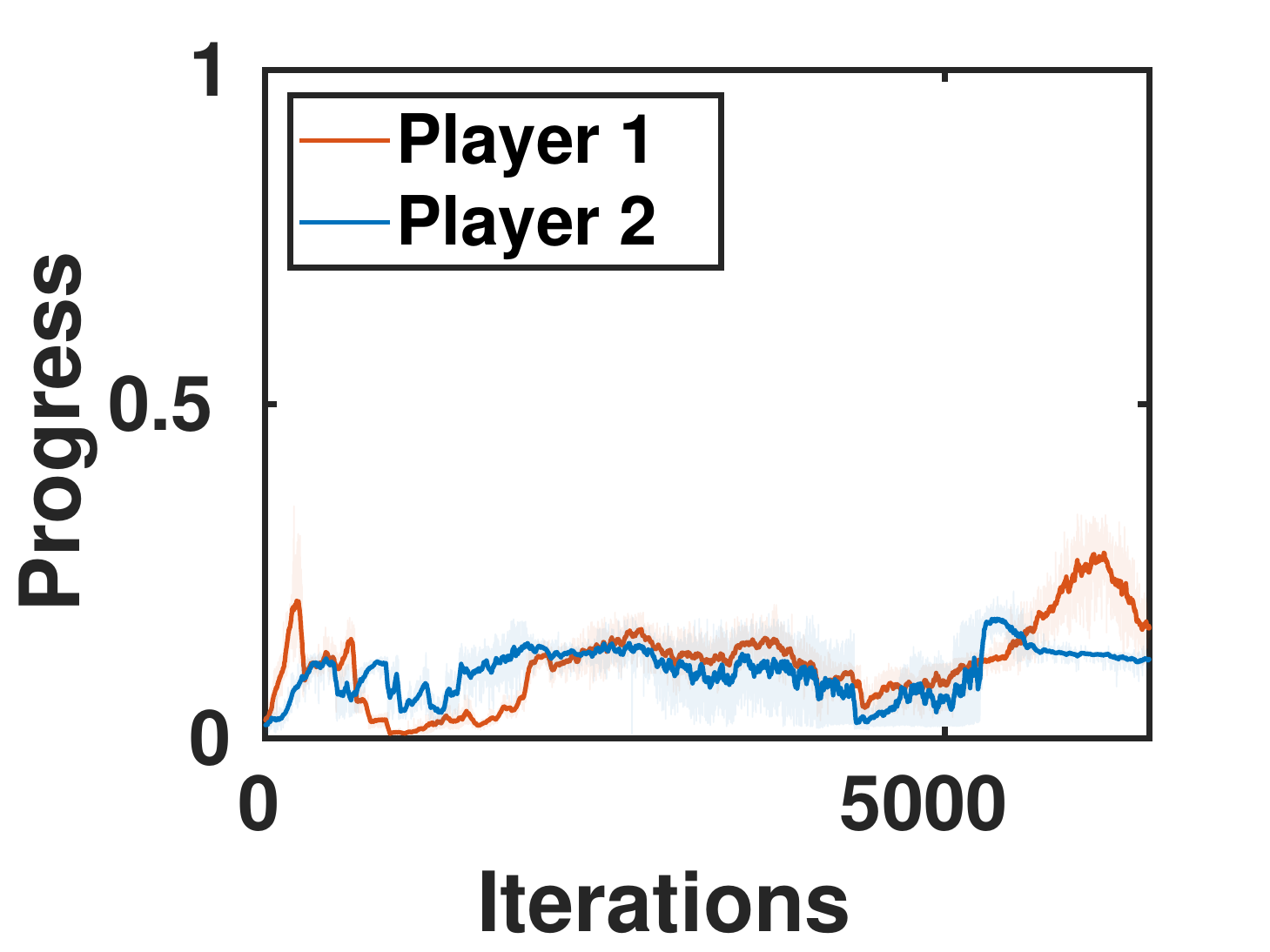}
    \caption{\trpogda}
    \label{fig: trpo2gda_oscillations}
    \end{subfigure}
    ~\hspace{-6.20mm}
    ~
    \begin{subfigure}[t]{0.21\columnwidth}
  	\centering
  	\includegraphics[width=\linewidth]{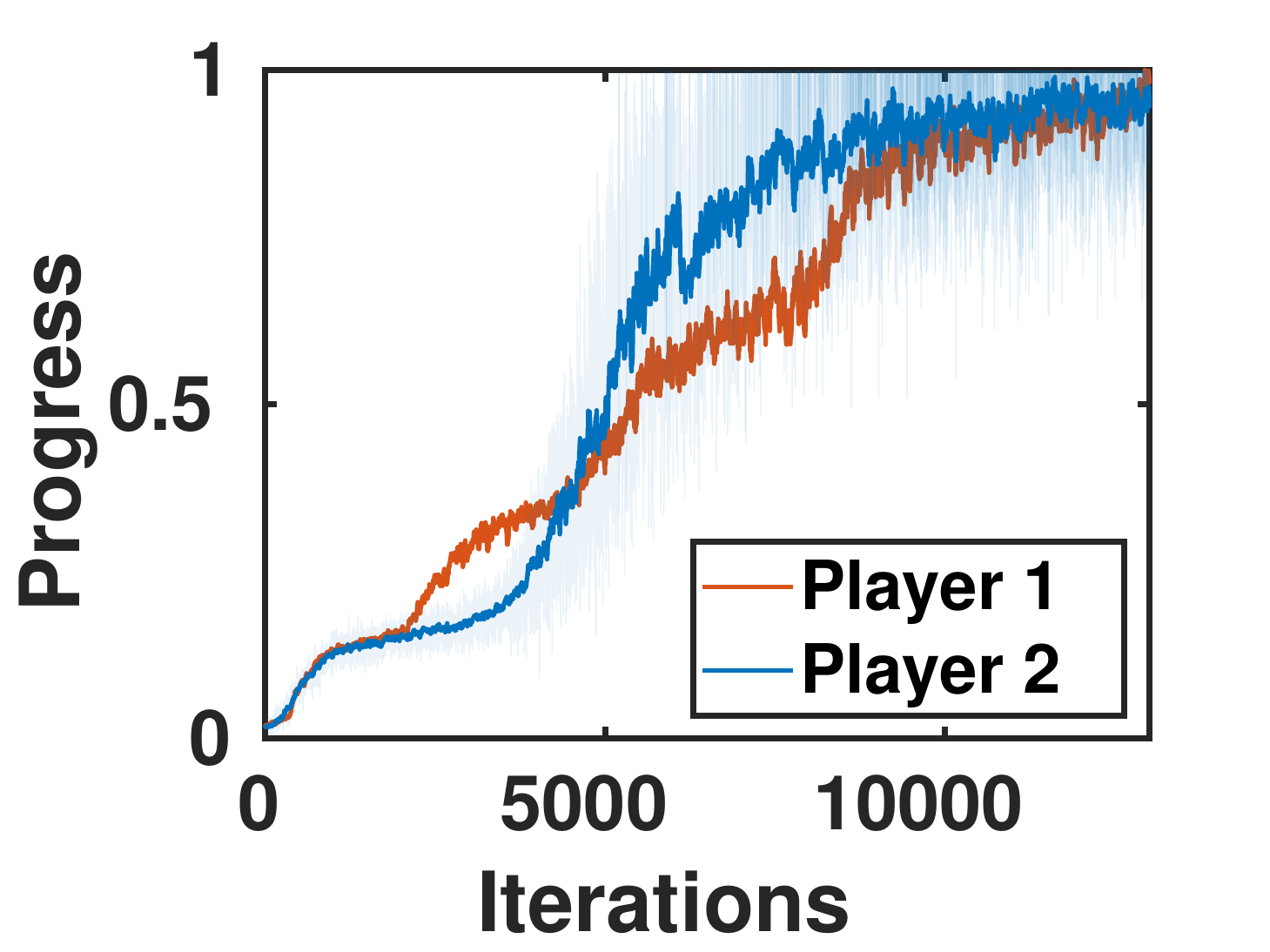}
    \caption{\trcpo}
    \label{fig: progress_plot_orca_trcpo}
    \end{subfigure}
    ~\hspace{-6.20mm}
        ~
    \begin{subfigure}[t]{0.21\columnwidth}
  	\centering
  	\includegraphics[width=\linewidth]{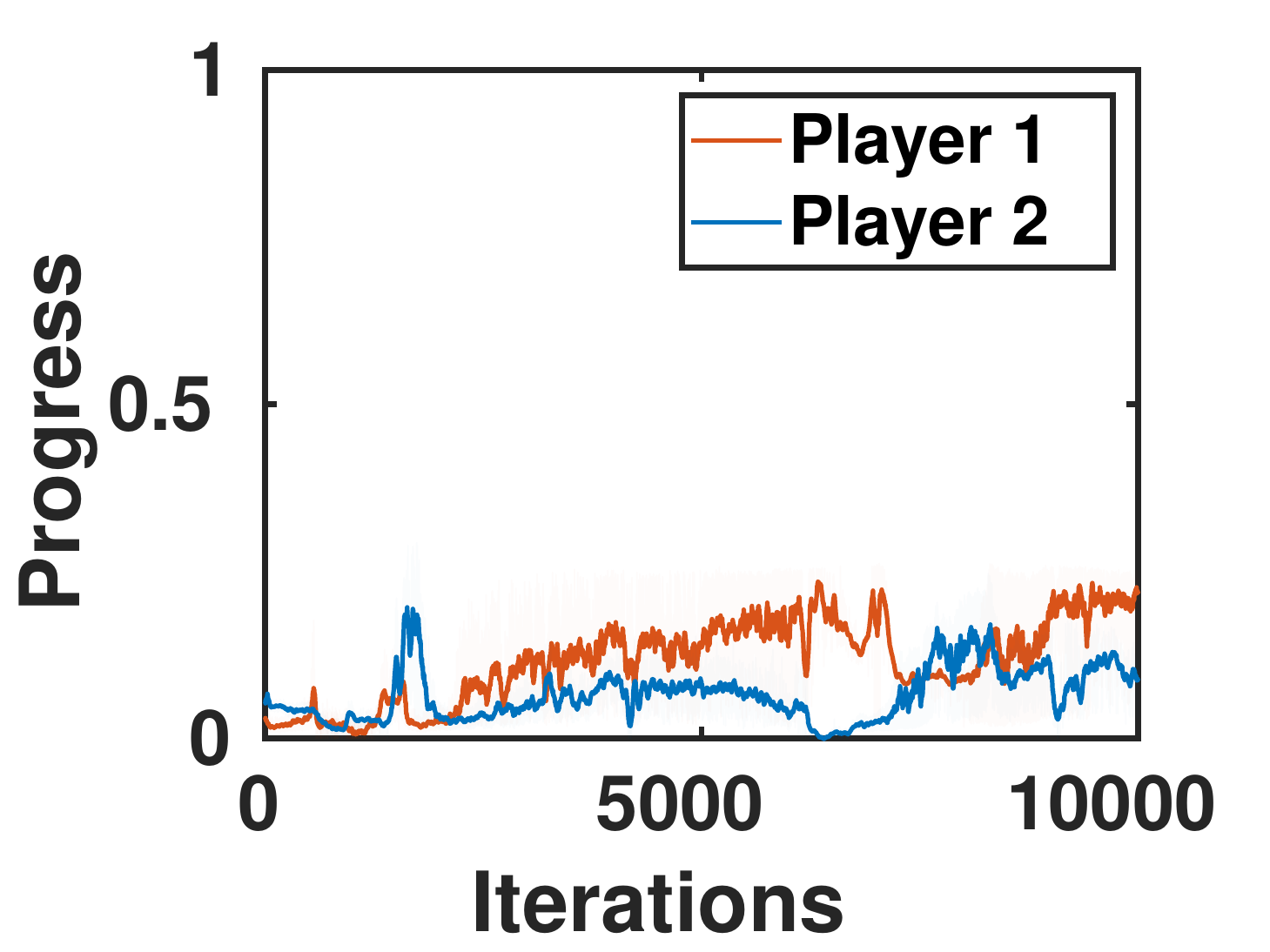}
    \caption{\maddpg}
    \label{fig: progress_plot_orca_maddpg}
    \end{subfigure}
    \vspace{-0.1em}
    ~\hspace{-6.00mm}
\caption{Normalised progress of agents in one lap of car racing game during training}
  \label{fig: ProgressOrca}
\end{figure}
\label{sec: CPG_ORCA}
\textbf{Car Racing} is another interesting game, with continuous state-action space, where two race cars competing against each other to finish the race first~\cite{alex2017noncooperative}. The game is accompanied by two important challenges, 1) learning a policy that can maneuver the car at the limit of handling, 2) strategic interactions with opponents. The track is challenging, consisting of 13 turns with different curvature (\fig~\ref{fig: overtaking_manuever_northsouth}). The game is formulated as a zero-sum, with reward $r(s_k,a^1_{k},a^2_{k}) = \Delta \rho_{car_1} - \Delta \rho_{car_2}$, where $\Delta \rho = \rho_{k+1} - \rho_k$ and $\rho_k$ is the car's progress along the track at the $k^{th}$ time step. If a car crosses track boundaries (e.g., hit the wall), it is penalized, and the opponent receives rewards, this encourages cars to play rough and push each other into the track boundaries. When a collision happens, the rear car is penalized, and the car in the front receives a reward; it promotes blocking by the car in front and overtaking by the car in the rear.

We study agents trained with all \gda, \trpogda, \maddpg, \lola, \cpg and \trcpo in this game, and show that even though all players were able to learn to "drive" only \cpg and \trcpo were able to learn how to "race". Using \gda, only one player was able to learn, which manifested in either one player completely failing and the other finishing the track (\fig~\ref{fig: orca_sgd}), or by an oscillation behavior where one player learns to go ahead, the other agent stays at lower progress(\fig~\ref{fig: trpo2gda_oscillations}, \fig~\ref{fig: progress_plot_orca_maddpg}). Even if one of the players learns to finish one lap at some point, this player does not learn to interact with its opponent(\href{https://youtu.be/rxkGW02GwvE}{https://youtu.be/rxkGW02GwvE}). In contrast, players trained with \cpg and \trpogda, both progress, learn to finish the lap, and race (interact with each other) (See \fig~\ref{fig: orca_cpg_progress_plot} and \fig~\ref{fig: progress_plot_orca_trcpo}). 
%
%
%
%
%
%
To test the policies, we performed races between \cpg and \gda, \trcpo and \trpogda, and \cpg and \trcpo. As shown in \tab~\ref{tab: orca_diff_players}, \cpg and \trcpo win almost all races against their counterparts. 
%
Overall, we see that both \cpg and \trcpo are able to learn policies that are faster, more precise, and interactive with the other player (e.g., learns to overtake).


\begin{table}[h]
\setlength{\tabcolsep}{12pt}
\centering
\caption{Normalized score computed based on 10000 races in the car racing game}
\label{tab: orca_diff_players}
\begin{tabular}{l c c c c c c}
\toprule
 & \multicolumn{2}{c}{\cpg vs \gda} & \multicolumn{2}{c}{\trcpo vs \trpogda} & \multicolumn{2}{c}{\makebox[0pt]{\cpg vs \trcpo}} \\
\midrule
\\\\[-3.9\medskipamount]
 Wins & \textbf{1} & 0 & \textbf{1} & 0 & 0.24 & \textbf{0.76}  \\
 Overtake  & \textbf{1.28} & 0.78 & \textbf{1.28} & 0.78 & 1.80 & \textbf{2.07} \\
 Collision & \textbf{0.17} & 16.11 & \textbf{0.25} & 1.87 & \textbf{0.30} &  0.31\\
\bottomrule
\end{tabular}
\end{table}

\section{Related Work}
\label{sec:related_works}

In tabular \CMDP, Q-learning and actor-critic have been deployed~\citep{littmansoccer, LittmanFoe, LittmanVal,GreenwaldCorrQ,Hu2003,GaruMoyaBalQLearning,QL2Frenay,PerolatActorCritic,SrinivasanActCric}, and recently, deep \RL methods have been extending to \CMDP{}s, with focus of modeling agents behaviour~\citep{tampuu2015multiagent,LeiboMALDQNSocial,RaghuSingleRL}. To mitigate the stabilization issues, centralized methods~\citep{matignon_laurent_le_fort-piat_2012,lowe2017multi,FoersterCOMA}, along with  opponent's behavior modeling~\citep{RobertaOppoModel,HeHeOppo} have been explored. 
Optimization in multi-agent learning can be interpreted as a game in the parameter space, and the main body of the mentioned literature does not take this aspect directly into account since they attempt to separately improve players' performance. Hence, they often fail to achieve desirable performance and oftentimes suffer from unstable training, especially in strategic games~\citep{Hernandez_Leal_2019,Buoniu2010MARL}.
In imperfect information games with known rules, e.g., poker ~\citep{Pocker_MoravcikSBLMBDW17}, a series of works study algorithmically computing Nash equilibra~\citep{shai2006_3107,Koller96,Gilpin2007,ZinkevichNIPS2007_3306, Bowling2017}. Also, studies in stateless episodic games shown convergence to coarse correlated equilibrium~\citep{singleshotCCE, singleshotCCE_RM}. In contrast \cpo converges to the Nash equilibrium in such games. In two-player competitive games, self-play is an approach of interest where a player plays against itself to improve its behavior~\citep{tesauro1995temporal,silver2016mastering}. But, many of these approaches are limited to specific domains~\citep{Heinrich2015,HeinrichS16}.

The closest approach to \cpo in the literature is \lola~\citep{FoersterLOLA} an opponent aware approach. \lola updates parameters using a second-order correction term, resulting in gradient updates corresponding to the following shortened recursion: if a player thinks that the other player thinks its strategy stays constant~\citep{schfer2019competitive}, whereas \cpo recovers the full recursion series until the Nash equilibrium is delivered. In contrast to ~\citep{FoersterLOLA} we also provide \cpo extension to value-based, and trust region-based methods, along with their efficient implementation. 

Our work is also related to GANs ~\citep{goodfellow2014generative}, which involves solving a zero sum two-player competitive game (\CMDP with single state). Recent development in nonconvex-nonconcave problems and GANs training show, simple and intuitive, \gda has undesirable convergence properties ~\citep{mazumdar2019policygradient}  
and exhibit strong rotation around fixed points ~\citep{BalduzziHamiltonian}. To overcome this rotation behaviour of \gda, various modifications have been proposed, including averaging ~\citep{yazc2018unusual}, negative momentum ~\citep{NegativeMomentumGidel2018} along many others ~\citep{EG_mertiko2018,OGDA_daskalakis2017,ConOpti_Mescheder2017,BalduzziHamiltonian,Gempmahadevan2015}.
Considering the game-theoretic nature of this problem, competitive gradient descent has been proposed as a natural generalization of gradient descent in two-players instead of \gda for GANs~\citep{schfer2019competitive}. This method, as the predecessor to \cpo, enjoys stability in training, robustness in choice of hyper-parameters, and has desirable performance and convergence properties. This method is also a special case of \cpo when the \CMDP of interest has only one state.

\section{Conclusion}
\label{sec:conclusion}

We presented competitive policy optimization \cpo, a novel \PG-based \RL method for two player strictly competitive game. In \cpo, each player optimizes strategy by considering the interaction with the environment and the opponent through game theoretic bilinear approximation to the game objective. This method is instantiated to competitive policy gradient (\cpg) and trust region competitive policy optimisation (\trcpo) using value based and trust region approaches. We theoretically studied these methods and provided \PG theorems to show the properties of these model-free \RL approaches. We provided efficient implementation of these methods and empirically showed that they provide stable and faster optimization, and also converge to more sophisticated and competitive strategies. 
We dedicated this paper to two player zero-sum games, however, the principles provided in this paper can be used for multi-player general games. In the future, we plan to extend this study to multi-player general-sum games along with efficient implementation of methods. Moreover, we plan to use the techniques proposed in partially observable domains, and study imperfect information games~\citep{azizzadenesheli2020policy}. Also, we aim to explore opponent modelling approaches to relax the requirement of knowing the opponent’s parameters during optimization.

\newpage
\section*{Acknowledgements}
The main body of this work took place when M. Prajapat was a visiting scholar at Caltech. The authors would like to thank Florian Sch\"afer for his support. M. Prajapat is thankful to Zeno Karl Schindler foundation for providing him with a Master thesis grant. K. Azizzadenesheli is supported in part by Raytheon and Amazon Web Service. A. Anandkumar is supported in part by Bren endowed chair, DARPA PAIHR00111890035 and LwLL grants, Raytheon, Microsoft, Google, and Adobe faculty fellowships.
\bibliography{ref}
\bibliographystyle{plainnat}

\newpage
\appendix

\allowdisplaybreaks
\begin{center}

{\huge Appendix}
\end{center}
\label{sec:appendix}
\section{Algorithms}
In this section, we briefly present the algorithms discussed in the paper, namely gradient descent ascent (\gda), competitive policy gradient (\cpg), trust region gradient descent ascent (\trpogda) and trust region competitive policy optimization (\trcpo). 
\begin{algorithm}[!ht]
\SetAlgoLined
\textbf{Initialize}  $(\theta^1,\theta^2)$\\ 
\For{$epoch:1,2,3,..$ until termination}{
Collect samples under $\pi(.|.;\theta^1)$, $\pi(.|.;\theta^2)$,\\
Estimate $Q$, then $D_{\theta^i} \eta$ in \eq\ref{eq:CPGGradHessianVal1}\\
Update $\theta^1,\theta^2$ using \eq~\ref{eq:GDAlinearised_game}\\
}
\caption{Gradient Descent Ascent Policy Gradient}
\label{alg:gda_short}
\end{algorithm}
\vspace{-1.00mm}
\vspace{-1.00mm}

\begin{algorithm}[!ht]
\SetAlgoLined
\textbf{Initialize}  $(\theta^1,\theta^2)$\\ 
\For{$epoch:1,2,3,..$ until termination}{
Collect samples under $\pi(.|.;\theta^1)$, $\pi(.|.;\theta^2)$,\\
Estimate $Q$, then $D_{\theta^i} \eta, D_{\theta^i \theta^j} \eta$ in Eqs.~\ref{eq:CPGGradHessianVal1},\ref{eq:CPGGradHessianVal}\\
Update $\theta^1,\theta^2$ using \eq~\ref{eqn: CGDSolution}\\
}
\caption{Competitive Policy Gradient}
\label{alg:cpg_short}
\end{algorithm}
\vspace{-1.00mm}
\vspace{-1.00mm}
\begin{algorithm}[!ht]
\SetAlgoLined
\textbf{Initialize}  $(\theta^1,\theta^2)$\\ 
\For{$epoch:1,2,3,..$ until termination}{
Collect samples under $\pi(.|.;\theta^1)$, $\pi(.|.;\theta^2)$,\\
Estimate $A$, and $D_{\theta^i} L$ using \eq~\ref{eqn: GradHessianTRCPO},\\
Update $\theta^i$ using \eq~\ref{eqn: trcpo_game}, with bilinear term as a null matrix. \\
}
\caption{Trust Region Gradient Descent Ascent}
\label{alg: trgda_short}
\end{algorithm}
\vspace{-1.00mm}
\vspace{-1.00mm}

\begin{algorithm}[!ht]
\SetAlgoLined
\textbf{Initialize}  $(\theta^1,\theta^2)$\\ 
\For{$epoch:1,2,3,..$ until termination}{
Collect samples under $\pi(.|.;\theta^1)$, $\pi(.|.;\theta^2)$,\\
Estimate $A$, and $D_{\theta^i} L, D_{\theta^i \theta^j} L$ using \eq~\ref{eqn: GradHessianTRCPO},\\
Update $\theta^1,\theta^2$ using \eq~\ref{eqn: trcpo_game}, \\
}
\caption{Trust Region Competitive Policy Optimisation}
\label{alg: trcpo_short}
\end{algorithm}

\label{appx: AlgoCPG}
\section{Recursion reasoning}\label{sec:Neuman}
\begin{figure}[ht]
  	\centering
  	\includegraphics[width=0.6\linewidth]{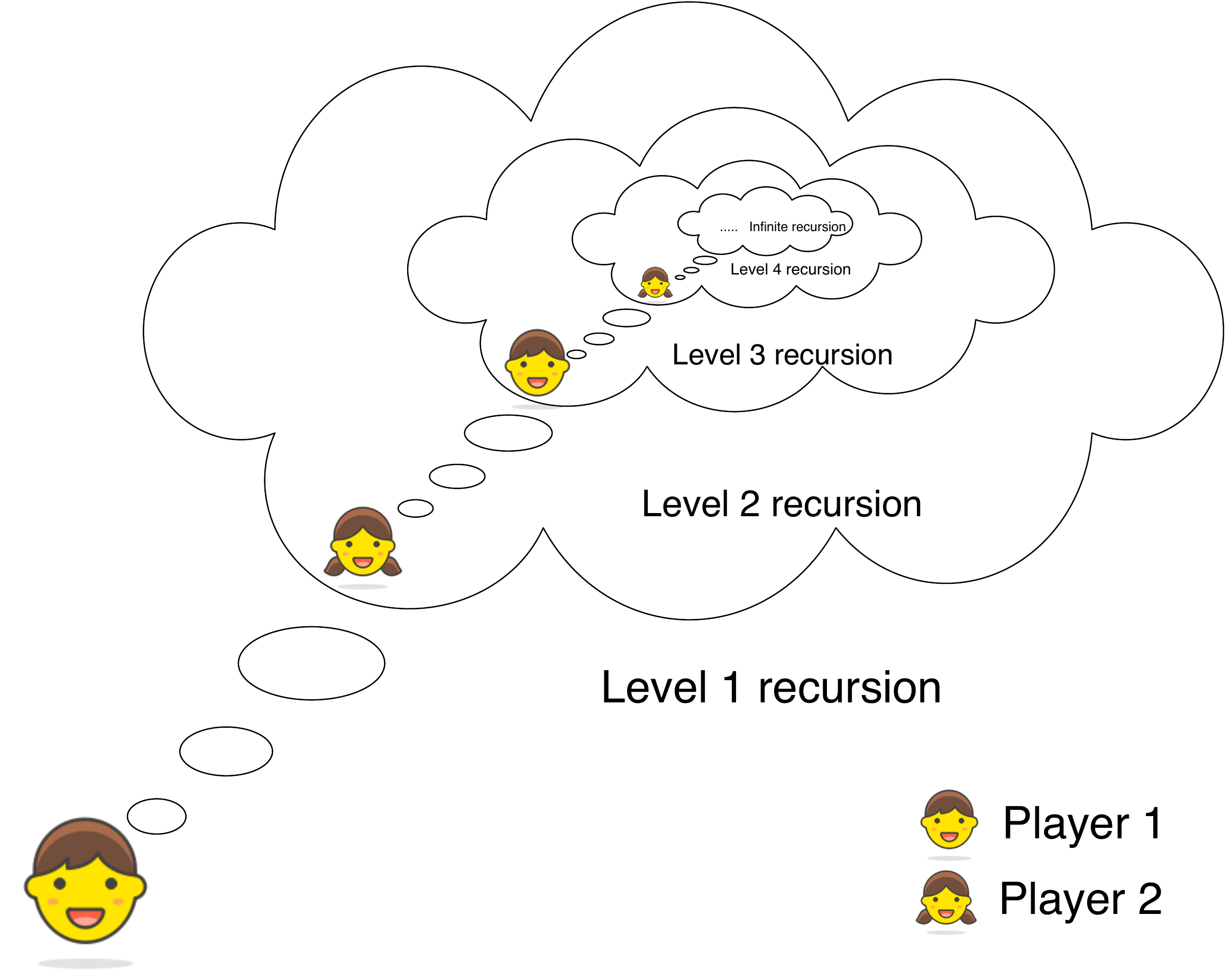}
    \caption{Recursion reasoning. The notion of "\textit{Level N recursion}" in the figure assumes that there is no further recursive thinking (cloud) beyond that level.}
    \label{fig: recursion}
\end{figure}

When players play strategic games, in order to maximize their payoffs, they need to consider their opponents' strategies which in turn may depend on their strategies (and so on), resulting in the well known infinite recursion in game theory. This is the recursive reasoning of the form what do I think that you think that I think (and so on) and arises while acting rationally in multi-player games. In this section we elaborate on the recursion in \CMDP and latter show how \cpo recovers the infinite recursion reasoning \fig~\ref{fig: recursion}. Recursion reasoning in \CMDP works as follows:
\begin{itemize}
    \item Level 1 recursion: Each player considers how the environment, as the results of the agents’ current and future moves temporally evolves in favor of each agent, 
    \item Level 2 recursion: Considering that the other agent also considers how the environment, as the results of the agents’ current and future moves temporally evolves in favor of each agent, 
    \item Level 3 recursion: Considering that the other agent also considers that the other agent also considers how the environment, as the results of the agents’ current and future moves temporally evolves in favor of each agent, 
    \item .......
\end{itemize}

In the special case of single state games (non sequential setting), this recursive reasoning boils down to what each player thinks, the other player thinks that the other player thinks, .... . ( For readers interested in the recursion reasoning in non-sequential but repeated episodic competitive games (single state (or stateless) episodic \CMDP), stability and convergence of competitive gradient optimization and the derivation of Eq.~\ref{eqn: CGDSolution}, we refer to~\citep{schfer2019competitive}). 

In the following we provide an intuitive explanation for the updates prescribed by \cpo.
Let us restate the Nash equilibrium of the bilinear game in \eq~\ref{eq:CGDBilinearised_game} in its matrix form,
\begin{align}\label{eq: MatrixCGDUpdate}
    \begin{bmatrix}
\theta^1 \\
 \theta^2 \\
\end{bmatrix} \leftarrow \begin{bmatrix}
\theta^1 \\
 \theta^2 \\
\end{bmatrix} + \alpha \begin{bmatrix}
I & - \alpha D_{\theta^1\theta^2} \eta\\
 \alpha D_{\theta^2\theta^1} \eta & I\\
\end{bmatrix}^{-1} \begin{bmatrix}
D_{{\theta}^1} \eta \\
- D_{{\theta}^2} \eta \\
\end{bmatrix}.
\end{align}{}
We rewrite the above mentioned update as 
\begin{align}\label{eq: MatrixCGDUpdateAlex}
    \begin{bmatrix}
\theta^1 \\
 \theta^2 \\
\end{bmatrix} \leftarrow \begin{bmatrix}
\theta^1 \\
 \theta^2 \\
\end{bmatrix} + \alpha (I -A)^{-1} \begin{bmatrix}
D_{{\theta}^1} \eta \\
- D_{{\theta}^2} \eta \\
\end{bmatrix}, \quad \text{with } \quad A  = \begin{bmatrix}
0 & \alpha D_{\theta^1\theta^2} \eta\\
 -\alpha D_{\theta^2\theta^1} \eta & 0\\
\end{bmatrix}.
\end{align}{}
If it holds that the spectral radius of $A$ is smaller than one, then we can use the Neumann series to compute the inverse,
\begin{align}\label{eq: series}
(I - A)^{-1} = \lim_{N\rightarrow\infty}\sum_{k=0}^N A^k = I + A + A^2 + A^3 +... \,.
\end{align}
Using this expansion and its different orders as an approximation, e.g., up to $k\in\lbrace 0,\ldots , N\rbrace$ for a finite $N$,  we derive policy update rules that have different levels of reasoning. For example, consider the following levels of recursion reasoning:
\begin{itemize}
\item N = 0, $(I - A)^{-1} \rightarrow I$, i.e., we approximate $(I - A)^{-1}$ with the first term in the series presented in \eq~\ref{eq: series}, therefore, the update rule in \eq~\ref{eq: MatrixCGDUpdate} reduces to the \gda update rule. In this case, each agent updates its parameters considering its current and future moves to gain a higher total cumulative reward in the environment assuming the policy of the opponent  stays unchanged,
    \begin{align}
    \begin{bmatrix}
\theta^1 \\
 \theta^2 \\
\end{bmatrix} \leftarrow \begin{bmatrix}
\theta^1 \\
 \theta^2 \\
\end{bmatrix} + \alpha \begin{bmatrix}
D_{{\theta}^1} \eta \\
- D_{{\theta}^2} \eta \\
\end{bmatrix}.\nonumber
\end{align}{}
    \item N = 1, $(I - A)^{-1} \rightarrow I + A$ and the update rule in \eq~\ref{eq: MatrixCGDUpdate} reduce to the \lola update rule. In this case, each agent updates its parameters considering that the opponent updates its parameters considering the other player stays unchanged. For example, player $1$ updates its parameter, assuming that player $2$ also updates its parameter, but player $1$ assumes that the player $2$ updates its parameter assuming the player $1$ stays unchanged.
\begin{align}
\begin{bmatrix}
\theta^1 \\
 \theta^2 \\
\end{bmatrix} \leftarrow \begin{bmatrix}
\theta^1 \\
 \theta^2 \\
\end{bmatrix} + \alpha \begin{bmatrix}
D_{{\theta}^1} \eta - \alpha D_{\theta^1\theta^2} \eta D_{\theta^2} \eta\\
- D_{{\theta}^2} \eta - \alpha D_{\theta^2\theta^1} \eta D_{\theta^1} \eta\\
\end{bmatrix}.\nonumber
\end{align}{}
    \item N = 2, $(I - A)^{-1} \rightarrow I + A + A^2$, and the update rule is given by \eq~\ref{eq: CPG_N2_approx}. In this case, the agent updates its parameters considering that the opponent also considering that the agent is considering to update parameters, such that its current and future moves increase the probability of earning a higher total cumulative reward in the environment.
\begin{align} \label{eq: CPG_N2_approx}
\begin{bmatrix}
\theta^1 \\
 \theta^2 \\
\end{bmatrix} \leftarrow \begin{bmatrix}
\theta^1 \\
 \theta^2 \\
\end{bmatrix} + \alpha \begin{bmatrix}
D_{{\theta}^1} \eta - \alpha D_{\theta^1\theta^2} \eta D_{\theta^2} \eta - \alpha^2 D_{\theta^1\theta^2} \eta D_{\theta^2\theta^1} \eta D_{\theta^1} \eta \\
- D_{{\theta}^2} \eta - \alpha D_{\theta^2\theta^1} \eta D_{\theta^1} \eta + \alpha^2 D_{\theta^2\theta^1} \eta D_{\theta^1\theta^2} \eta D_{\theta^2} \eta \\
\end{bmatrix}.
\end{align}{}
    \item $\lim$ N $\rightarrow$ $\infty$, $(I - A)^{-1} \rightarrow I + A + A^2 + A^3 + ...$, we recover the Nash equilibrium in the limit which corresponds to infinite recursion reasoning. In this case, the update rule in \eq~\ref{eq: MatrixCGDUpdate} is given by \eq~\ref{eq: CPG_Ninf_approx} which resembles the \eq~\ref{eqn: CGDSolution}.
    \begin{align} \label{eq: CPG_Ninf_approx}
 \begin{bmatrix}
 \theta^1\\
 \theta^2
\end{bmatrix} 
\leftarrow
 \begin{bmatrix}
 \theta^1\\
 \theta^2
\end{bmatrix} +
\alpha \begin{bmatrix}
\bigl( Id + \alpha^2 D_{\theta^1\theta^2} \eta D_{\theta^2\theta^1}\eta \bigr)^{-1} \bigl( D_{\theta^1} \eta - \alpha D_{\theta^1\theta^2}\eta D_{\theta^2}\eta \bigr)\\
- \bigl( Id + \alpha^2 D_{\theta^2\theta^1}\eta D_{\theta^1\theta^2}\eta \bigr)^{-1} \bigl( D_{\theta^2}\eta + \alpha D_{\theta^2\theta^1}\eta D_{\theta^1} \eta \bigr)\\
\end{bmatrix}.
\end{align}
\end{itemize}

We provided this discussion to motivate the importance of taking the game theoretic nature of a given problem into account, in order to design an update.


\section{Proof of competitive policy gradient theorem}
In this section, we first derive the gradient and bilinear terms in the form of score function. In the subsequent subsections we provide complete proof of the competitive policy gradient theorem.

\subsection{Proof for \prop~\ref{prop: CPOGradHessianTrajDist}:}
\label{appx: CPOGradHessianTrajDist}

\begin{proof}
Given that the agents' policies are parameterized with $\theta^1$ and $\theta^2$, we know that the game objective is,
\begin{align}
\eta(\theta^1,\theta^2) &= \int_{\tau} f(\tau;\theta^1,\theta^2) R(\tau) d\tau, 
\shortintertext{thus, the gradient with respect to $\theta_i$ is given by}
D_{\theta^i}\eta &= \int_{\tau} D_{\theta_i} f(\tau;\theta^1,\theta^2) R(\tau) d\tau, \; i\in\!\lbrace 1,2\rbrace\,,
\shortintertext{for any $p_{\theta}(\tau) \neq 0$ using $D_{\theta} \log p_{\theta}(\tau) = \frac{D_{\theta} p_{\theta}(\tau)}{p_{\theta}(\tau)}$ from standard calculus and the definition of $f(\tau;\theta^1,\theta^2)$ in \eq~\ref{eq: trajectory_distribution}, we can compute the gradient of the game objective,}
D_{\theta^i}\eta &= \int_{\tau} f(\tau;\theta^1,\theta^2) (D_{\theta^i}\log(\prod_{k=0}^{|\tau|-1} \pi(a_k|s_k;\theta^i))) R(\tau) d\tau\,, \nonumber\\
\shortintertext{Let us define the score function as  $g(\tau, \theta^i) = D_{\theta^i}(\log(\prod_{k=0}^{|\tau|-1} \pi(a_{k}|s_k;\theta^i)))$ which results in, }
D_{\theta^i}\eta &= \int_{\tau} f(\tau;\theta^1,\theta^2) g(\tau, \theta^i) R(\tau) d\tau .
\shortintertext{
 By definition we know that the second order bilinear approximation of the game objective is $D_{\theta^2\theta^1} \eta(\theta^1,\theta^2) = D_{\theta^2} D_{\theta^1}\eta(\theta^1,\theta^2)$. Hence,}
D_{\theta^2\theta^1} \eta &= \int_{\tau} D_{\theta^2}f(\tau;\theta^1,\theta^2) g(\tau, \theta^1)^\top R(\tau) d\tau\,. \nonumber\\
&= \int_{\tau} f(\tau;\theta^1,\theta^2) (D_{\theta^2}\log(\prod_{k=0}^{|\tau|-1} \pi(a_k|s_k;\theta^2))) g(\tau, \theta^1)^\top R(\tau) d\tau\,, \nonumber\\
&= \int_{\tau} f(\tau;\theta^1,\theta^2) g(\tau, \theta^2) g(\tau, \theta^1)^\top R(\tau) d\tau \, . \nonumber
\end{align}
which concludes the proof.
\end{proof}

\subsection{Proof for \thm~\ref{thm: CPGGradHessianVal}}
\label{appx: CPGGradHessianVal}
The competitive policy gradient theorem (\thm~\ref{thm: CPGGradHessianVal}) consists of two parts. We first provide the derivation for the gradient and utilise this result to derive the bilinear term in the subsequent subsection.   
\subsubsection{Derivation of gradient term \texorpdfstring{$D_{\theta^1} \eta$}{Lg}}
\begin{proof}
Using \eq~\ref{eqn: val_rew}, which defines $V(s_k;\theta^1, \theta^2)$ and $Q(s_k,a^1_{k},a^2_{k};\theta^1, \theta^2)$. We can write $V(s_0;\theta^1, \theta^2)$ as,
\begin{align}\label{eqn: appxValIntegral}
V(s_0;\theta^1,\theta^2) &= \!\int_{a_0^1} \! \int_{a_0^2} \!\!\!\pi(a_0^1|s_0;\theta^1) \pi(a_0^2|s_0;\theta^2) Q(s_0,a_0^1,a_0^2;\theta^1\!\!,\theta^2) da_0^1 da_0^2\,, \\
\text{where, } \, Q(s_0,a_0^1,a_0^2;\theta^1\!\!,\theta^2) &= r(s_0,a_0^1,a_0^2) + \gamma \int_{s_1} T(s_1|s_0,a_0^1,a_0^2) V(s_0;\theta^1,\theta^2) ds_1\,. \nonumber
\shortintertext{Hence, the gradient of the state-value $V$ is,}
\label{eqn: gradv}
D_{\theta^1} V(s_0;\theta^1,\theta^2) &= \int_{a_0^1} \int_{a_0^2} \frac{\partial \pi(a_0^1|s_0;\theta^1)}{\partial \theta^1} \pi(a_0^2|s_0;\theta^2) Q(s_0,a_0^1,a_0^2;\theta^1\!\!,\theta^2) da_0^1  da_0^2 \nonumber\\
& + \!\!\int_{a_0^1} \! \int_{a_0^2} \!\! \pi(a_0^1|s_0;\theta^1)  \pi(a_0^2|s_0;\theta^2) \! \frac{\partial Q(s_0,a_0^1,a_0^2;\theta^1\!\!,\theta^2)} {\partial \theta^1}  da_0^1  da_0^2 .
\shortintertext{The gradient of the state-action value $Q$, in term of the gradient of the state-value $V$ is given by}
\label{eqn: gradq}
D_{\theta^1} Q(s_0,a_0^1,a_0^2;\theta^1\!\!,\theta^2) &= \gamma \int_{s_1} T(s_1|s_0,a_0^1,a_0^2) \frac{\partial V(s_1;\theta^1,\theta^2)}{\partial \theta^1} ds_1 \numberthis \,.
\end{align}
Substituting \eq~\ref{eqn: gradq} in \eq~\ref{eqn: gradv} results in,
\begin{align}\label{eqn: recursive_gradv}
\!D_{\theta^1} \!V(\!s_0;\theta^1,\theta^2) & = \!  \int_{a_0^1} \!\! \int_{a_0^2} \!\!\!\!\frac{\partial \pi(\!a_0^1|s_0;\theta^1)}{\partial \theta^1} \pi(\!a_0^2|s_0;\theta^2) Q(\!s_0,a_0^1,a_0^2;\theta^1\!\!,\theta^2) da_0^1 da_0^2 \\
 &+ \! \int_{a_0^1} \!\! \int_{a_0^2} \!\!\!\! \pi(a_0^1|s_0;\theta^1)  \pi(a_0^2|s_0;\theta^2)   \gamma \!\!\int_{s_1} \!\!\!\!T(s_1|s_0,a_0^1,a_0^2)   \frac{\partial V(s_1;\theta^1,\theta^2)}{\partial \theta^1}  ds_1  da_0^1  da_0^2\,. \nonumber
\shortintertext{By exploring the recursive nature of the state-value $V$ and $D_{\theta^1} V(s_0;\theta^1,\theta^2)$, we can unroll \eq~\ref{eqn: recursive_gradv} as,}
 \!D_{\theta^1}  \!V( s_0;\theta^1,\theta^2)& = \int_{a_0^1} \int_{a_0^2} \frac{\partial \pi(a_0^1|s_0;\theta^1)}{\partial \theta^1}\pi(a_0^2|s_0;\theta^2) Q(s_0,a_0^1,a_0^2;\theta^1,\theta^2) da_0^1 da_0^2 \label{eqn: recursive_gradv2}\\
 &+\int_{a_0^1}  \int_{a_0^2}\pi(a_0^1|s_0;\theta^1) \pi(a_0^2|s_0;\theta^2) \gamma \int_{s_1} T(s_1|s_0,a_0^1,a_0^2) \nonumber\\
 &\quad \quad \Biggl(  \int_{a_1^1} \int_{a_1^2}\frac{\partial \pi(a_1^1|s_1;\theta^1)}{\partial \theta^1}  \pi(a_1^2|s_1;\theta^2) Q(s_1,a_1^1,a_1^2;\theta^1,\theta^2) da_1^1 da_1^2 \nonumber\\
 &\quad \quad \quad \quad \quad  +\int_{a_1^1}  \int_{a_1^2}   \pi(a_1^1|s_1;\theta^1)  \pi(a_1^2|s_1;\theta^2) \gamma \int_{s_2}  T(s_2|s_1,a_1^1,a_1^2) \nonumber\\
 & \ \ \quad\quad\quad \quad \quad\quad\quad\quad\quad \quad\quad\quad \frac{\partial V(s_2;\theta^1,\theta^2)}{\partial \theta^1} ds_2 da_1^1 da_1^2 \Biggr) ds_1  da_0^1 da_0^2 \nonumber \, .
\shortintertext{To find the derivative of the expected return, we can use the definition of the expected return $\eta (\theta^1,\theta^2)$ and \eq~\ref{eqn: recursive_gradv2}, which results in}
\label{eqn: recursive_grad2_traj}
\!\!\!\!\!\!\!\!\!\!D_{\theta^1}\eta(\theta^1,\theta^2) & = \!\int_{s_0} p(s_0) D_{\theta^1} V(s_0;\theta^1,\theta^2) ds_0 \nonumber\\
= \!& \int_{s_0} p(s_0) \int_{a_0^1} \int_{a_0^2} \frac{\partial \pi(a_0^1|s_0;\theta^1)}{\partial \theta^1}  \pi(a_0^2|s_0;\theta^2) Q(s_0,a_0^1,a_0^2;\theta^1\!\!,\theta^2) da_0^1 da_0^2 ds_0 \nonumber\\
 & + \!\gamma \!\! \int_{\tau} \!\! f(\tau_{0:0},s_1;\theta^1\!\!,\theta^2) 
\frac{\partial \! \pi(a_1^1|s_1;\theta^1)}{\partial \theta^1}  \pi(a_1^2|s_1;\theta^2) Q(s_1,a_1^1,a_1^2;\theta^1\!\!,\theta^2) d\tau_{0:1} \nonumber\\
 &+ \gamma^2 \int_{\tau} f(\tau_{0:1},s_2;\theta^1,\theta^2) \frac{\partial V(s_2;\theta^1,\theta^2)}{\partial \theta^1} ds_2 d\tau_{0:1}.
\shortintertext{
Similar to the two previous step, unrolling and marginalisation can be continued for the entire length of the trajectory, which results in,}
\!\!\!\!\!\!\!\!\!\!D_{\theta^1} \eta(\theta^1,\theta^2) = \! \int_{\tau}& \!\sum_{k=0}^{|\tau|-1} \!\!\! \gamma^k \! f(\!\tau_{0:k-1}\!,\!s_k\!;\theta^1\!\!,\theta^2) \!\!\! \int_{a_k^1} \!\! \int_{a_k^2} \!\!\!\!\!\frac{\partial \pi(\!a_k^1|s_k;\!\theta^1\!)}{\partial \theta^1} \!\pi(\!a_k^2|s_k;\!\theta^2) Q(\!s_k,\!a_k^1,\!a_k^2;\!\theta^1\!\!,\theta^2) d\tau \\
= \int_{\tau} &\sum_{k=0}^{|\tau|-1} \!\!\gamma^k \! f(\tau_{0:k};\theta^1,\theta^2) \frac{\partial \log \pi(a_k^1|s_k;\theta^1)}{\partial \theta^1} Q(s_k,a_k^1,a_k^2;\theta^1\!\!,\theta^2)  d\tau.
\end{align}
This concludes our derivation of the gradient term of the game objective for the two agent case.
\end{proof}
\subsubsection{Derivation of Bilinear term  \texorpdfstring{$D_{\theta^1 \theta^2} \eta$}{Lg}}

\textit{Proof:} By definition we know that $D_{\theta^1 \theta^2} V(s_0;\theta^1,\theta^2) = D_{\theta^1}(D_{\theta^2} V(s_0;\theta^1,\theta^2))$. Thus we can use $D_{\theta^2} V(s_0;\theta^1,\theta^2)$ as defined in \eq~\ref{eqn: gradv} and differentiating with respect to $\theta^1$, thereby we get,
\begin{align}
D_{\theta^1 \theta^2} V(s_0;\theta^1,\theta^2) &= \int_{a_0^1} \int_{a_0^2} \frac{\partial \pi(a_0^1|s_0;\theta^1)}{\partial \theta^1}  \frac{\partial \pi(a_0^2|s_0;\theta^2)}{\partial \theta^2}^\top Q(s_0,a_0^1,a_0^2;\theta^1\!\!,\theta^2) da_0^1 da_0^2 \nonumber\\
&+ \int_{a_0^1} \int_{a_0^2} \frac{\partial \pi(a_0^1|s_0;\theta^1)}{\partial \theta^1}  \pi(a_0^2|s_0;\theta^2) \frac{\partial Q(s_0,a_0^1,a_0^2;\theta^1\!\!,\theta^2)}{\partial \theta^2}^\top da_0^1 da_0^2 \nonumber\\
&+ \int_{a_0^1} \int_{a_0^2}  \pi(a_0^1|s_0;\theta^1)  \frac{\partial Q(s_0,a_0^1,a_0^2;\theta^1\!\!,\theta^2)} {\partial \theta^1} \frac{\partial \pi(a_0^2|s_0;\theta^2)}{\partial \theta^2}^\top da_0^1 da_0^2 \nonumber\\
&+ \int_{a_0^1} \int_{a_0^2}  \pi(a_0^1|s_0;\theta^1)  \pi(a_0^2|s_0;\theta^2) \frac{\partial^2 Q(s_0,a_0^1,a_0^2;\theta^1\!\!,\theta^2)} {\partial \theta^1 \partial \theta^2} da_0^1 da_0^2. \label{eqn: doublegradv}
\end{align}
Exploiting the recursive structure of $D_{\theta^1 \theta^2} V(s_0;\theta^1,\theta^2)$, we can evaluate $D_{\theta^1 \theta^2} V(s_1;\theta^1,\theta^2)$ and substitute it back in \eq~\ref{eqn: doublegradv},we get,
\begin{align}
\label{eqn: subs_double_gradv}
D_{\theta^1 \theta^2} V(s_0;\theta^1,\theta^2) =& \int_{a_0^1} \int_{a_0^2} \frac{\partial \pi(a_0^1|s_0;\theta^1)}{\partial \theta^1} \frac{\partial \pi(a_0^2|s_0;\theta^2)}{\partial \theta^2}^\top Q(s_0,a_0^1,a_0^2;\theta^1\!\!,\theta^2) da_0^1 da_0^2 \\
& + \int_{a_0^1} \int_{a_0^2} \frac{\partial \pi(a_0^1|s_0;\theta^1)}{\partial \theta^1}  \pi(a_0^2|s_0;\theta^2) \frac{\partial Q(s_0,a_0^1,a_0^2;\theta^1\!\!,\theta^2)}{\partial \theta^2}^\top da_0^1 da_0^2 \nonumber\\
&+ \int_{a_0^1} \int_{a_0^2}  \pi(a_0^1|s_0;\theta^1)  \frac{\partial Q(s_0,a_0^1,a_0^2;\theta^1\!\!,\theta^2)} {\partial \theta^1} \frac{\partial \pi(a_0^2|s_0;\theta^2)}{\partial \theta^2}^\top da_0^1 da_0^2 \nonumber\\ 
&+ \int_{a_0^1} \! \int_{a_0^2} \!\!\! \pi(a_0^1|s_0;\theta^1) \pi(a_0^2|s_0;\theta^2) \Biggl( \!\!\gamma \!\! \int_{s_1}\!\! T(s_1|s_0,a_0^1,a_0^2) \! \nonumber \\
& \quad \quad \biggl(\! \int_{a_1^1} \!\!\int_{a_1^2} \!\! \frac{\partial \pi(a_1^1|s_1;\theta^1\!)}{\partial \theta^1} \frac{\partial \pi(a_1^2|s_1;\theta^2\!)}{\partial \theta^2}^\top  Q(s_1,a_1^1,a_1^2;\theta^1\!,\theta^2\!) da_1^1 da_1^2 \nonumber\\
&+ \int_{a_1^1} \int_{a_1^2} \frac{\partial \pi(a_1^1|s_1;\theta^1)}{\partial \theta^1}  \pi(a_1^2|s_1;\theta^2) \frac{\partial Q(s_1,a_1^1,a_1^2;\theta^1\!\!,\theta^2)}{\partial \theta^2}^\top da_1^1 da_1^2 \nonumber\\
&+ \int_{a_1^1} \int_{a_1^2}  \pi(a_1^1|s_1;\theta^1)  \frac{\partial Q(s_1,a_1^1,a_1^2;\theta^1\!\!,\theta^2)} {\partial \theta^1} \frac{\partial \pi(a_1^2|s_1;\theta^2)}{\partial \theta^2}^\top da_1^1 da_1^2 \nonumber\\
& \!\!\!\!\!\!\!\!\!\!\!\!\!\!\!\!\!\!\!\!\!\!\!\!\!\!\!\!\!\!\!\!\!\!\!\!\!\!\!\!\!\!\!\!\!\!\! + \!\int_{a_1^1}\!\! \int_{a_1^2} \!\!\! \pi(a_1^1|s_1;\theta^1)   \pi(a_1^2|s_1;\theta^2)  \! \gamma \!\!\int_{s_2}\!\!\!\! T(s_2|s_1,a_1^1,a_1^2)  \frac{\partial^2 V(s_2;\theta^1,\theta^2)}{\partial \theta^2 \partial \theta^1} ds_2  da_1^1  da_1^2 \!\biggr) ds_1 \!\!\! \Biggr) da_0^1 da_0^2. \nonumber
\end{align}
Exploring the recursive structure, quickly leads to long sequences of terms, to deal with this let us define a short hand notation, the first term in \eq~\ref{eqn: doublegradv} is denoted by $b_0$, second term by $c_0$, third term by $d_0$, forth term as $e_0$, note that the subscript indicates the time step and let $Pr_k = Pr(s_k\rightarrow s_{k+1}, \theta^1,\theta^2)$ be the state transition probability summed over all actions. When investigating \eq~\ref{eqn: subs_double_gradv} note that the forth term of \eq~\ref{eqn: doublegradv} $e_0$, expanded to $e_0 = \gamma \int_{s_1} Pr_0 ( b_1 + c_1 + d_1 + e_1) ds_1$. Thus, recursively applying this insight and using our short hand notation results in,
\begin{align}
D_{\theta^1 \theta^2} V(s_0;\theta^1,\theta^2) = &b_0 \!+\! c_0 \!+\! d_0 
\!+\! \gamma \int_{s_1}\!\!Pr_0 \Bigl(\!b_1 \!+\! c_1 \!+\! d_1 
\!+\! \gamma \int_{s_2}\!\!Pr_1 \Bigl(\!b_2 \!+\! c_2 \!+\! d_2 \nonumber\\
&\!+\! \gamma \int_{s_3}\!\!Pr_2 \Bigl(\!b_3 \!+\! c_3 \!+\! d_3 
\!+\! \gamma \int_{s_4}\!\!Pr_3 \Bigl(\!...\!\Bigr)ds_4\!\Bigr)ds_3\!\Bigr)ds_2\!\Bigr)ds_1 \nonumber\\
 = & b_0 \!+\! \gamma \!\!\int_{s_1}\!\!Pr_0 \Bigl(\!b_1 \!+\! \gamma \!\! \int_{s_2}\!\!Pr_1 \Bigl(\!b_2 \!+\! \gamma \!\!\int_{s_3}\!\!Pr_2 \Bigl(\!b_3 \!+\! \gamma \!\! \int_{s_4}\!\!Pr_3 \Bigl(\!...\!\Bigr)ds_4\!\Bigr)ds_3\!\Bigr)ds_2\!\Bigr)ds_1 \nonumber\\
&\!\!\!\!\!\!\!\! \!+\! c_0 \!+\!  \gamma \!\! \int_{s_1}\!\!Pr_0 \Bigl(\!c_1 \!+\! \gamma \!\! \int_{s_2}\!\!Pr_1 \Bigl(\!c_2 \!+\! \gamma \int_{s_3}\!\!Pr_2 \Bigl(\!c_3 \!+\! \gamma \!\! \int_{s_4}\!\!Pr_3 \Bigl(\!...\!\Bigr)ds_4\!\Bigr)ds_3\!\Bigr)ds_2\!\Bigr)ds_1 \nonumber\\
&\!\!\!\!\!\!\!\! \!+\! d_0 \!+\! \gamma \!\! \int_{s_1}\!\!Pr_0 \Bigl(\!d_1 \!+\! \gamma \!\! \int_{s_2}\!\!Pr_1 \Bigl(\!d_2 \!+\! \gamma \!\! \int_{s_3}\!\!Pr_2 \Bigl(\!d_3 \!+\! \gamma \!\! \int_{s_4}\!\!Pr_3 \Bigl(\!...\!\Bigr)ds_4\!\Bigr)ds_3\!\Bigr)ds_2\!\Bigr)ds_1. \nonumber
\end{align}
Summing over the initial state distribution results in,
\begin{align}
\!\!D_{\theta^1 \theta^2} \eta(\theta^1,\theta^2) \!&= \!\!\int_{s_0}\!\!\!\! p(s_0) D_{\theta^1 \theta^2} V(s_0;\theta^1,\theta^2) ds_0 \nonumber\\
&\!\!\!\!\!\!\!\!\!\!\!\!\!\!\!\!\!\!\!\!= \!\!\int_{s_0}\!\!\!\! p(s_0) \Bigl(\!b_0 \!\!+\!\! \gamma \!\!\int_{s_1}\!\!\!\!Pr_0 \Bigl(\!b_1 \!\!+\!\! \gamma \!\!\int_{s_2}\!\!\!\!Pr_1 \Bigl(\!b_2 \!\!+\!\! \gamma \!\!\int_{s_3}\!\!\!\!Pr_2 \Bigl(\!b_3 \!\!+\!\! \gamma \!\!\int_{s_4}\!\!\!\!Pr_3 \Bigl(\!...\!\Bigr)ds_4\!\Bigr)ds_3\!\Bigr)ds_2\!\Bigr)ds_1\!\Bigr)ds_0\nonumber\\
&\quad\!\!\!\!\!\!\!\!\!\!\!\!\!\!\!\!\!\!\!\!+\! \!\!\int_{s_0}\!\!\!\! p(s_0)\Bigl(\!c_0 \!\!+\!\! \gamma \!\!\int_{s_1}\!\!\!\!Pr_0 \Bigl(\!c_1 \!\!+\!\! \gamma \!\!\int_{s_2}\!\!\!\!Pr_1 \Bigl(\!c_2 \!\!+\!\! \gamma \!\!\int_{s_3}\!\!\!\!Pr_2 \Bigl(\!c_3 \!\!+\!\! \gamma \!\!\int_{s_4}\!\!\!\!Pr_3 \Bigl(\!...\!\Bigr)ds_4\!\Bigr)ds_3\!\Bigr)ds_2\!\Bigr)ds_1\!\Bigr)ds_0\nonumber\\
&\quad\!\!\!\!\!\!\!\!\!\!\!\!\!\!\!\!\!\!\!\!+\! \!\!\int_{s_0}\!\!\!\!\! p(s_0)\Bigl(\!d_0 \!\!+\!\! \gamma \!\!\int_{s_1}\!\!\!\!\!Pr_0 \Bigl(\!d_1 \!\!+\!\! \gamma \!\!\int_{s_2}\!\!\!\!\!Pr_1 \Bigl(\!d_2 \!\!+\!\! \gamma \!\!\int_{s_3}\!\!\!\!\!Pr_2 \Bigl(\!d_3 \!\!+\!\! \gamma \!\!\int_{s_4}\!\!\!\!\!Pr_3 \Bigl(\!...\!\Bigr)ds_4\!\Bigr)ds_3\!\Bigr)ds_2\!\Bigr)ds_1\!\Bigr)ds_0. \label{eq: shortbilinear}
\end{align}
\eq~\ref{eq: shortbilinear} is clearly build up by three terms the first depending on $b_k$, the second on $c_k$ and the last on $d_k$, let us denote these terms as $\mathscr{T}_1$,$\mathscr{T}_2$ and $\mathscr{T}_3$ respectively. Thus we have $D_{\theta^1 \theta^2} \eta(\theta^1,\theta^2) = \mathscr{T}_1 + \mathscr{T}_2 + \mathscr{T}_3$. Given this separation, we can tackle each part separately.
\vspace{\baselineskip}
\noindent\textbf{Solving for the $\mathscr{T}_1$:}\\
To bring $\mathscr{T}_1$ in \eq~\ref{eq: shortbilinear}, to the final form in our theorem, let us substitute the terms for $b_k$ back in,
\begin{align}
\mathscr{T}_1 &= \int_{s_0} \!\!\!\! p(s_0) \Biggl(\int_{a_0^1} \!\! \int_{a_0^2} \!\! \frac{\partial \pi(a_0^1|s_0;\theta^1)}{\partial \theta^1} \frac{\partial \pi(a_0^2|s_0;\theta^2)}{\partial \theta^2}^\top \!\!\!Q(s_0,a_0^1,a_0^2;\theta^1\!\!,\theta^2)  da_0^1  da_0^2 \nonumber\\
&+ \!\!\gamma\!\! \int_{s_1}\!\!\! Pr(s_1|s_0;\theta^1,\theta^2) \biggl( \int_{a_1^1} \!\! \int_{a_1^2} \!\! \frac{\partial \pi(a_1^1|s_1;\theta^1)}{\partial \theta^1}  \frac{\partial \pi(a_1^2|s_1;\theta^2)}{\partial \theta^2}^\top\!\!\!  Q(s_1,a_1^1,a_1^2;\theta^1\!\!,\theta^2) da_1^1 da_1^2 \nonumber\\
&+ \!\!\int_{a_1^1} \!\! \int_{a_1^2} \!\!\!\! \pi(a_1^1|s_1;\theta^1) \pi(a_1^2|s_1;\theta^2)  \gamma \int_{s_2} T(s_2|s_1,a_1^1,a_1^2) \Bigl(...\Bigr)  ds_2 da_1^1 da_1^2 \!\! \biggr) ds_1 da_0^1 da_0^2 \!\! \Biggr) ds_0 . \numberthis
\shortintertext{Similar to \eq~\ref{eqn: recursive_grad2_traj} we can unrolling and marginalizing, which allows us to bring it into a compact form,} 
\label{eqn: summed_gradv}
\mathscr{T}_1 &=\!\! \int_{\tau} \!\sum_{k=0}^{|\tau|-1} \!\!\gamma^k f(\tau_{0:k-1},s_k;\theta^1,\theta^2) \frac{\partial \pi(a_k^1|s_k;\theta^1)}{\partial \theta^1} \frac{\partial \pi(a_k^2|s_k;\theta^2)}{\partial \theta^2}^\top \!\!\!Q(s_k,a_k^1,a_k^2;\theta^1\!\!,\theta^2)  d\tau \nonumber\\
&=\!\!\int_{\tau}\! \sum_{k=0}^{|\tau|-1} \!\! \gamma^k \! f(\tau_{0:k};\theta^1\!\!,\theta^2) \frac{\partial \log \pi(a_k^1|s_k;\theta^1)}{\partial \theta^1} \frac{\partial \log \pi(a_k^2|s_k;\theta^2)}{\partial \theta^2}^\top \!\!\!Q(s_k,a_k^1,a_k^2;\theta^1\!\!,\theta^2) d\tau .
\end{align}

\vspace{\baselineskip}
\noindent\textbf{Solving for the $\mathscr{T}_2$:}\\
For $\mathscr{T}_2$ in \eq~\ref{eq: shortbilinear}, we again substitute the $c_k$ terms back in, which gets us,
\begin{align}
\mathscr{T}_2 = \int_{s_0}  p(s_0) \Biggl( & \int_{a_0^1} \int_{a_0^2}  \frac{\partial \pi(a_0^1|s_0;\theta^1)}{\partial \theta^1} \pi(a_0^2|s_0;\theta^2) \frac{\partial Q(s_0,a_0^1,a_0^2;\theta^1,\theta^2)}{\partial \theta^2}^\top  da_0^1 da_0^2 \nonumber\\
&\quad+\int_{a_0^1}\int_{a_0^2}\pi(a_0^1|s_0;\theta^1)  \pi(a_0^2|s_0;\theta^2)\gamma\int_{s_1}T(s_1|s_0,a_0^1,a_0^2) \nonumber\\
&\qquad\qquad \biggl( \int_{a_1^1}\int_{a_1^2}\frac{\partial \pi(a_1^1|s_1;\theta^1)}{\partial \theta^1}  \pi(a_1^2|s_1;\theta^2) \frac{\partial Q(s_1,a_1^1,a_1^2;\theta^1,\theta^2)}{\partial \theta^2}^\top  da_1^1 da_1^2 \nonumber\\
&\qquad\qquad\qquad \ \quad+\int_{a_1^1} \int_{a_1^2}  \pi(a_1^1|s_1;\theta^1)  \pi(a_1^2|s_1;\theta^2) \gamma  \int_{s_2}  T(s_2|s_1,a_1^1,a_1^2) \nonumber\\
&\qquad\qquad\qquad\qquad\quad\qquad\qquad\qquad \Bigl(...\Bigr)ds_2\ da_1^1 da_1^2  \biggr)ds_1 da_0^1 da_0^2 \Biggr) ds_0 . \label{eqn: term2_unrolling} 
\end{align}
Note that $\mathscr{T}_2$ contains $\frac{\partial Q(s_k,a_k^1,a_k^2;\theta^1\!\!,\theta^2) }{\partial \theta^2}$. We can unroll $\frac{\partial Q(s_0,a_0^1,a_0^2;\theta^1\!\!,\theta^2)}{\partial \theta^2}$ using \eq~\ref{eqn: gradq} and \eq~\ref{eqn: recursive_gradv}, till the next state $s_1$, which results in,
\begin{align}
\mathscr{T}_2 \! = \!\! \int_{s_0} \!\! p(s_0) \Biggl( & \int_{a_0^1}  \int_{a_0^2}  \frac{\partial \pi(a_0^1|s_0;\theta^1)}{\partial \theta^1} \pi(a_0^2|s_0;\theta^2)  \int_{s_1}  \gamma T(s_1|s_0,a_0^1,a_0^2) \nonumber \\
& \quad \biggl(
\int_{a_1^1}  \int_{a_1^2}  \pi(a_1^1|s_1;\theta^1) \frac{\partial \pi(a_1^2|s_1;\theta^2)}{\partial \theta^2} \nonumber
Q(s_1,a_1^1,a_1^2;\theta^1,\theta^2) da_1^1 da_1^2 \\
& \qquad  +  \int_{a_1^1} \!\! \int_{a_1^2} \!\!\!\! \pi(\!a_1^1|s_1;\theta^1)  \pi(\!a_1^2|s_1;\theta^2) \frac{\partial Q(\!s_1,a_1^1,a_1^2;\theta^1,\theta^2)}{\partial \theta^2} da_1^1 da_1^2 \biggr)^\top \!\!\!\! ds_1 da_0^1 da_0^2 \nonumber\\
&+\int_{a_0^1}\int_{a_0^2}\pi(a_0^1|s_0;\theta^1)  \pi(a_0^2|s_0;\theta^2)\gamma\int_{s_1}T(s_1|s_0,a_0^1,a_0^2) \nonumber \\
& \quad \qquad \biggl( \int_{a_1^1}\int_{a_1^2}\frac{\partial \pi(a_1^1|s_1;\theta^1)}{\partial \theta^1}  \pi(a_1^2|s_1;\theta^2) \frac{\partial Q(s_1,a_1^1,a_1^2;\theta^1,\theta^2)}{\partial \theta^2}^\top  da_1^1 da_1^2\nonumber\\
&\qquad \quad \qquad +\int_{a_1^1}\int_{a_1^2}  \pi(a_1^1|s_1;\theta^1)  \pi(a_1^2|s_1;\theta^2) \gamma  \int_{s_2}  T(s_2|s_1,a_1^1,a_1^2)\nonumber\\
&\qquad\qquad\qquad\quad\qquad\qquad\qquad\qquad\qquad
\ \Bigl(...\Bigr) ds_2 da_1^1 da_1^2 \biggr) ds_1  da_0^1 da_0^2 \Biggr) ds_0 . \nonumber  
\end{align}
Using log trick $\left( D_{\theta} \log p_{\theta}(\tau) = D_{\theta} p_{\theta}(\tau)/p_{\theta}(\tau)\right)$ and combining terms together, we get,
\begin{align} 
\mathscr{T}_2 \!=\! \int_{s_0} \!\! p(s_0) \!\Biggl( &\! \int_{a_0^1} \! \int_{a_0^2} \!\!\! \pi(a_0^1|s_0;\theta^1) \pi(a_0^2|s_0;\theta^2)  \int_{s_1} \!\!\! \gamma T(s_1|s_0,a_0^1,a_0^2) \int_{a_1^1} \! \int_{a_1^2} \!\!\!\! \pi(a_1^1|s_1;\theta^1) \pi(a_1^2|s_1;\theta^2)\nonumber\\  
& \quad \biggl( \frac{\partial \log \pi(a_0^1|s_0;\theta^1)}{\partial \theta^1} \frac{\partial \log \pi(a_1^2|s_1;\theta^2)}{\partial \theta^2}^\top  Q(s_1,a_1^1,a_1^2;\theta^1,\theta^2) \nonumber\\
& \quad \qquad + \biggl(  \frac{\partial \log \pi(a_0^1|s_0;\theta^1)}{\partial \theta^1}  +  \frac{\partial \log \pi(a_1^1|s_1;\theta^1)}{\partial \theta^1}  \biggr)\nonumber
\frac{\partial Q(s_1,a_1^1,a_1^2;\theta^1,\theta^2)}{\partial \theta^2}^\top  \\
& \qquad\qquad\qquad\quad\quad +  \gamma  \int_{s_2} \!\!\!\! T(s_2|s_1,a_1^1,a_1^2) \Bigl(...\Bigr) ds_2  \biggr) da_1^1 da_1^2 ds_1 da_0^1 da_0^2 \Biggr) ds_0 \nonumber
\end{align}
\begin{align}
\!\!\!\!\!\!\!&= \int_{\tau}f(\tau_{0:1};\theta^1,\theta^2) \gamma \Biggl( \frac{\partial \log \pi(a_0^1|s_0;\theta^1)}{\partial \theta^1} \frac{\partial \log \pi(a_1^2|s_1;\theta^2)}{\partial \theta^2}^\top \!\! Q(s_1,a_1^1,a_1^2;\theta^1\!\!,\theta^2) \nonumber \\
&\quad\!\! + \!\! \biggl( \! \frac{\partial \log \pi(\!a_0^1|s_0;\!\theta^1\!)}{\partial \theta^1} \! + \! \frac{\partial \log \pi(\!a_1^1|s_1;\!\theta^1\!)}{\partial \theta^1} \! \biggr)\! \frac{\partial Q(\!s_1,a_1^1,a_1^2;\!\theta^1\!\!,\theta^2\!)}{\partial \theta^2}^\top \!\!\!\! + \!\gamma \!\! \int_{s_2} \!\!\!\! T(\!s_2|s_1,a_1^1,a_1^2) \Bigl(\!...\!\Bigr) ds_2 \!\! \Biggr) \! d\tau_{0:1} . \nonumber
\end{align}
We can use the identical method for the remaining  $\frac{\partial Q(s_k,a_k^1,a_k^2;\theta^1\!\!,\theta^2) }{\partial \theta^2}$. Thus unrolling this term for one time step and using the log trick to combine terms. For example unrolling the second step results in,
\begin{align}
\!\!\mathscr{T}_2\!&= \! \int_{\tau} \!\! f(\tau_{0:1};\!\theta^1\!\!,\theta^2) \gamma \frac{\partial \log \pi(a_0^1|s_0;\theta^1)}{\partial \theta^1} \frac{\partial \log \pi(a_1^2|s_1;\!\theta^2)}{\partial \theta^2}^\top \!\!\!  Q(s_1,a_1^1,a_1^2;\theta^1\!\!,\theta^2) d\tau_{0:1} \nonumber \\
&\quad+ \!\! \int_{\tau} \!\!\! f(\!\tau_{0:2};\!\theta^1\!\!,\theta^2) \gamma^2 \! \Biggl( \!\! \biggl( \! \frac{\partial \log \pi(\!a_0^1|s_0;\!\theta^1\!)}{\partial \theta^1} \!+\! \frac{\partial \log \pi(\!a_1^1|s_1;\!\theta^1\!)}{\partial \theta^1} \! \biggr) \! \frac{\partial \log \pi(\!a_2^2|s_2;\theta^2\!)}{\partial \theta^2}^\top \!\!\! Q(\!s_2,a_2^1,a_2^2;\!\theta^1\!\!,\theta^2\!) \nonumber\\
&\quad+ \!\! \biggl( \! \frac{\partial \log \pi(a_0^1|s_0;\!\theta^1)}{\partial \theta^1} \!+\! \frac{\partial \log \pi(a_1^1|s_1;\!\theta^1)}{\partial \theta^1} \!+\! \frac{\partial \log \pi(a_2^1|s_2;\!\theta^1)}{\partial \theta^1} \biggr) \!\frac{\partial Q(s_2,a_2^1,a_2^2;\!\theta^1\!\!,\theta^2)}{\partial \theta^2}^\top \!\!\!\! + \!\! \Bigl(\!\!...\!\!\Bigr) \!\!\! \Biggr)\! d\tau_{0:2} . \nonumber
\end{align}
Finally, on unrolling for $|\tau|$ steps we get,
\begin{align}
\mathscr{T}_2 \! &= \!\!\! \int_{\tau} \! \sum_{k=1}^{|\tau|-1} \!\!\! \gamma^k\! f(\!\tau_{0:k};\!\theta^1\!\!,\theta^2) \frac{\partial \log(\prod_{i=0}^{k-1}\!\pi(\!a_{i}^1|s_i;\!\theta^1)\!)}{\partial \theta^1} \frac{\partial \log \pi(\!a_{k}^2|s_{k};\!\theta^2)\!)}{\partial \theta^2}^\top  \!\!\! Q(\!s_k,a_{k}^1,a_{k}^2;\!\theta^1\!\!,\theta^2)d\tau.
\end{align}
\vspace{\baselineskip}
\noindent\textbf{Solving for the $\mathscr{T}_3$:}\\
In $\mathscr{T}_3$ in \eq~\ref{eq: shortbilinear}, similar to the previous two terms we again substitute $d_k$ back in, which results in,
\begin{align}
\mathscr{T}_3 = \int_{s_0} \!\!\! p(s_0) \Biggl(&  \int_{a_0^1}  \int_{a_0^2}  \pi(a_0^1|s_0;\theta^1) \frac{\partial Q(s_0,a_0^1,a_0^2;\theta^1,\theta^2)} {\partial \theta^1} \frac{\partial \pi(a_0^2|s_0;\theta^2)}{\partial \theta^2}^\top  da_0^1 da_0^2 \nonumber\\ 
& \quad + \int_{a_0^1}\int_{a_0^2}\pi(a_0^1|s_0;\theta^1) \pi(a_0^2|s_0;\theta^2) \gamma\int_{s_1}T(s_1|s_0;a_0^1,a_0^2) \nonumber\\
& \qquad \qquad \biggl(\int_{a_1^1} \int_{a_1^2}\pi(a_1^1|s_1;\theta^1)  \frac{\partial Q(s_1,a_1^1,a_1^2;\theta^1,\theta^2)} {\partial \theta^1}  \frac{\partial \pi(a_1^2|s_1;\theta^2)}{\partial \theta^2}^\top  da_1^1  da_1^2 \nonumber\\
& \qquad \qquad \qquad +\int_{a_1^1} \int_{a_1^2}  \pi(a_1^1|s_1;\theta^1)  \pi(a_1^2|s_1;\theta^2) \gamma  \int_{s_2}  T(s_2|s_1,a_1^1,a_1^2) \nonumber\\
& \qquad\qquad\qquad\qquad\qquad\qquad\qquad\qquad\qquad\Bigl(...\Bigr)ds_2 da_1^1 da_1^2  \biggr)ds_1 da_0^1 da_0^2 \Biggr) ds_0. \nonumber
\label{eqn: term3_unrolling}
\end{align}
Similar to $\mathscr{T}_2$, $\mathscr{T}_3$ contains $\frac{\partial Q(s_k,a_k^1,a_k^2;\theta^1\!\!,\theta^2) }{\partial \theta^1}$, using the same approach and unfold this term using \eq~\ref{eqn: gradq} and \eq~\ref{eqn: recursive_gradv} for one step results in, 
\begin{align}
\!\!\!\!\mathscr{T}_3 \! = \! \int_{s_0} \!\!\!  p(s_0) \Biggl(& \int_{a_0^1}  \int_{a_0^2}  \pi(a_0^1|s_0;\theta^1)  \int_{s_1} \gamma T(s_1|s_0;a_0^1,a_0^2) \nonumber \\
& \quad \biggl(
\int_{a_1^1}  \int_{a_1^2}  \frac{\partial \pi(a_1^1|s_1;\theta^1)}{\partial \theta^1}  \pi(a_1^2|s_1;\theta^2) Q(s_1,a_1^1,a_1^2;\theta^1,\theta^2) da_1^1  da_1^2 \nonumber\\
& \qquad \quad + \int_{a_1^1}  \int_{a_1^2}  \pi(\!a_1^1|s_1;\theta^1)
\pi(\!a_1^2|s_1;\theta^2) \frac{\partial Q(s_1,a_1^1,a_1^2;\theta^1,\theta^2)}{\partial \theta^1} da_1^1 da_1^2 \nonumber \\ 
& \qquad \qquad \qquad \qquad \qquad \qquad \qquad \qquad \qquad \biggl) ds_1 \frac{\partial \pi(a_0^2|s_0;\theta^2)}{\partial \theta^2}^\top  da_0^1 da_0^2 \nonumber\\
&+\int_{a_0^1}\int_{a_0^2}\pi(a_0^1|s_0;\theta^1) \pi(a_0^2|s_0;\theta^2) \gamma\int_{s_1}T(s_1|s_0;a_0^1,a_0^2) \nonumber \\
& \qquad \biggl(\int_{a_1^1} \int_{a_1^2}\pi(a_1^1|s_1;\theta^1)  \frac{\partial Q(s_1,a_1^1,a_1^2;\theta^1,\theta^2)} {\partial \theta^1}  \frac{\partial \pi(a_1^2|s_1;\theta^2)}{\partial \theta^2}^\top  da_1^1  da_1^2 \nonumber\\
& \qquad \qquad \qquad + \int_{a_1^1} \int_{a_1^2}  \pi(a_1^1|s_1;\theta^1)  \pi(a_1^2|s_1;\theta^2) \gamma  \int_{s_2}  T(s_2|s_1,a_1^1,a_1^2)\nonumber \\
& \qquad\qquad\qquad\qquad\qquad\qquad\qquad\qquad\Bigl(...\Bigr)  ds_2  da_1^1  da_1^2  \biggr) ds_1 da_0^1 da_0^2  \Biggr) ds_0 . \nonumber
\end{align}

Using the log trick ($D_{\theta} \log p_{\theta}(\tau) = D_{\theta} p_{\theta}(\tau)/p_{\theta}(\tau)$) and combining terms together, we get:
\begin{align}
\mathscr{T}_3 = \! \int_{s_0} \!\! p(s_0) \Biggl( & \! \int_{a_0^1} \! \int_{a_0^2} \!\!\! \pi(a_0^1|s_0;\theta^1) \pi(a_0^2|s_0;\theta^2) \!\! \int_{s_1} \!\!\!\! \gamma T(s_1|s_0;a_0^1,a_0^2) \!
\int_{a_1^1} \! \int_{a_1^2} \!\!\! \pi(a_1^1|s_1;\theta^1) \pi(a_1^2|s_1;\theta^2) \nonumber\\
& \quad \biggl(\frac{\partial \log \pi(a_0^1|s_0;\theta^1)}{\partial \theta^1} \frac{\partial \log \pi(a_1^2|s_1;\theta^2)}{\partial \theta^2}^\top   Q(s_1,a_1^1,a_1^2;\theta^1,\theta^2) \nonumber\\
& \qquad \quad + \frac{\partial Q(s_1,a_1^1,a_1^2;\theta^1,\theta^2)}{\partial \theta^1}  \biggl(  \frac{\partial \log \pi(a_0^2|s_0;\theta^2)}{\partial \theta^2} +  \frac{\partial \log \pi(a_1^2|s_1;\theta^2)}{\partial \theta^2}  \biggr)^\top \nonumber \\
& \qquad\qquad\qquad\qquad + \gamma  \int_{s_2}  T(s_2|s_1,a_1^1,a_1^2) \Bigl(...\Bigr)\ ds_2  \biggr) da_1^1 da_1^2 ds_1 da_0^1 da_0^2  \Biggr) ds_0, \nonumber 
\end{align}
by simplifying the notation we get,
\begin{align}
\mathscr{T}_3 =\!\!& \int_{\tau} \!\! f(\tau_{0:1};\theta^1,\theta^2) \gamma \biggl( \frac{\partial \log \pi(a_0^1|s_0;\theta^1)}{\partial \theta^1} \frac{\partial \log \pi(a_1^2|s_1;\theta^2)}{\partial \theta^2}^\top \!\!\!  Q(s_1,a_1^1,a_1^2;\theta^1\!\!,\theta^2) \nonumber \\
&\!\!\!+ \!\! \frac{\partial Q(\!s_1,a_1^1,a_1^2;\!\theta^1\!\!,\theta^2)}{\partial \theta^1} \!\biggl(\!\frac{\partial \log \pi(\!a_0^2|s_0;\!\theta^2)}{\partial \theta^2} \!+\! \frac{\partial \log \pi(\!a_1^2|s_1;\!\theta^2)}{\partial \theta^2} \! \biggr)^\top \!\!\!\!\! + \! \gamma \!\! \int_{s_2} \!\!\! T(\!s_2|s_1,\!a_1^1,\!a_1^2) \Bigl(\!...\!\Bigr) ds_2 \!\! \biggr)\! d\tau_{0:1}. \nonumber
\end{align}
The remaining terms in the recursive structure again contain terms of the form $\frac{\partial Q(s_k,a_k^1,a_k^2;\theta^1\!\!,\theta^2) }{\partial \theta^1}$, which we simplify in a similar fashion by unfolding for one step, for example unfolding the second term for one step results in,
\begin{align}
\mathscr{T}_3 = & \!\! \int_{\tau} \!\! f(\tau_{0:1};\theta^1\!\!,\theta^2) \gamma \frac{\partial \log \pi(a_0^1|s_0;\theta^1)}{\partial \theta^1} \frac{\partial \log \pi(a_1^2|s_1;\theta^2)}{\partial \theta^2}^\top \!\!\! Q(s_1,a_1^1,a_1^2;\theta^1\!\!,\theta^2) d\tau_{0:1} \nonumber \\
&\!\!+ \!\! \int_{\tau}\!\!\!f(\!\tau_{0:2};\!\theta^1\!\!,\theta^2) \gamma^2 \!\Biggl( \!\! \frac{\partial \log \pi(\!a_2^1|s_2;\theta^1\!)}{\partial \theta^1} \!  \biggl( \! \frac{\partial \log \pi(\!a_0^2|s_0;\!\theta^2)}{\partial \theta^2} \!+\! \frac{\partial \log \pi(\!a_1^2|s_1;\!\theta^2)}{\partial \theta^2} \! \biggr)^\top \!\!\!\!  Q(\!s_2,a_2^1,a_2^2;\!\theta^1\!\!,\theta^2) \nonumber\\
&\!\!+ \!\! \frac{\partial Q(\!s_2,a_2^1,a_2^2;\theta^1\!\!,\theta^2)}{\partial \theta^1} \! \biggl( \! \frac{\partial \log \pi(\!a_0^2|s_0;\theta^2)}{\partial \theta^2} \!+\! \frac{\partial \log \pi(\!a_1^2|s_1;\theta^2)}{\partial \theta^2} \!+\! \frac{\partial \log \pi(\!a_2^2|s_2;\theta^2)}{\partial \theta^2} \! \biggr)^\top \!\!\!\! + \Bigl(\!...\!\Bigr) \!\! \Biggr)\!d\tau_{0:2} . \nonumber
\end{align}
Finally on unrolling for $|\tau|$ steps we get,
\begin{align}
\!\!\!\!= \! \int_{\tau} \! \sum_{k=1}^{|\tau|-1} \!\!\! \gamma^k f(\tau_{0:k};\theta^1\!\!,\theta^2) \frac{\partial \log \pi(a_{k}^1|s_{k};\theta^1)}{\partial \theta^1} \frac{\partial \log(\prod_{l=0}^{k-1}\pi(a_{l}^2|s_l;\theta^2))}{\partial \theta^2}^\top \!\!\! Q(s_k,a_{k}^1,a_{k}^2;\theta^1\!\!,\theta^2) d\tau .
\end{align}

\vspace{\baselineskip}
\noindent\textbf{Combining all 3 terms:}\\
By \eq~\ref{eq: shortbilinear} we know that the bilinear term of the game objective is given by $D_{\theta^1 \theta^2} \eta(\theta^1,\theta^2) = \mathscr{T}_1 + \mathscr{T}_2 + \mathscr{T}_3$, or in other words the sum of the three terms derived above. Thus, we derived that,
\begin{align}
\!\!\!\!D_{\theta^1 \theta^2} \eta(\theta^1\!\!,\theta^2)\!\! &= \!\!\! \int_{\tau} \! \sum_{k=0}^{|\tau|-1} \!\!\! \gamma^k f(\tau_{0:k};\theta^1\!\!,\theta^2) \frac{\partial \log \pi(a_k^1|s_k;\theta^1)}{\partial \theta^1} \frac{\partial \log \pi(a_k^2|s_k;\theta^2)}{\partial \theta^2}^\top \!\!\! Q(s_k,a_k^1,a_k^2;\theta^1\!\!,\theta^2)  d\tau \nonumber \\
& \!\!\!\!\!\!\!\!\!\!\!\!\!\!\!\!\!\!\!\!\!\!\!\!\!\!\!\!\! + \!\!\! \int_{\tau} \! \sum_{k=1}^{|\tau|-1} \!\!\! \gamma^k \! f(\tau_{0:k};\theta^1\!\!,\theta^2) \frac{\partial \log(\prod_{l=0}^{k-1}\pi(a_{l}^1|s_l;\theta^1))}{\partial \theta^1} \frac{\partial \log \pi(a_{k}^2|s_{k};\theta^2)}{\partial \theta^2}^\top \!\!\!  Q(s_k,a_{k}^1,a_{k}^2;\theta^1\!\!,\theta^2) d\tau \nonumber\\
& \!\!\!\!\!\!\!\!\!\!\!\!\!\!\!\!\!\!\!\!\!\!\!\!\!\!\!\!\! + \!\!\! \int_{\tau} \! \sum_{k=1}^{|\tau|-1} \!\!\! \gamma^k \! f(\tau_{0:k};\theta^1\!\!,\theta^2) \frac{\partial \log \pi(a_{k}^1|s_{k};\theta^1)}{\partial \theta^1} \frac{\partial \log(\prod_{l=0}^{k-1} \! \pi(a_{l}^2|s_l;\theta^2))}{\partial \theta^2}^\top \!\!\! Q(s_k,a_{k}^1,a_{k}^2;\theta^1\!\!,\theta^2) d\tau .
\end{align}
This concludes our derivation of the bilinear term of the game objective.
\section{Generalization of \cpg to variants of return}
\label{secappx: advantagebilinear}
\subsection{Estimation of gradient and bilinear term using a baseline}
The gradient and bilinear terms can be estimated using \eq~\ref{eq:CPGGradHessianVal1} and \eq~\ref{eq:CPGGradHessianVal} based on $Q(s_k, a^1_k, a^2_k;\theta^1,\theta^2)$ the state-action value function. But in practice, this can perform poorly due to variance and missing the notion of relative improvement. During learning, policy parameters are updated to increase the probability of visiting good trajectories and reducing chances of encountering bad trajectories. But, let us assume good samples have zero reward and bad samples have a negative reward. In this case, the policy distribution will move away from the negative samples but can move in any direction as zero reward does not provide information on being better a state-action pair. This causes high variance and slow learning. In another case, let us assume that a non zero constant is added to the reward function for all states-action pairs, ideally this constant factor should not affect the policy update but in the current formulation using $Q(s_k, a^1_k,a^2_k;\theta^1,\theta^2)$ it does because the notion of better or worse sample is with respect to zero. Hence, missing the notion of relative improvement.

To mitigate the issues, a baseline $b$ can be subtracted from $Q(s, a^1_k, a^2_k;\theta^1\!\!,\theta^2)$. The baseline can be any random variable as long as it is not a function of $a^1_k$ and $a^2_k$ ~\citep{Sutton2018}. For an \MDP, a good baseline $b$ is a function of the state $b(s)$, because let say in some states all actions have high values and we need a high baseline to differentiate the higher valued actions from the less highly valued ones; in other states all actions will have low values and a low baseline is appropriate. This is equivalent to $Q(s_k, a^1_k, a^2_k;\theta^1,\theta^2) \xleftarrow{} Q(s_k, a^1_k, a^2_k;\theta^1,\theta^2) - b(s_k)$ in \eq~\ref{eq:CPGGradHessianVal} and \eq~\ref{eq:CPGGradHessianVal1}.

\subsection{Extending \thm~\ref{thm: CPGGradHessianVal} for gradient and bilinear estimation using advantage function}
A popular choice for a baseline $b(s)$ is $V(s,\theta^1,\theta^2)$, which results in the advantage function defined in \sect~\ref{sec:cpg_theory}. In \thm~\ref{thm: CPGGradHessianAdv} we extend \thm~\ref{thm: CPGGradHessianVal} to estimate gradient and bilinear terms using the advantage function.

\begin{theorem}\label{thm: CPGGradHessianAdv}
Given that the advantage function is defined as $A(s,a^1,a^2; \theta^1,\theta^2) = Q(s,a^1,a^2; \theta^1,\theta^2) - V(s; \theta^1,\theta^2)$ and given \eq~\ref{eq:CPGGradHessianVal1}, then for player $i,j\in\lbrace1,2\rbrace$ and $i \neq j$, it holds that,
\begin{align}
D_{\theta^i} \eta\! &= \! \int_{\tau} \! \sum_{k=0}^{|\tau|-1} \!\! \gamma^k f(\tau_{0:k};\theta^1\!\!,\theta^2) D_{\theta^i} (\log\pi(a_k^i|s_k;\theta^i)) A(s_k,a_k^1,a_k^2;\theta^1\!\!,\theta^2) d\tau ,\label{eq:CPGGradHessianAdv1}\\
\!\!\!\! D_{\theta^i \theta^j} \eta &=\!\!\!\int_{\tau} \!\sum_{k=0}^{|\tau|-1} \!\! \gamma^k f(\tau_{0:k};\theta^1\!\!,\theta^2)D_{\theta^i}( \log\pi(a_k^i|s_k;\theta^i)) D_{\theta^j}( \log\pi(a_k^j|s_k;\theta^j))^{\top} \!\! A(s_k,a_k^1,a_k^2;\theta^1\!\!,\theta^2) d\tau \nonumber \\
&\!\!\!\!\!\!\!\!\!\!\!\!\!+\!\!\! \int_{\tau} \!\sum_{k=1}^{|\tau|-1} \!\! \gamma^k \! f(\tau_{0:k};\theta^1\!\!,\theta^2) D_{\theta^i} (\log \!\prod_{l=0}^{k-1}\!\!\pi(a_{l}^i|s_l;\theta^i)) D_{\theta^j}(\log \pi(a_{k}^j|s_{k};\theta^j))^\top \!\! A(s_k,a_{k}^1,a_{k}^2;\theta^1\!\!,\theta^2) d\tau \nonumber\\
&\!\!\!\!\!\!\!\!\!\!\!\!\!+\!\!\! \int_{\tau} \!\sum_{k=1}^{|\tau|-1} \!\!\! \gamma^k \! f(\tau_{0:k};\theta^1\!\!,\theta^2) D_{\theta^i} (\log \pi(a_{k}^i|s_{k};\theta^i)) D_{\theta^j} (\log \! \prod_{l=0}^{k-1}\!\!\pi(a_{l}^j|s_l;\theta^j))^\top \!\!\! A(s_k,a_{k}^1,a_{k}^2;\theta^1\!\!,\theta^2) d\tau  \!.\!\!\!\label{eq:CPGGradHessianAdv}
\end{align}
\end{theorem}
\begin{proof} The RHS of \eq~\ref{eq:CPGGradHessianAdv1} is, 
\begin{align}
&\int_{\tau} \sum_{k=0}^{|\tau|-1} \gamma^k f(\tau_{0:k};\theta^1\!\!,\theta^2) D_{\theta^i} (\log\pi(a_k^i|s_k;\theta^i)) A(s_k,a_k^1,a_k^2;\theta^1\!\!,\theta^2)d\tau \nonumber\\
&= \int_{\tau} \sum_{k=0}^{|\tau|-1} \gamma^k f(\tau_{0:k};\theta^1\!\!,\theta^2) D_{\theta^i} (\log\pi(a_k^i|s_k;\theta^i)) ( Q(s, a^1, a^2;\theta^1,\theta^2) - V(s;\theta^1,\theta^2))d\tau\nonumber\\
&= D_{\theta^i} \eta
- \int_{\tau} \sum_{k=0}^{|\tau|-1} \gamma^k f(\tau_{0:k};\theta^1\!\!,\theta^2) D_{\theta^i} (\log\pi(a_k^i|s_k;\theta^i)) V(s;\theta^1,\theta^2))d\tau \nonumber\\
&= D_{\theta^i} \eta
- \int_{\tau} \sum_{k=0}^{|\tau|-1} \gamma^k f(\tau_{0:k-1},s_k;\theta^1,\theta^2) D_{\theta^i} (\pi(a_k^i|s_k;\theta^i))\pi(a_k^j|s_k;\theta^j) V(s;\theta^1,\theta^2))d\tau\\
&=D_{\theta^i}\eta. \nonumber
\end{align}
The last equality holds since the value function $V(s;\theta^1,\theta^2)$ is independent of the actions, and this concludes the proof for $D_{\theta^i}\eta$.
Now consider, the RHS of \eq~\ref{eq:CPGGradHessianAdv},

\begin{align}
&\!\!\int_{\tau} \!\sum_{k=0}^{|\tau|-1} \gamma^k f(\tau_{0:k};\theta^1\!\!,\theta^2)D_{\theta^i}( \log\pi(a_k^i|s_k;\theta^i)) D_{\theta^j}( \log\pi(a_k^j|s_k;\theta^j))^{\top}  A(s_k,a_k^1,a_k^2;\theta^1\!\!,\theta^2) d\tau \\
&\!\!+\!\!\! \int_{\tau} \!\sum_{k=1}^{|\tau|-1} \gamma^k f(\tau_{0:k};\theta^1\!\!,\theta^2) D_{\theta^i} (\log \prod_{l=0}^{k-1}\pi(a_{l}^i|s_l;\theta^i)) D_{\theta^j}(\log \pi(a_{k}^j|s_{k};\theta^j))^\top  A(s_k,a_{k}^1,a_{k}^2;\theta^1\!\!,\theta^2) d\tau \nonumber\\
&\!\!+\!\!\! \int_{\tau} \!\sum_{k=1}^{|\tau|-1} \gamma^k f(\tau_{0:k};\theta^1\!\!,\theta^2) D_{\theta^i} (\log \pi(a_{k}^i|s_{k};\theta^i)) D_{\theta^j} (\log \prod_{l=0}^{k-1}\pi(a_{l}^j|s_l;\theta^j))^\top  A(s_k,a_{k}^1,a_{k}^2;\theta^1\!\!,\theta^2) d\tau  \!.\!\!  \nonumber\\
\shortintertext{Using definition of Advantage function, we can equivalently write, }
&\!\!\int_{\tau} \!\! \sum_{k=0}^{|\tau|-1} \!\!\!\!\gamma^k \!f(\!\tau_{0:k};\!\theta^1\!\!,\theta^2)D_{\theta^i}( \log\pi(\!a_k^i|s_k;\!\theta^i\!)\!) D_{\theta^j}( \log\pi(\!a_k^j|s_k;\!\theta^j\!)\!)^{\!\top} \!\! (\!Q(\!s_k,\!a_k^1,\!a_k^2;\!\theta^1\!\!,\theta^2) \!-\! V\!(s_k;\!\theta^1\!\!,\theta^2\!)\!) d\tau \nonumber \\
&\!\!+\!\!\! \int_{\tau} \!\! \sum_{k=1}^{|\tau|-1} \!\!\!\! \gamma^k \!f(\!\tau_{0:k};\!\theta^1\!\!,\theta^2) D_{\theta^i} (\log \!\! \prod_{l=0}^{k-1}\!\!\pi(\!a_{l}^i|s_l;\!\theta^i\!)\!) D_{\theta^j}(\log \pi(\!a_{k}^j|s_{k};\!\theta^j\!)\!)^{\!\top} \!\! (\!Q(\!s_k,\!a_k^1,\!a_k^2;\!\theta^1\!\!,\theta^2) \!-\! V\!(\!s_k;\!\theta^1\!\!,\theta^2\!)\!)\! d\tau \nonumber\\
&\!\!+\!\!\! \int_{\tau} \!\!\sum_{k=1}^{|\tau|-1} \!\!\!\! \gamma^k \! f(\!\tau_{0:k};\!\theta^1\!\!,\theta^2) D_{\theta^i} (\log \pi(\!a_{k}^i|s_{k};\!\theta^i\!)\!) D_{\theta^j} (\log \!\! \prod_{l=0}^{k-1}\!\!\pi(\!a_{l}^j|s_l;\!\theta^j\!)\!)^{\!\top}  \!\! (\!Q(\!s_k,\!a_k^1,\!a_k^2;\!\theta^1\!\!,\theta^2) \!-\! V\!(\!s_k;\!\theta^1\!\!,\theta^2\!)\!) d\tau  \!.\!\! \nonumber\\
\shortintertext{Using \eq~\ref{eq:CPGGradHessianVal}, we can equivalently separate the expression in $D_{\theta^i \theta^j} \eta$ and terms depending on the value function,}
&\!\! D_{\theta^i \theta^j} \eta - \!\!\int_{\tau} \!\sum_{k=0}^{|\tau|-1} \gamma^k f(\tau_{0:k};\theta^1\!\!,\theta^2)D_{\theta^i}( \log\pi(a_k^i|s_k;\theta^i)) D_{\theta^j}( \log\pi(a_k^j|s_k;\theta^j))^{\top} V(s_k;\theta^1,\theta^2) d\tau \nonumber \\
&\!\!-\!\! \int_{\tau} \!\sum_{k=1}^{|\tau|-1} \!\! \gamma^k f(\tau_{0:k};\theta^1\!\!,\theta^2) D_{\theta^i} (\log \prod_{l=0}^{k-1}\pi(a_{l}^i|s_l;\theta^i)) D_{\theta^j}(\log \pi(a_{k}^j|s_{k};\theta^j))^\top V(s_k;\theta^1,\theta^2) d\tau \nonumber\\
&\!\!-\!\! \int_{\tau} \!\sum_{k=1}^{|\tau|-1} \!\! \gamma^k f(\tau_{0:k};\theta^1\!\!,\theta^2) D_{\theta^i} (\log \pi(a_{k}^i|s_{k};\theta^i)) D_{\theta^j} (\log \prod_{l=0}^{k-1}\pi(a_{l}^j|s_l;\theta^j))^\top V(s_k;\theta^1,\theta^2) d\tau  \!,\!\!\\
\shortintertext{note that for any $p_{\theta}(\tau) \neq 0$ using $D_{\theta} \log p_{\theta}(\tau) = D_{\theta} p_{\theta}(\tau)/p_{\theta}(\tau)$, which allows us to further simplify the expression to,}
&\!\! D_{\theta^i \theta^j} \eta - \!\!\int_{\tau} \!\sum_{k=0}^{|\tau|-1} \gamma^k f(\tau_{0:k-1},s_k;\theta^1,\theta^2)D_{\theta^i}(\pi(a_k^i|s_k;\theta^i)) D_{\theta^j}( \pi(a_k^j|s_k;\theta^j))^{\top} V(s_k;\theta^1,\theta^2) d\tau \nonumber \\
&\!\!\!\!-\!\! \int_{\tau}\! \sum_{k=1}^{|\tau|-1}\!\!\gamma^k\!f(\tau_{0:k-1},s_k, a^i_k;\theta^1,\theta^2) D_{\theta^i} (\log \prod_{l=0}^{k-1}\pi(a_{l}^i|s_l;\theta^i)) D_{\theta^j}( \pi(a_{k}^j|s_{k};\theta^j))^\top V(s_k;\theta^1,\theta^2) d\tau \nonumber\\
&\!\!\!\!-\!\! \int_{\tau}\! \sum_{k=1}^{|\tau|-1}\!\!\!\gamma^k f(\tau_{0:k-1},s_k, a^j_k;\theta^1,\theta^2) D_{\theta^i} (\pi(a_{k}^i|s_{k};\theta^i)) D_{\theta^j} (\log \prod_{l=0}^{k-1}\pi(a_{l}^j|s_l;\theta^j))^\top V(s_k;\theta^1,\theta^2) d\tau  \!\!\! \nonumber\\
&\!\!= D_{\theta^i \theta^j} \eta , \;\;\;\;\;\;\;\;\;\;\;\;\;\;
\end{align}
where the last equality holds since $V(s_k;\theta^1,\theta^2)$ is independent of the actions of any player. This concludes our proof.
\end{proof}
\subsection{Approaches to estimate advantage function}
There exist multiple approaches to estimate $A(s_k,a^1_k,a^2_k; \theta^1,\theta^2)$, for example
Monte Carlo (\eq~\ref{eqn: MonteCarloAdv}), 1-step TD residual (\eq~\ref{eqn: step1Adv}) and t-step TD residual (\eq~\ref{eqn: steptAdv})
\begin{align}
    A(s_k,a^1_k,a^2_k; \theta^1,\theta^2) =& \sum_{j=k}^{|\tau|-1} \gamma^{j-k} r(s_j,a^1_j,a^2_j) - V(s_k;\theta^1,\theta^2), \label{eqn: MonteCarloAdv}\\
     A(s_k,a^1_k,a^2_k; \theta^1,\theta^2) =& r(s_k,a^1_k,a^2_k) + \gamma V(s_{k+1};\theta^1,\theta^2) - V(s_k;\theta^1,\theta^2),\label{eqn: step1Adv}\\
     A(s_k,a^1_k,a^2_k; \theta^1,\theta^2) =& r(\!s_k,\!a^1_k,\!a^2_k) \!+\! \gamma r(\!s_{k+1},\!a^1_{k+1},\!a^2_{k+1}) \!+\!...\!+\! \gamma^{t-1}r(\!s_{k+t-1},\!a^1_{k+t-1},\!a^2_{k+t-1})\nonumber\\ 
     &+ \gamma^t V(\!s_{k+t};\!\theta^1\!,\theta^2) - V(\!s_k;\!\theta^1\!,\theta^2).\label{eqn: steptAdv}
\end{align}
The Monte Carlo approach experience high variance in gradient estimation, particularly with long trajectories, which may result in slow learning or not learning at all. On the other hand 1-step TD residual has bias towards initialization of the value function. t-step TD residual can lower this effect up to an extent. To trade between bias and variance we propose to use generalized advantage estimation (\GAE) ~\citep{schulman2015highdimensional}, which is an exponentially $\lambda$ weighted average over t-step estimators. It is given by,
\begin{align}
    A^{\GAE}(s_k,a^1_k,a^2_k; \theta^1,\theta^2) &= \sum_{l=0}^{|\tau|-1} (\lambda\gamma)^l\delta^V_{l+k}, \nonumber\\
    \text{where, } \delta^V_{k} &= r(s_k,a^1_k,a^2_k) + \gamma V(s_{k+1};\theta^1,\theta^2) - V(s_k;\theta^1,\theta^2). \nonumber
\end{align}{}
Note that at $\lambda = 0$, $A^{\GAE}(s_k,a^1_k,a^2_k; \theta^1,\theta^2)$ reduces to 1-step TD residual and at $\lambda = 1$ it is equivalent to the Monte Carlo approach to estimate the advantage function. We refer the interested reader to ~\citep{schulman2015highdimensional} for more details.

\section{Training by self-play}\label{appx: selfplay}
As proposed in \alg~\ref{appxalg: cpg}, we use \cpg to learn a policy for each player, however, it is also possible to use the competitive policy gradient algorithm to learn game strategies by self-play.
In contrast to two player training where each player has its own policy model, in self-play, we have one model for both players. Each player then samples actions from the same policy model for two different state instances
$s^{1}=[s^{l1}, s^{l2}]$ and $s^{2}=[s^{l2}, s^{l1}]$, where $s^{l1}$ and $s^{l2}$ are the local states of player 1 and player 2 respectively. Let $\tau = \big( (s^{l1}_k$,$s^{l2}_k,a^1_{k},a^2_{k},r_{k})_{k=0}^{|\tau^1|-1} ,s_{|\tau|}\big)$ be a trajectory obtained in the game. Using \eq~\ref{eq:CPGGradHessianVal1} and \eq~\ref{eq:CPGGradHessianVal}, we can define the gradient and bilinear terms as $D_{\theta}\eta^i$ and $D_{\theta \theta} \eta^i$ , $i,j \in \lbrace 1,2 \rbrace$, $i \neq j$,
\begin{align}
\!\!\!\!\!\!\!\!\!\!\!\!\!\!\!\!\!\!\!\!\!\!\!\!\!\!\!D_{\theta} \eta^i =&\int_{\tau} \sum_{k=0}^{|\tau|-1} \gamma^k f(\tau_{0:k};\theta) D_{\theta} (\log\pi(a_k^i|s_k^i;\theta)) Q(s_k^1,a_k^1,a_k^2;\theta) d\tau,\!\!\!\!\! \label{eq: selfplayGrad}\\
\!\! D_{\theta \theta} \eta^i =& \int_{\tau} \!\sum_{k=0}^{|\tau|-1}\!\!\! \gamma^k f(\tau_{0:k};\theta) D_{\theta}( \log\pi(a_k^i|s_k^i;\theta)) D_{\theta}( \log\pi(a_k^j|s_k^j;\theta))^{\top}  Q(s_k^1,a_k^1,a_k^2;\theta) d\tau \nonumber \\
&+\!\! \int_{\tau}\!\!\sum_{k=1}^{|\tau|-1}\!\!\!\gamma^k \!f(\tau_{0:k};\theta) D_{\theta} (\log\!\! \prod_{l=0}^{k-1}\!\!\pi(a_{l}^i|s_l^i;\theta)) D_{\theta}(\log \pi(a_{k}^j|s_{k}^j;\theta))\!^\top \! Q(s_k^1,a_{k}^1,a_{k}^2;\theta) d\tau \nonumber\\
&+\!\! \int_{\tau}\!\! \sum_{k=1}^{|\tau|-1}\!\!\!  \gamma^k  \!f(\tau_{0:k};\theta) D_{\theta} (\log \pi(a_{k}^i|s_{k}^i;\theta)) D_{\theta} (\log\!\! \prod_{l=0}^{k-1}\!\!\pi(a_{l}^j|s_l^j;\theta))\!^\top \! Q(s_k^1,a_{k}^1,a_{k}^2;\theta) d\tau.\label{eq: selfplayHessian}
\end{align}

Utilizing the gradients and bilinear term in \eq~\ref{eq: selfplayGrad} and \eq~\ref{eq: selfplayHessian}, the parameters are updated twice which results in the following update rule for self-play, 
\begin{align}\label{eqn: selfplayCGDSolution}
\theta^t &\leftarrow \theta + \alpha \bigl( Id + \alpha^2 D_{\theta\theta} \eta^1 D_{\theta\theta}\eta^2 \bigr)^{-1} \bigl( D_{\theta} \eta^1 - \alpha D_{\theta\theta}\eta^1 D_{\theta}\eta^2 \bigr) , \nonumber \\
\theta &\leftarrow \theta^t - \alpha \bigl( Id + \alpha^2 D_{\theta\theta}\eta^2 D_{\theta \theta}\eta^1 \bigr)^{-1} \bigl( D_{\theta}\eta^2 + \alpha D_{\theta\theta}\eta^2 D_{\theta} \eta^1 \bigr) ,
\end{align}
where $\theta^t$ is a temporary intermediate variable.\footnote{Of course, there are other ways to come up with self-play updates. For example, one can use \eq~\ref{eqn: CGDSolution} with $\theta^1=\theta^2=\theta$ to update parameters, then set $\theta\leftarrow(\theta^1+\theta^2)/2$.}

\begin{algorithm}[h!]
\SetAlgoLined
\textbf{Input:} $\pi(.|.;\theta)$, N, B, $\alpha$, Choice of $A(s^1,a^1,a^2;\theta)$ mentioned in Sec~\ref{sec:cpg_theory}\\ 
\For{$epoch:e=1$ {\bfseries to} $N$}{
 \For{$batch:b=1$ {\bfseries to} $B$}{
  \For{$k=0$ {\bfseries to} $|\tau|-1$}{
  Sample $a^1_k \sim \pi(a^1_k|s^{1}_k;\theta)$, $a^2_k \sim \pi(a^2_k|s^{2}_k;\theta)$.\\
  Execute actions $a_k = (a_k^1,a_k^2)$ \\
  Record $\lbrace s^{l1}_k,s^{l2}_k, a^1_k, a^2_k, r_k \rbrace$ in $\tau_{b}$
  }
  Record $\tau_b$ in \textit{M}
  }
\textbf{1.} Estimate $A(s^1,a^1,a^2;\theta)$ with $\tau$ in M\\
\textbf{2.} Estimate $D_{\theta} \eta^1 $, \, $D_{\theta} \eta^2$, \, $D_{\theta\theta} \eta^1$, \, $D_{\theta\theta} \eta^2$  using \eq~\ref{eq: selfplayGrad} and \eq~\ref{eq: selfplayHessian}\\
 \textbf{3.} Update $\theta$ using \eq~\ref{eqn: selfplayCGDSolution}
}
\caption{Competitive Policy Gradient with self-play}
\label{appxalg: selfplaycpg}
\end{algorithm}

We tested the self-play algorithm on the Markov soccer game, which is discussed in more detail in \sect~\ref{appxsec: soccer}.
We focus in this comparison on the Markov soccer game since it is strategic enough while being representative. The game has two player A and B. The state vector in perspective of player A is $s^1 = [s^A,s^B]$ and in player B $s^2 = [s^B,s^A]$, in accordance with the self-play setting described above. The game description is given in \sect~\ref{appxsec: soccer}. The player was trained while competing with itself using \alg~\ref{appxalg: selfplaycpg}. \fig~\ref{fig: cpgsp_cpgsp_interaction_plot} below show the interaction plot for \cpg agents trained with self-play with both players winning almost 50\% of the games. On competing agents trained under self-play (\cpgsp) against agents trained with two different networks (\cpg), we observe that self-play had slight edge in number of game wins. The is probably due to better generalisation of agents in self-play obtained by exploiting symmetricity in the game. We also trained a \gda agent using self-play and saw no clear improvements compared to normal \gda as tested in \sect~\ref{appxsec: soccer}. 


\begin{figure}[!h]
\hspace{-4.00mm}
	\centering
    \begin{subfigure}[t]{0.30\columnwidth}
  	\centering
  	\includegraphics[width=\linewidth]{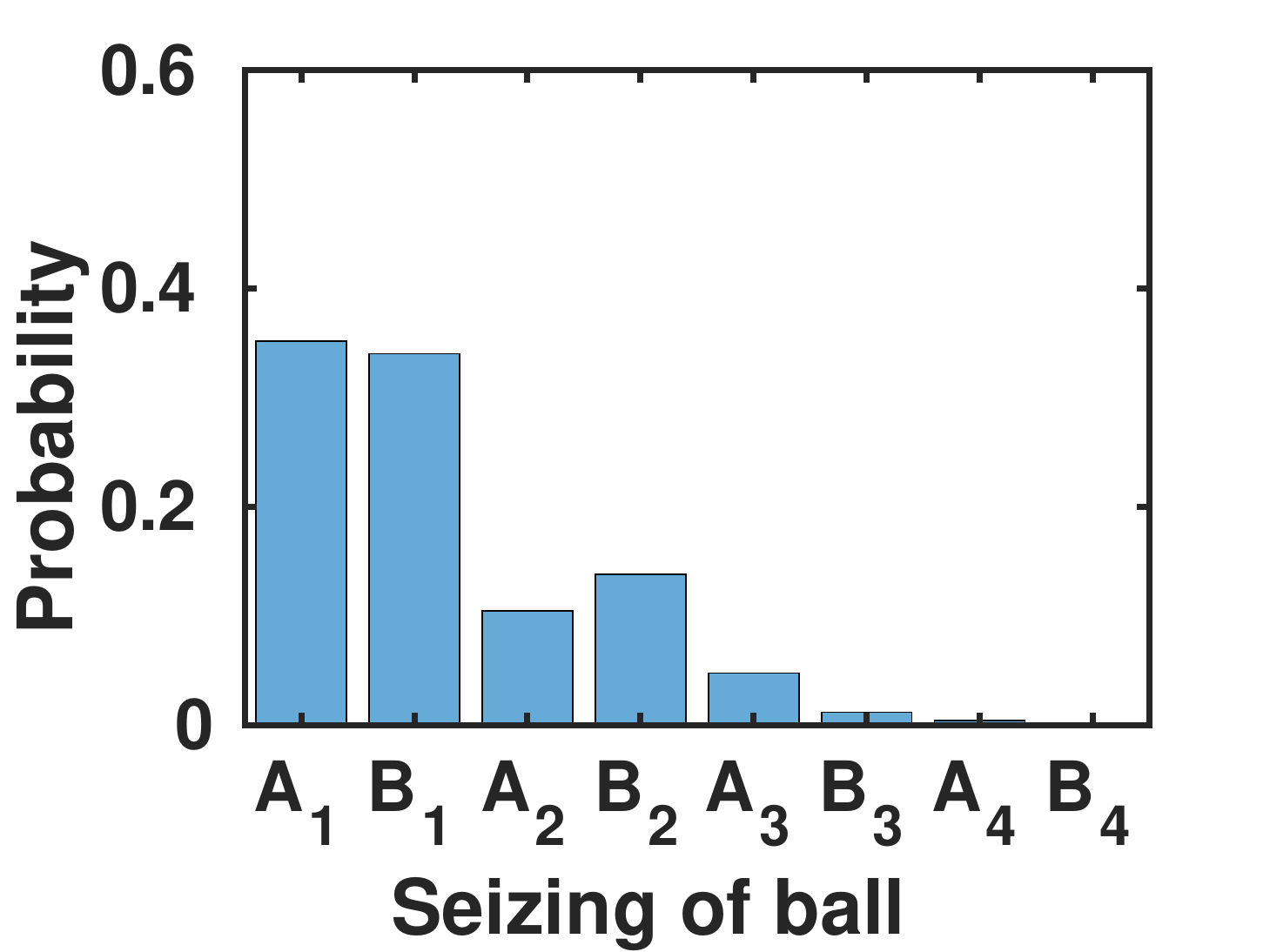}
    \caption{\cpgsp vs \cpgsp}
    \label{fig: cpgsp_cpgsp_interaction_plot}
    \end{subfigure}
    \hspace{-5.00mm}
  ~
    \begin{subfigure}[t]{0.30\columnwidth}
  	\centering
  	\includegraphics[width=\linewidth]{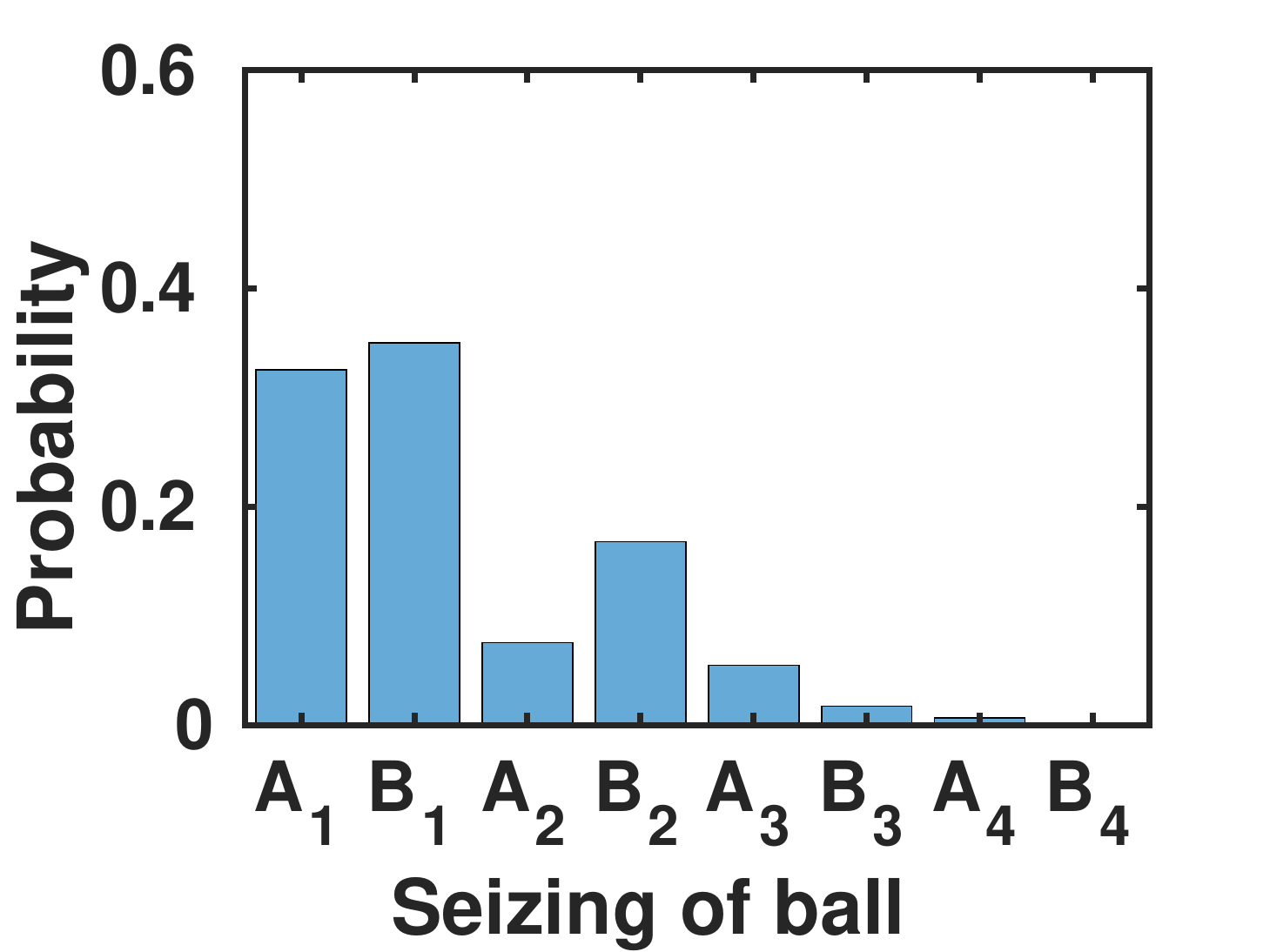}
    \caption{\cpg vs \cpgsp}
    \label{fig: cpg_cpgsp_interaction_plot}
    \end{subfigure}
    \hspace{-5.00mm}
~
  \begin{subfigure}[t]{0.30\columnwidth}
  	\centering
  	\includegraphics[width=\linewidth]{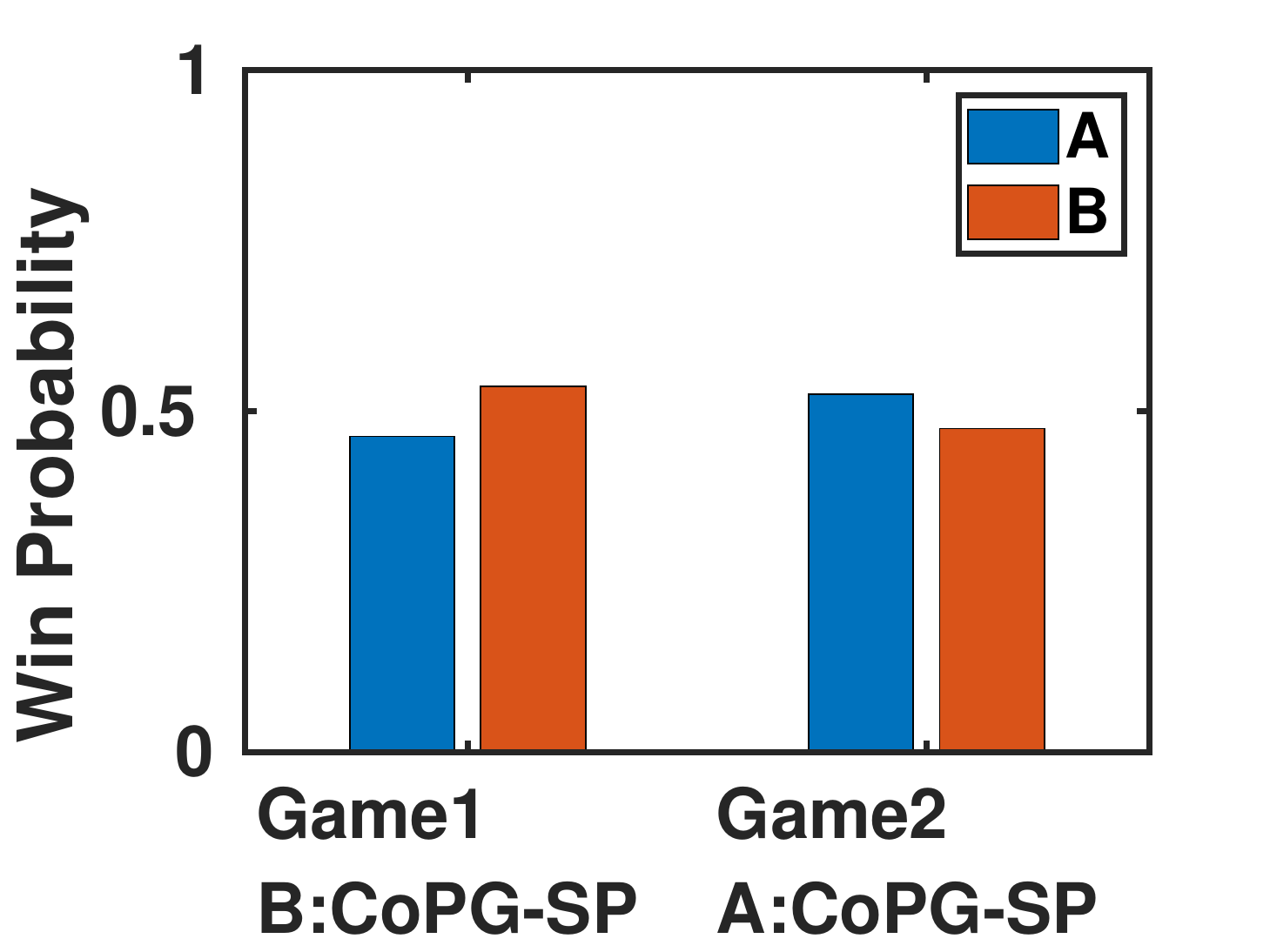}  
    \caption{\cpg vs \cpgsp}
    \label{fig: EX_cpg_cpgsp_winning_plot}
  \end{subfigure}
  \hspace{-4.00mm}
  \caption{ a-b) Interaction plots, representing probability of seizing ball in the game between $A$ vs $B$. X-axis convention, for player $A$. $A_1$: $A$ scored goal, $A_2$: $A$ scored goal after seizing ball from $B$, $A_3$: $A$ scored goal by seizing ball from $B$ which took the ball from $A$ and so forth. Vice versa for player $B$. N: No one scored goal both kept on seizing ball. c) probability of games won. 
  }
  \label{Fig: InteractionBilinear_selfplay}
\end{figure}

\section{Trust region competitive policy optimisation}
Following our discussion in the \sect~\ref{sec:trcpo} on trust region competitive policy optimization, in this section we will go into more details on constraint competitive optimization and formally derive and proof all our results presented in \sect~\ref{sec:trcpo}. More precisely in this section we first proof \lem~\ref{lemma: trcpo_advantage} and \thm~\ref{theorem: BoundTRCPO}, second we will derive how we can efficiently approximate \eq~\ref{eqn: TRCPOFinalObjective} using a bilinear approximation of the objective and a second order approximation of the KL divergence constraint and how this allows us to efficiently find an approximate solution. 

\subsection{Proof for \lem~\ref{lemma: trcpo_advantage}}
\label{appx: trcpo_advantage}
\begin{proof}
The second term on the RHS of \eq~\ref{eqn: TRCPOAdvantage} can be simplified as,
\begin{align}
    \mathbb{E}_{\tau \sim  f(.;\theta'^1,\theta'^2)}& \Bigl[\sum_{k=0}^{|\tau|-1} \gamma^k A(s_k, a^1_k, a^2_k;\theta_{}^1,\theta_{}^2) \Bigr] \nonumber\\
    &= \mathbb{E}_{\tau \sim   f(.;\theta'^1,\theta'^2)} \Bigl[ \sum_{k=0}^{|\tau|-1} \gamma^k \bigl( Q(s_k, a^1_k,a^2_k;\theta_{}^1,\theta_{}^2) - V(s_k;\theta_{}^1,\theta_{}^2) \bigr) \Bigr] \nonumber\\
    &= 
    \mathbb{E}_{\tau \sim   f(.;\theta'^1,\theta'^2)} \Bigl[ \sum_{k=0}^{|\tau|-1} \gamma^k \bigl( r(s_k,a^1_k,a^2_k) + \gamma V(s_{k+1};\theta_{}^1,\theta_{}^2) - V(s_k;\theta_{}^1,\theta_{}^2) \bigr) \Bigr]\nonumber\\
    &= 
    \mathbb{E}_{\tau \sim   f(.;\theta'^1,\theta'^2)} \Bigl[ \sum_{k=0}^{|\tau|-1} \gamma^k r(s_k,a^1_k,a^2_k) + \gamma^{k+1} V(s_{k+1};\theta_{}^1,\theta_{}^2) - \gamma^k V(s_k;\theta_{}^1,\theta_{}^2) \Bigr]\nonumber\\
    &= 
    \mathbb{E}_{\tau \sim   f(.;\theta'^1,\theta'^2)} \Bigl[ \sum_{k=0}^{|\tau|-1} \gamma^k r(s_k,a^1_k,a^2_k) \Bigr] - \mathbb{E}_{s_0} \Bigl[V^1(s_0;\theta_{}^1,\theta_{}^2)\Bigr]\nonumber\\
    &= \eta(\theta'^1,\theta'^2) - \eta(\theta_{}^1,\theta_{}^2).
\end{align}
This concludes our proof that,
\begin{align}
\eta(\theta'^1,\theta'^2) = \eta(\theta_{}^1,\theta_{}^2) +
\mathbb{E}_{\tau \sim  f(\cdot;\theta'^1,\theta'^2)}\sum\nolimits_{k=0}^{|\tau|-1} \gamma^k A(s,a^1,a^2;\theta_{}^1,\theta_{}^2). \nonumber
\end{align}
\end{proof}{}

\subsection{Proof for \thm~\ref{theorem: BoundTRCPO}}
\label{appx: ProofBoundTRPO}
\begin{proof}
Let $\theta_{}^1, \theta_{}^2$ be the parametrization of the policies, when these parameters are updated to $\theta'^1, \theta'^2$, by \lem~\ref{lemma: trcpo_advantage} we know that the improved game objective $\eta(\theta'^1,\theta'^2)$ is, 
\begin{align}
\label{eqn: appxTrueGO}
    \eta(\theta'^1,\theta'^2) = \eta(\theta_{}^1,\theta_{}^2) +
\mathbb{E}_{\tau \sim   f(.;\theta'^1,\theta'^2)}\sum_{k}^{|\tau|-1} \gamma^k A(s_k,a^1_k,a^2_k;\theta_{}^1,\theta_{}^2) .
\end{align}
Furthermore, we defined $L_{\theta_{}^1,\theta_{}^2}(\theta'^1,\theta'^2)$ which denotes the surrogate game objective where the trajectory depends on policies $(\theta^1,\theta^2)$ but actions are sampled from $ (\theta'^1,\theta'^2)$.
\begin{align}
\label{eqn: appxSurrGO}
L_{\theta^1\!\!,\theta^2}(\theta'^1\!\!,\theta'^2)\!\!= \!\eta(\theta_{}^1\!\!,\theta_{}^2)\! +\!
\mathbb{E}_{\tau \sim  f(\cdot;\theta^1\!\!,\theta^2)} \!\!\! \sum_{k=0}^{|\tau|-1} \!\!\! \gamma^k \mathbb{E}_{ \pi(a'^1_k|s_k;\theta'^1),\pi(a'^2_k|s_k;\theta'^2)} A(s_k,a'^1_k,a'^2_k;\theta_{}^1\!\!,\theta_{}^2) .\!\!
\end{align}
Note that $L_{\theta_{}^1,\theta_{}^2}(\theta'^1,\theta'^2)$ is an approximation of true game objective $\eta(\theta'^1,\theta'^2)$. Using the average advantage function,
\begin{align}
\overline{A}(s;\theta_{}^1,\theta_{}^2) = \sum_{a^1,a^2}  \pi(a^1|s;\theta'^1)\pi(a^2|s;\theta'^2)A(s,a^1,a^2;\theta_{}^1,\theta_{}^2), 
\end{align}
one can write \eq~\ref{eqn: appxTrueGO} and \eq~\ref{eqn: appxSurrGO} as,
\begin{align}\label{eqn: appxTrueGOAvgerage}
\eta(\theta'^1,\theta'^2)-\eta(\theta_{}^1,\theta_{}^2) =
\mathbb{E}_{\tau \sim  f(.;\theta'^1,\theta'^2)}\Bigl[\sum_{k=0}^{|\tau|-1} \gamma^k \overline{A}(s;\theta_{}^1,\theta_{}^2) \Bigr], \\\label{eqn: appxSurrGOAvgerage}
L_{\theta_{}^1,\theta_{}^2}(\theta'^1,\theta'^2)-\eta(\theta_{}^1,\theta_{}^2) =
\mathbb{E}_{\tau \sim  f(.;\theta_{}^1,\theta_{}^2)}\Bigl[\sum_{k=0}^{|\tau|-1} \gamma^k \overline{A}(s;\theta_{}^1,\theta_{}^2) \Bigr].
\end{align}
To finish our proof, we subtract \eq~\ref{eqn: appxSurrGOAvgerage} from \eq~\ref{eqn: appxTrueGOAvgerage} and take the absolute value, this allows to bound $|\eta(\theta'^1\!\!,\theta'^2)\!-\!L_{\theta_{}^1\!\!,\theta_{}^2}(\theta'^1\!\!,\theta'^2)|$. To derive this bound let $\epsilon = \max_{s} \sum_{k}^{|\tau|-1} \gamma^k \overline{A}(s;\theta^1,\theta^2)$, which allows us to perform the following steps,
\begin{align}
\!\!\!\!\!\!|\eta(\theta'^1\!\!,\theta'^2)\!-\!L_{\theta_{}^1\!\!,\theta_{}^2}(\theta'^1\!\!,\theta'^2)|\! &=\! \Bigl|
\mathbb{E}_{\tau \sim  f(\!.;\theta'^1\!\!,\theta'^2)}\Bigl[\sum_{k}^{|\tau|-1} \!\!\! \gamma^k \overline{A}(\!s;\theta_{}^1\!\!,\theta_{}^2) \Bigr] \!-\! 
\mathbb{E}_{\tau \sim  f(\!.;\theta_{}^1\!\!,\theta_{}^2)}\Bigl[\sum_{k}^{|\tau|-1} \!\!\! \gamma^k \overline{A}(\!s;\theta_{}^1\!\!,\theta_{}^2) \Bigr] \Bigr| \nonumber\\ &=\int_{\tau}\Bigl|f(\tau;\theta'^1,\theta'^2)-f(\tau;\theta_{}^1,\theta_{}^2)\Bigr|\Bigl[\sum_{k}^{|\tau|-1} \gamma^k \overline{A}(s;\theta_{}^1,\theta_{}^2) \Bigr] d\tau \nonumber\\
&\leq \epsilon D_{TV}\bigl(\tau \sim(\theta_{}^1,\theta_{}^2),\tau \sim(\theta'^1,\theta'^2)\bigr)\nonumber\\
&\leq \epsilon  \sqrt{\frac{1}{2}D_{KL}(\tau \sim(\theta_{}^1,\theta_{}^2),\tau \sim(\theta'^1,\theta'^2))}\nonumber\\
&= \epsilon/\sqrt{2} \sqrt{D_{KL}((\theta'^1,\theta'^2), (\theta_{}^1,\theta_{}^2))}.\nonumber
\end{align}
This concludes our proof.
\end{proof}{} 

\subsection{Constraint Competitive Optimization of a Trust Region}
\label{appx: trcpoOptimization}
In this section, we approach the constrained optimization problem of \trcpo given by \eq~\ref{eqn: TRCPOFinalObjective} in 3 steps. First, motivated by \cgd we seek for a Nash equilibrium of the bilinear approximation of the game objective. Secondly, in contrast to \cpg, where an L2 penalty term is used along with the bilinear approximation denoting limited confidence in parameter deviation, here a KL divergence constraint will influence the gradient direction. Note that we incorporate this KL divergence constraint into the objective using the method of Lagrange multipliers. This results in the game theoretic update rule defined in \eq~\ref{eqn: trcpo_game}. 
\begin{align}\label{eqn: trcpo_game}
    &\theta^1\leftarrow\theta^1 \!\!+\!\!\!\!\argmax_{{\Delta\theta^1};{\Delta\theta^1}+\theta^1\in \Theta^1} \!\!\!\! {\Delta\theta^1}^{\!\!\top}\!\!D_{\theta^1}L \!+\! {\Delta\theta^1}^{\!\!\top} \!\! D_{\theta^1\theta^2}L \Delta \theta^2 \!-\! \frac{\lambda}{2}({\Delta\theta^1}^{\!\!\top} \!\! A_{11} {\Delta\theta^1} \!+\! \Delta {\theta^2}^{\!\!\top} \!\! A_{22} \Delta \theta^2 \!-\! 2 \delta'), \nonumber\\
    &\theta^2\leftarrow\theta^2 \!\!+\!\!\!\! \argmin_{{\Delta\theta^2};{\Delta\theta^2} + \theta^2\in \Theta^2}\!\!\!\! {\Delta\theta^2}^{\!\!\top} \!\! D_{\theta^2}L \!+\! {\Delta\theta^2}^{\!\!\top} \!\! D_{\theta^2\theta^1}L \Delta \theta^1 \!+\! \frac{\lambda}{2}(\Delta {\theta^1}^{\!\!\top} \!\! A_{11} \Delta \theta^1 \!+\! \Delta {\theta^2}^{\!\!\top} \!\! A_{22} \Delta \theta^2 \!-\! 2 \delta').
\end{align}
Here, $\lambda$ is the Lagrange multiplier. Note that the same $\lambda$ is used for both the agents and that we use a quadratic approximation of the KL divergence, which is $D_{KL}\bigl( (\theta_{}^1,\theta_{}^2),(\theta'^1,\theta'^2)\bigr) \approx  \frac{1}{2} \sum_{i}^2\sum_{j}^2 {\Delta\theta^i}^\top A_{ij} {\Delta\theta^j}$, where $A_{ij} = D_{\theta^i\theta^j} D_{KL}\bigl( (\theta_{}^1,\theta_{}^2),(\theta'^1,\theta'^2)\bigr)$, and it holds that for $i \neq j, \ A_{ij} = 0$. Note that we will go into more details later on the approximation of the KL divergence.

\textbf{\trcpo update:} As said the update step in \eq~\ref{eqn: trcpo_game} is given by the Nash equilibrium of the game within \eq~\ref{eqn: trcpo_game}. We can find the Nash equilibrium that fulfills the approximated KL divergence constraint efficiently by solving the following equation using a line-search approach, 
 \begin{flalign}\label{eqn: TRCO}
\Delta \theta = -(B + \lambda A')^{-1}C, \, \, \, 
     \text{subject to:} \, {\Delta \theta}^\top A \Delta \theta \leq \delta ,
 \end{flalign}
where, let $A'$, $B$, $C$, $\delta$, and $\Delta \theta$ be defined as,
\begin{align}
&A' = \begin{bmatrix}
-A_{11} & 0\\
 0 & A_{22}\\
\end{bmatrix},\; B = \begin{bmatrix}
0 & D_{\theta^1\theta^2}L\\
 D_{\theta^2\theta^1}L & 0\\
\end{bmatrix},
\; \nonumber \\
&C =  \begin{bmatrix}
 D_{\theta^1}L\\
 D_{\theta^2}L
\end{bmatrix}, \;\delta=2\delta' \text{, and } \Delta \theta = \begin{bmatrix}
\Delta \theta^1 \\
 \Delta \theta^2 \\
\end{bmatrix}\,. \label{eqn: nash_notation}
\end{align}

\textbf{\trcpo constraint optimization:} After this quick introduction on how to practically update the policy parameters using \trcpo, let us derive in detail how to get to \eq~\ref{eqn: TRCO}. As explained above \eq~\ref{eqn: TRCO} is the Nash equilibrium of an approximation of \eq~\ref{eqn: TRCPOFinalObjective}. Before starting to approximate \eq~\ref{eqn: TRCPOFinalObjective} let us rewrite it by using the definition of $L_{\theta^1\!\!,\theta^2}(\theta'^1\!\!,\theta'^2)$ as given in \eq~\ref{eqn: appxSurrGO}. Note that we can neglect the constant term $\eta(\theta_{}^1\!\!,\theta_{}^2)$, which results in the following game,
\begin{align}
    \max_{\theta'^1} \min_{\theta'^2} \,\; &\mathbb{E}_{\tau \sim  f(\cdot;\theta^1\!\!,\theta^2)} \!\!\! \sum_{k=0}^{|\tau|-1} \!\!\! \gamma^k \mathbb{E}_{ \pi(a'^1_k|s_k;\theta'^1),\pi(a'^2_k|s_k;\theta'^2)} A(s_k,a'^1_k,a'^2_k;\theta_{}^1\!\!,\theta_{}^2) , \\
    \text{subject to: } &D_{KL}((\theta_{}^1,\theta_{}^2),(\theta'^1,\theta'^2)) \leq \delta'. \nonumber
\end{align}
Furthermore, we can use an importance sampling factor instead of considering the expected value with respect to the policies parametrized with $(\theta'^1,\theta'^2)$. This results in the following optimization problem, 
\begin{align} \label{eqn: importanceSamplingGame}
    \max_{\theta'^1} \min_{\theta'^2} \,\; &E_{\tau \sim f(\cdot;\theta^1,\theta^2)}  \sum_{k=0}^{|\tau|-1} \gamma^k \frac{\pi(a^1_k|s_k;\theta'^1)\pi(a^2_k|s_k;\theta'^2)}{\pi(a^1_k|s_k;\theta_{}^1)\pi(a^2_k|s_k;\theta_{}^2)}A(s_k,a^1_k,a^2_k;\theta_{}^1,\theta_{}^2),  \\
    \text{subject to: } &D_{KL}((\theta_{}^1,\theta_{}^2),(\theta'^1,\theta'^2)) \leq \delta'. \nonumber
\end{align}
For the game in \eq~\ref{eqn: importanceSamplingGame} we can compute the gradient $D_{\theta^i}L$ and bilinear term $D_{\theta^i\theta^j}L$ given player $i,j \in \lbrace 1,2\rbrace$, $i \neq j$ as
\begin{align}\label{eqn: GradHessianTRCPO}
    D_{\theta^i} L &= E_{\tau \sim f(\cdot;\theta^1,\theta^2)}  \sum_{k=0}^{|\tau|-1} \gamma^k \frac{D_{\theta^i}\pi(a^i_k|s_k;\theta^i)}{\pi(a^i_k|s_k;\theta^i)}A(s_k,a^1_k,a^2_k;\theta^1,\theta^2), \nonumber \\
    D_{\theta^i\theta^j} L &= E_{\tau \sim f(\cdot;\theta^1,\theta^2)} \sum_{k=0}^{|\tau|-1} \gamma^k \frac{D_{\theta^i}\pi(a^i_k|s_k;\theta^i)D_{\theta^j}\pi(a^j_k|s_k;\theta^j)}{\pi(a^i_k|s_k;\theta^i)\pi(a^j_k|s_k;\theta^j)}A(s_k,a^1_k,a^2_k;\theta^1,\theta^2) ,
\end{align}
which allows us to formulate the following bilinear approximation of the optimization problem, 
\begin{align}
    &\max_{{\Delta\theta^1};{\Delta\theta^1}+\theta^1\in \Theta^1} \min_{{\Delta\theta^2};{\Delta\theta^2}+\theta^2\in \Theta^2} {\Delta\theta^1}^{\top} \!\!D_{\theta^1}L + {\Delta\theta^1}^{\!\!\top} \!\! D_{\theta^1\theta^2}L \Delta \theta^2
    + {\Delta\theta^2}^{\top} \!\!D_{\theta^2}L + {\Delta\theta^2}^{\!\!\top} \!\! D_{\theta^2\theta^1}L \Delta \theta^1 , \nonumber\\
    &\qquad \qquad \text{subject to: } \qquad \qquad  D_{KL}((\theta_{}^1,\theta_{}^2),(\theta'^1,\theta'^2)) \leq \delta' . \nonumber
\end{align}

However, to efficiently solve the optimization problem we also need to approximate the KL divergence for the two agent case, similar to~\cite{schulman2015trust} we use a quadratic approximation of the constraint. 

\textbf{Approximation of KL divergence}
\label{appx: KLdivergenceHessian}
Let us derive an approximation of the KL divergence for the two player case that can be estimated efficiently using Monte Carlo samples. As defined in \sect~\ref{sec:trcpo}, the KL divergence between $(\theta^1,\theta^2)$ and $(\theta'^1,\theta'^2)$ is given by,
\begin{align}\label{eqn: appxKLDivergence}
D_{KL}\bigl( (\theta_{}^1,\theta_{}^2),(\theta'^1,\theta'^2)\bigr) = \int_{\tau} f(\tau;\theta_{}^1,\theta_{}^2)\log\biggl(\frac{f(\tau;\theta_{}^1,\theta_{}^2)}{f(\tau;\theta'^1,\theta'^2)}\biggr) d\tau,
\end{align}
Let us define $\theta = \bigl({\theta^1}^\top,{\theta^2}^\top \bigr)^\top$\!\!, and $\Delta \theta = \theta' - \theta$, thus, using a Taylor series to approximate $D_{KL}$ around $\theta_{}$ we get,
\begin{align}\label{eq: taylorKL}
D_{KL}(\theta_{},\theta') &= D_{KL}(\theta_{},\theta')\Bigl|_{\theta' = \theta_{}} \!\! + \frac{\partial D_{KL}(\theta_{},\theta')}{\partial \theta'}\biggl|_{\theta' = \theta_{}} \!\! \!\! \!\! \Delta \theta_{} + \frac{1}{2}{\Delta \theta}^\top\!\! A\bigl|_{\theta' = \theta_{}}  \Delta \theta + \varepsilon\bigr(||\Delta \theta||^3 \bigl),
\end{align}
where,
\begin{align} \label{eqn: KLHession_A}
A =
\begin{bmatrix}
D_{\theta'^1\theta'^1} D_{KL}\bigl( (\theta_{}^1,\theta_{}^2),(\theta'^1,\theta'^2)\bigr)
& \!\!\!D_{\theta'^1\theta'^2}D_{KL}\bigl( (\theta_{}^1,\theta_{}^2),(\theta'^1,\theta'^2)\bigr)\vspace{0.5 mm}\\
D_{\theta'^2\theta'^1}D_{KL}\bigl( (\theta_{}^1,\theta_{}^2),(\theta'^1,\theta'^2)\bigr)
& \!\!\!D_{\theta'^2\theta'^2}D_{KL}\bigl( (\theta_{}^1,\theta_{}^2),(\theta'^1,\theta'^2)\bigr)
\end{bmatrix}.
\end{align}
In the Taylor expansion \eq~\ref{eq: taylorKL}, the zero order term $D_{KL}(\theta_{},\theta')|_{\theta' = \theta_{}} = 0$ and the first order term can be expanded as, 
\label{appx: KLDivergenceApprox}
\begin{gather} 
    \frac{\partial D_{KL}(\theta_{},\theta')}{\partial \theta'}\biggl|_{\theta' = \theta_{}} \!\!\! = \Bigl(\!D_{\theta'^1} D_{KL}\bigl( (\theta_{}^1\!\! ,\theta_{}^2),(\theta'^1\!\! ,\theta'^2)\bigr)^{\!\top}\!\! ,D_{\theta'^2} D_{KL}\bigl( (\theta_{}^1\!\! ,\theta_{}^2),(\theta'^1\!\!,\theta'^2)\bigr)^{\!\top} \Bigr)^\top \! \biggr|_{\substack{(\theta'^1\!\!, \theta'^2) = (\theta_{}^1\!\!, \theta_{}^2)}} .\nonumber
\end{gather}
When solving for $D_{\theta'^1} D_{KL}\bigl( (\theta_{}^1\!\! ,\theta_{}^2),(\theta'^1\!\! ,\theta'^2)\bigr)$, we get,
\begin{align}
    D_{\theta'^1} D_{KL}\bigl( (\theta_{}^1\!\!,\theta_{}^2),(\theta'^1,\theta'^2)\bigr)\Bigr|_{\theta' = \theta_{}}  \!\!\!\!\!&= D_{\theta'^1} \biggl(\int_{\tau} f(\tau;\theta_{}^1,\theta_{}^2)\log\Bigl(\frac{f(\tau;\theta_{}^1,\theta_{}^2)}{f(\tau;\theta'^1,\theta'^2)}\Bigr) d\tau \biggr) \biggr|_{\substack{(\theta'^1\!\!, \theta'^2) = (\theta_{}^1\!\!, \theta_{}^2)}}\nonumber\\
    &=\!\!\int_{\tau} \!\!D_{\theta'^1} \! \bigl(\! f(\!\tau;\!\theta_{}^1\!\!,\theta_{}^2) \! \log \! f(\!\tau;\!\theta_{}^1\!\!,\theta_{}^2) \!\!-\!\! f(\!\tau;\!\theta_{}^1\!\!,\theta_{}^2) \! \log \! f(\!\tau;\!\theta'^1\!\!,\theta'^2)  \! \bigr) d\tau\!
    \nonumber\\
    &= \int_{\tau} \!\! - \frac{f(\tau;\theta_{}^1,\theta_{}^2)}{f(\tau;\theta'^1,\theta'^2)}D_{\theta'^1}f(\tau;\theta'^1,\theta'^2) d\tau \biggl|_{\substack{(\theta'^1\!\!, \theta'^2) = (\theta_{}^1\!\!, \theta_{}^2)}}  \label{eqn: KL1derivative}\\
    &=- D_{\theta'^1} \Bigl( \int_{\tau} f(\tau;\theta'^1,\theta'^2) d\tau \Bigr) =  - D_{\theta'^1} (1) = 0 .\nonumber
\end{align}
Similarly, $ D_{\theta'^2} D_{KL}\bigl( (\theta_{}^1,\theta_{}^2),(\theta'^1,\theta'^2)\bigr) = 0$. 
Hence, we write that,
\begin{align}\frac{\partial D_{KL}(\theta_{},\theta')}{\partial \theta'}\Bigl|_{\theta' =\theta_{} } = 0 .\nonumber
\end{align}
Ignoring higher order terms in the approximation, we get
\begin{align}
    D_{KL}\bigl((\theta_{}^1,\theta_{}^2),(\theta'^1,\theta'^2)\bigr) \approx \frac{1}{2} \begin{bmatrix}
\Delta \theta^1 & \Delta \theta^2 \\
\end{bmatrix} \begin{bmatrix}
A_{11} & A_{12}\\
 A_{21} & A_{22}\\
\end{bmatrix} \begin{bmatrix}
\Delta \theta^1 \\ \Delta \theta^2 \\
\end{bmatrix},
\end{align}
where, $A_{ij}$ is defined in \eq~\ref{eqn: KLHession_A}.
Let us now derive all the $A_{ij}$ terms, starting with $A_{11}$,
\begin{align}
    A_{11} &= D_{\theta'^1\theta'^1}\biggl(\int_{\tau} f(\tau;\theta_{}^1,\theta_{}^2)\log\biggl(\frac{f(\tau;\theta_{}^1,\theta_{}^2)}{f(\tau;\theta'^1,\theta'^2)}\biggr) d\tau \biggr)\nonumber\\
    &= \int_{\tau} D_{\theta'^1\theta'^1} \bigl(f(\tau;\theta_{}^1,\theta_{}^2)\log f(\tau;\theta_{}^1,\theta_{}^2)  -f(\tau;\theta_{}^1,\theta_{}^2)\log f(\tau;\theta'^1,\theta'^2) \bigr) d\tau  \label{eqn: KLderivativeexpansion}\\
    &= \int_{\tau} D_{\theta'^1\theta'^1} \bigl(-f(\tau;\theta_{}^1,\theta_{}^2)\log f(\tau;\theta'^1,\theta'^2) \bigr) d\tau \nonumber\\
    &= -\int_{\tau}f(\tau;\theta_{}^1,\theta_{}^2) D_{\theta'^1\theta'^1}\log(f(\tau;\theta'^1,\theta'^2)) d\tau, \nonumber\\
    \shortintertext{and now using the definition of $f(\tau;\theta^1,\theta^2)$ in \eq~\ref{eq: trajectory_distribution}, we get,}
    A_{11} &= -\!\!\int_{\tau} f( \tau; \theta_{}^1,\theta_{}^2 ) D_{\theta'^1\theta'^1} \Bigl( \log \bigl( p(s_0) \prod_{k=0}^{|\tau|-1} \pi( a_{k}^1|s_k;\theta'^1)\nonumber\\ 
    &\qquad\qquad\qquad\qquad\qquad\qquad\pi(a_{k}^2|s_k;\theta'^2 ) R(r_k|s_k,a^1_k,a^2_k ) T(s_{k+1}|s_k,a^1_{k},a^2_{k})\bigr)\Bigr) d\tau \nonumber\\
    &= -\int_{\tau}f(\tau;\theta_{}^1,\theta_{}^2) D_{\theta'^1\theta'^1} \Bigl( \sum_k^{|\tau|-1} \log \pi(a^1_{k}|s_k;\theta'^1)+ \log \pi(a^2_{k}|s_k;\theta'^2)  \Bigr) d\tau \label{eqn: HessianKL}\\
    &= -\int_{\tau}f(\tau;\theta_{}^1,\theta_{}^2) D_{\theta'^1\theta'^1} \Bigl( \sum_k^{|\tau|-1} \log \pi(a^1_{k}|s_k;\theta'^1) \Bigr) d\tau.\nonumber
\end{align}
This formulation allows us to estimate $A_{11}$ using Monte Carlo samples. Note that an interesting alternative to estimate $A_{11}$ only using first order derivatives is in terms of the Fisher matrix. Using \eq~\ref{eqn: KLderivativeexpansion}, we get,  
\begin{align}
    A_{11} &= \int_{\tau} D_{\theta'^1}\Bigl( -f(\tau;\theta_{}^1,\theta_{}^2)D_{\theta'^1}\log(f(\tau;\theta'^1,\theta'^2))\Bigr) d \tau \nonumber \\ 
    &= \int_{\tau}  D_{\theta'^1}\Bigl( -\frac{f(\tau;\theta_{}^1,\theta_{}^2)}{f(\tau;\theta'^1,\theta'^2)}D_{\theta'^1}f(\tau;\theta'^1,\theta'^2)\Bigr) d \tau \nonumber\\ 
    &= \int_{\tau} \!\! -\frac{f(\tau;\theta_{}^1,\theta_{}^2)}{f(\tau;\theta'^1,\theta'^2)}D_{\theta'^1\theta'^1}f(\tau;\theta'^1,\theta'^2) \nonumber \\
    &\qquad + \frac{f(\tau;\theta_{}^1,\theta_{}^2)}{f(\tau;\theta'^1,\theta'^2)^2}D_{\theta'^1}f(\tau;\theta'^1,\theta'^2)D_{\theta'^1}f(\tau;\theta'^1,\theta'^2)^\top d\tau. \nonumber
\end{align}
Similar to \eq~\ref{eqn: KL1derivative}, using the above equation we can write $A_{11}$ at $\theta'^1 = \theta^1, \theta'^2=\theta^2$ as,  
\begin{align}
    A_{11}\Bigl|_{\theta'= \theta_{}}&= \int_{\tau} f(\tau;\theta_{}^1,\theta_{}^2) \frac{D_{\theta'^1}f(\tau;\theta'^1,\theta'^2)}{f(\tau;\theta'^1,\theta'^2)}\frac{D_{\theta'^1}f(\tau;\theta'^1,\theta'^2)}{f(\tau;\theta'^1,\theta'^2)}^\top d\tau \nonumber\\
    &= \mathbb{E}_{f(\tau;\theta_{}^1,\theta_{}^2)}\biggl[D_{\theta^1}\log(f(\tau;\theta'^1,\theta'^2))D_{\theta^1}\log(f(\tau;\theta'^1,\theta'^2))^\top\biggr].\nonumber
\end{align}
Which by definition is the Fisher information matrix. Using the definition of $f(\tau;\theta^1,\theta^2)$ in \eq~\ref{eq: trajectory_distribution} and further reformulations, we get the following estimate for $A_{11}$,
\begin{align}
    A_{11}\Bigl|_{\theta'= \theta_{}}&= \mathbb{E}_{f(\tau;\theta_{}^1,\theta_{}^2)}\biggl[D_{\theta^1}\!\!\sum_t^{|\tau|-1}\Bigl(\log( \pi(a^1_{t+1}|s_k;\theta'^1))\Bigr)D_{\theta^1}\!\!\sum_t^{|\tau|-1}\Bigl(\log( \pi(a^1_{t+1}|s_k;\theta'^1))^\top\Bigr)\biggr]. \nonumber
\end{align}
Note that for our implementation, we use the first approach \eq~\ref{eqn: HessianKL}, but the second approach has the advantage that it avoids second order derivatives, which can potentially improve computation times. 
Using the same ideas as in the derivation of $A_{11}$ in \eq~\ref{eqn: HessianKL}, we can find the $A_{22}$ term, which is
\begin{align} 
A_{22} = -\int_{\tau}f(\tau;\theta_{}^1,\theta_{}^2) D_{\theta'^2\theta'^2} \Bigl( \sum_k^{|\tau|-1} \log \pi(a^2_{k}|s_k;\theta'^2) \Bigr) d\tau \ . \nonumber
\end{align} 

Next, in order to estimate the Hessian of the KL divergence, we want to derive a formulation for $A_{12}$. Similar to \eq~\ref{eqn: HessianKL}, $A_{12}$ can be written as,
\begin{align}
A_{12} &=  -\int_{\tau} f(\tau;\theta_{}^1,\theta_{}^2) D_{\theta'^1\theta'^2} \Bigl( \sum_t^{|\tau|-1}\log \pi(a^1_{t+1}|s_k;\theta'^1) + \log \pi(a^2_{t+1}|s_k;\theta'^2) \Bigr) d \tau \nonumber\\
&= -\int_{\tau} \!\! f(\tau;\theta_{}^1,\theta_{}^2) \Bigl(\!D_{\theta'^1\theta'^2} \!\! \sum_t^{|\tau|-1}\!\Bigl(\log \pi(a^1_{t+1}|s_k;\theta'^1)\Bigr) + D_{\theta'^1\theta'^2}\!\!\sum_t^{|\tau|-1}\!\Bigl(\log \pi(a^2_{t+1}|s_k;\theta'^2)\Bigr) \Bigr) d\tau \nonumber\\
&=0+0\nonumber\\
&= 0.\nonumber
\end{align}
Using the same derivation we also get $A_{21} = 0$.

Thus, the constraint on the KL divergence $D_{KL}((\theta_{}^1,\theta_{}^2),(\theta'^1,\theta'^2)) \leq \delta'$ can be  approximated by the following quadratic inequality constraint, 
\begin{align} \label{eqn: quad_KL_con}
    {\Delta\theta^1}^{\top} A_{11} {\Delta\theta^1} + \Delta {\theta^2}^\top A_{22} \Delta \theta^2 \leq 2 \delta',
\end{align}
which allows us to efficiently find an approximate solution for \eq~\ref{eqn: TRCPOFinalObjective}.

\textbf{Derivation of the \trcpo Update:} Given the bilinear approximation of the objective \eq~\ref{eqn: GradHessianTRCPO} and the quadratic approximation of the KL divergence constraint \eq~\ref{eqn: quad_KL_con}, we can use the the method of Lagrange multiplier to reformulate the problem as the following optimization problem:
\begin{align}\label{eqn: appxtrcpo_game}
    &\max_{{\Delta\theta^1};{\Delta\theta^1}+\theta^1\in \Theta^1}{\Delta\theta^1}^{\top} \!\!D_{\theta^1}L + {\Delta\theta^1}^{\!\!\top} \!\! D_{\theta^1\theta^2}L \Delta \theta^2 - \frac{\lambda}{2}({\Delta\theta^1}^{\top} A_{11} {\Delta\theta^1} + \Delta {\theta^2}^\top A_{22} \Delta \theta^2 - 2 \delta'), \nonumber\\
    &\min_{{\Delta\theta^2};{\Delta\theta^2}+\theta^2\in \Theta^2} {\Delta\theta^2}^{\top} \!\!D_{\theta^2}L + {\Delta\theta^2}^{\!\!\top} \!\! D_{\theta^2\theta^1}L \Delta \theta^1 + \frac{\lambda}{2}({\Delta\theta^1}^{\top} A_{11} {\Delta\theta^1} + \Delta {\theta^2}^\top A_{22} \Delta \theta^2 - 2 \delta').  
\end{align}
Note that in \eq~\ref{eqn: trcpo_game} the same game was used, and thus finding a Nash equilibrium $(\Delta \theta^{1,*},\Delta \theta^{2,*})$ to \eq~\ref{eqn: appxtrcpo_game} allows to define the update rule as $(\theta'^1,\theta'^2) = (\theta_{}^1,\theta_{}^2) + (\Delta\theta^{1,*},\Delta\theta^{2,*})$. The Nash equilibrium $(\Delta\theta^{1,*},\Delta\theta^{2,*})$ of the game, for a fixed $\lambda$, can be obtained by differentiating \eq~\ref{eqn: appxtrcpo_game} with respect to $\theta'^1$, $\theta'^2$ and setting the derivatives to zero:
\begin{align*}
\label{eqn: DiffBilinear_noconstraint}
     &D_{\theta^1}L + D_{\theta^1\theta^2}L \Delta \theta^2  - \lambda(A_{11} \Delta \theta^{1}) = 0, \nonumber\\
     &D_{\theta^2}L + D_{\theta^2\theta^1}L \Delta \theta^1 + \lambda(A_{22} \Delta \theta^2) = 0.\numberthis
\end{align*}
To solve this system of equations, we can formulate \eq~\ref{eqn: DiffBilinear_noconstraint} in matrix form, using $\Delta \theta$ = $[{\Delta \theta^1}^\top,{\Delta \theta^2}^\top]^\top$ and $\delta=2\delta'$, which results in
\begin{gather*}
 \begin{bmatrix}
 D_{\theta^1}L\\
 D_{\theta^2}L
\end{bmatrix}  +
\begin{bmatrix}
0 & D_{\theta^1\theta^2}L\\
 D_{\theta^2\theta^1}L & 0\\
\end{bmatrix}\Delta \theta +
\lambda \begin{bmatrix}
-A_{11} & 0\\
0 & A_{22}\\
\end{bmatrix}\Delta\theta = 0~.
\end{gather*}
This gives us a one dimensional manifold of solutions for $\Delta \theta$ when varying $\lambda$. Using the notation in \eq~\ref{eqn: nash_notation}, we can reformulate the system of equation as,
\begin{align*}
    C \; + \; B \Delta \theta  +  \; \lambda A' \Delta \theta= 0,
\end{align*}
which for non-singular $B + \lambda A'$ has the following solution,
\begin{align*}
    \Delta \theta = -(B \;+\; \lambda A')^{-1}\;C
\end{align*}
For any $\lambda$, in case of a unique solution, we obtain a point $\Delta \theta$. In the end we are interested in a  $\Delta \theta$ solution that satisfies the approximate KL divergence constraint \eq~\ref{eqn: quad_KL_con} or in other words, 
\begin{align}
    {\Delta \theta^1}^\top A_{11} \Delta \theta^{1} + {\Delta \theta^2}^\top A_{22} \Delta \theta^2 \leq 2 \delta'  \nonumber
\end{align}
All $\lambda$ which satisfy this constraint are legitimate solutions, but we prefer the one which assigns higher value to the left hand side of this equation.  Therefore, considering $\frac{1}{\lambda}$ as step size, we select a $\lambda$ that satisfy this constraint with a high value for the left hand side using line search. The solution is given by computing $\Delta \theta = -(B + \lambda A')^{-1}C$ inside a line search, and choosing the solution that satisfies the following constraint with the biggest margin,
 \begin{gather*}
     C^\top(B + \lambda A')^{{-1}^\top}A(B + \lambda A')^{-1}C \leq \delta .
 \end{gather*}
This concludes our derivation.

\section{Implementation of algorithms}
\label{appx: Implementation}
In this section we talk about final setup of \cpg and \trcpo algorithms and their efficient implementation. 
\subsection{Competitive Policy Gradient}
\label{appx: CPGImplementation}
\begin{algorithm}[h!]
\SetAlgoLined
\textbf{Input:} $\pi(.|.;\theta^1)$, $\pi(.|.;\theta^2)$, $V_{\phi}(s)$, N, B, $\alpha$, Choice of $A(s,a^1,a^2;\theta^1,\theta^2)$ mentioned in Sec~\ref{sec:cpg_theory}\\ 
\For{$epoch:e=1$ {\bfseries to} $N$}{
 \For{$batch:b=1$ {\bfseries to} $B$}{
  \For{$k=0$ {\bfseries to} $|\tau|-1$}{
  Sample $a^1_k \sim \pi(a^1_k|s_k;\theta^1)$, $a^2_k \sim \pi(a^2_k|s_k;\theta^2)$.\\
  Execute actions $a_k = (a_k^1,a_k^2)$ \\
  Record $\lbrace s_k, a^1_k, a^2_k, r^1_k, r^2_k \rbrace$ in $\tau_{b}$
  }
  Record $\tau_b$ in \textit{M}
  }
\textbf{1.} Estimate $A(s,a^1,a^2;\theta^1,\theta^2)$ with $\tau$ in M\\
  \textbf{2.} Estimate $D_{\theta^1} \eta $, \, $D_{\theta^2} \eta$, \, $D_{\theta^2\theta^1} \eta$, \, $D_{\theta^1\theta^2} \eta$  using \thm~\ref{thm: CPGGradHessianVal}\\
 \textbf{3.} Simultaneously update $(\theta^1,\theta^2)$ using \eq~\ref{eqn: CGDSolution}
}
\caption{Competitive Policy Gradient}
\label{appxalg: cpg}
\end{algorithm}

Based on our discussion in \sect~\ref{secappx: advantagebilinear} we propose \alg~\ref{appxalg: cpg} which is an extension of \alg~\ref{alg:cpg_short} to utilise advantage function estimation.
In this algorithm, simultaneous update  of $(\theta^1,\theta^2)$ using \eq~\ref{eqn: CGDSolution} requires computing inverse matrix product twice, one for each agent. Instead we can compute infinite recursion strategy using \eq~\ref{eqn: CGDSolution} for one player only and apply the counter strategy for the other player as proposed in~\cite{schfer2019competitive}. From \eq~\ref{eqn: CGDSolution}, for player 1 $\Delta \theta^1 =  \alpha \bigl( Id + \alpha^2 D_{\theta^1\theta^2} \eta D_{\theta^2\theta^1}\eta \bigr)^{-1} \bigl( D_{\theta^1} \eta - \alpha D_{\theta^1\theta^2}\eta D_{\theta^2}\eta \bigr)$. Using this, one can derive an optimal counter strategy $\Delta \theta^2$ for player 2 by solving the $\argmin$ in \eq~\ref{eq:CGDBilinearised_game}. This results in,
\begin{align}
   \Delta \theta^2 =  - \alpha \bigl( D_{\theta^2\theta^1} \eta \Delta \theta^1 + D_{\theta^2}\eta \bigr),\nonumber
\end{align}
which is computed without evaluating an inverse. Similarly, if one evaluates $\Delta \theta^2$ from \eq~\ref{eqn: CGDSolution}, the optimal counter strategy $\Delta \theta^1$ can be obtain by solving the $\argmax$ in \eq~\ref{eq:CGDBilinearised_game}, which is given by,
\begin{align}
   \Delta \theta^1 =  \alpha \bigl( D_{\theta^1\theta^2} \eta \Delta \theta^2 + D_{\theta^1}\eta \bigr).\nonumber
\end{align}
This reduces the requirement of computing inverse to only once per epoch. Further, \cpg has been extended to utilise RMSProp to iteratively adapt the step size as proposed in~\cite{schfer2019implicit}. In RMSProp, the learning rate is adapted based on the moving average of the square of gradients. The code for \cpg along with RMS Prop is also available in the code repository.    

\subsection{Trust Region Competitive Policy Optimization}
\label{appx: TRCPOImplemetation}
Based on our discussion in \sect~\ref{appx: trcpoOptimization}, we propose the trust region competitive policy optimization algorithm given in \alg~\ref{alg: trcpo}.
\label{appx: AlgoTRCPO}
\begin{algorithm}[ht!]
\SetAlgoLined
\textbf{Input:} $\pi(.|.;\theta^1)$, $\pi(.|.;\theta^2)$, N, B, $\delta$, Choice of $A(s,a^1,a^2;\theta^1,\theta^2)$ mentioned in Sec~\ref{sec:cpg_theory}\\  
\For{$epoch:e=1$ {\bfseries to} $N$}{
 \For{$batch:b=1$ {\bfseries to} $B$}{
  \For{$k=0$ {\bfseries to} $|\tau|-1$}{
  Sample $a^1_k \sim \pi(a^1_k|s_k;\theta_1)$, $a^2_k \sim \pi(a^2_k|s_k;\theta_2)$.\\
  Execute actions $a_k = (a_k^1,a_k^2)$ \\
  Record $\lbrace s_k, a^1_k, a^2_k, r^1_k, r^2_k \rbrace$ in $\tau_{b}$
  }
  Record $\tau_b$ in \textit{M}
  }
\textbf{1.} Estimate $A(s,a^1,a^2;\theta^1,\theta^2)$ with $\tau$ in M\\
\textbf{2.} Estimate $D_{\theta^1} L, \ D_{\theta^1} L, \ D^2_{\theta^1\theta^2} L \ D^2_{\theta^2\theta^1} L$ using \eq~\ref{eqn: GradHessianTRCPO}\\ 
\textbf{3.} Solve for $\Delta \theta = [{\Delta \theta^1}^\top , {\Delta \theta^2}^\top]^\top$ in ~\eq~\ref{eqn: TRCO} using line search,\\
 \While{${\Delta \theta}^\top A \Delta \theta<\delta$}{
 $\lambda \xleftarrow{} 2 \lambda$\\
 $\Delta \theta \xleftarrow{} -(B + \lambda A')^{-1}C$
 }
$\theta \xleftarrow{} \theta + \Delta \theta$
}
\caption{Trust Region Competitive Policy Optimisation}
\label{alg: trcpo}
\end{algorithm}

In the experiments, we used \eq~\ref{eqn: HessianKL} to estimate $A'$. As given in the \alg~\ref{alg: trcpo}, \textbf{3.}, we suggest to fulfill the constraint using a line search method. Start with small a value of $\lambda = \lambda_0$, and solve for $\Delta \theta = -(B + \lambda A')^{-1}C$. Check for ${\Delta \theta}^TA\Delta \theta < \delta$. If not satisfied change $\lambda$ to 2$\lambda$ and repeat it until the constraint is satisfied.

There could be multiple heuristics to initialize the $\lambda$ value for a fast convergence. We propose to solve for $\lambda_0$ assuming 2 agents solve for their objective independently (\trpogda). The $\lambda$ value can be obtained in closed form using single-agent \TRPO~\citep{schulman2015trust}. Let us say we obtain $\lambda_{min}$ and $\lambda_{max}$ corresponding to the minimizing and maximizing player. We propose to use $\lambda_0 = \min\lbrace\lambda_{min},\lambda_{max}\rbrace$ as an initialisation.

Similar to single-agent \TRPO, often the resulting step size from the algorithm might be large and thus probably does not benefit either of the agents with policy updates. During such cases, we can further refine the line search iteratively for a smaller step size and update the policies only if both players gain an advantage.

\section{Experiment details}
\label{appxsec: NashGame}
In this section, we provide all the details and explanations of the experiment platforms used in this work. First, we start with the games with a known closed-form Nash equilibrium. Later we talk about the Markov Soccer game and the car racing game.
\subsection{Linear Quadratic (LQ) Game} 
\label{appxsec: LQgame}
\begin{figure}[h]
\hspace{-4.00mm}
	\centering
    \begin{subfigure}[t]{0.25\columnwidth}
  	\centering
  	\includegraphics[width=\linewidth]{images/LQlr_1e2.pdf}
    \caption{}
    \label{Fig: LQlr_1e2}
    \end{subfigure}
    \hspace{-4.00mm}
~
    \begin{subfigure}[t]{0.25\columnwidth}
  	\centering
  	\includegraphics[width=\linewidth]{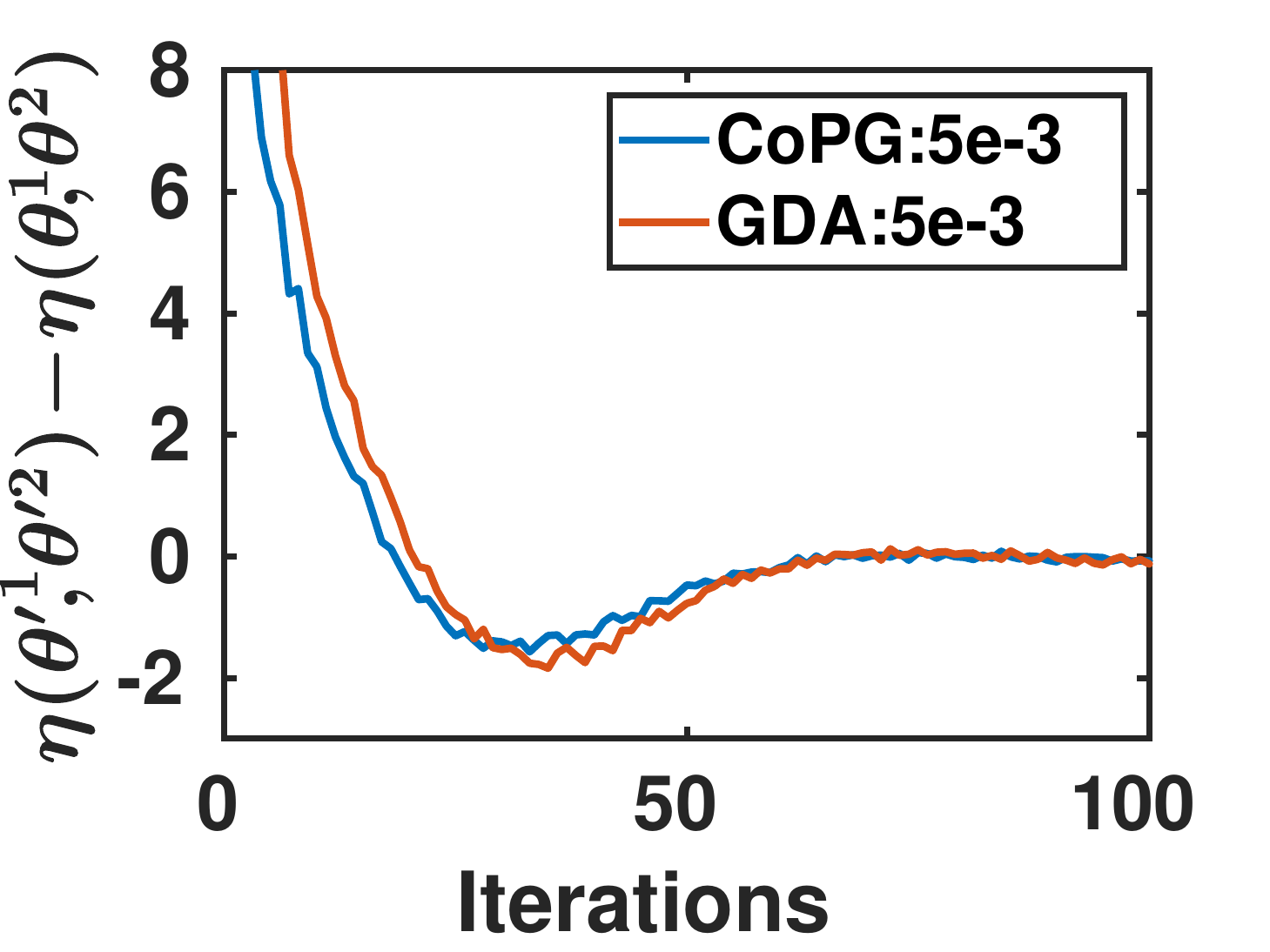}
    \caption{}
    \label{Fig: LQlr_5e3}
    \end{subfigure}
    \hspace{-4.00mm}
   ~
    \begin{subfigure}[t]{0.25\columnwidth}
  	\centering
  	\includegraphics[width=\linewidth]{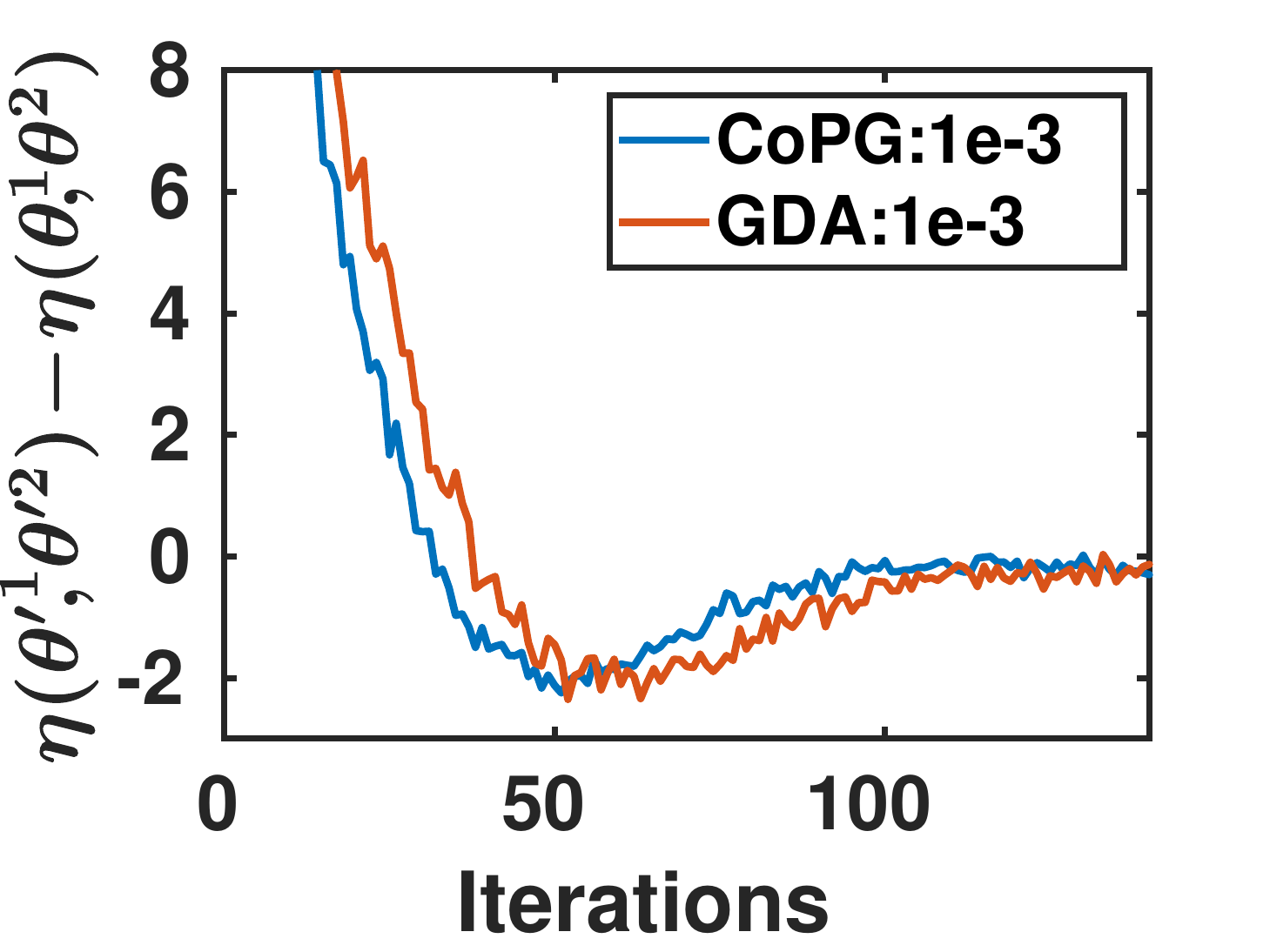}
    \caption{}
    \label{Fig: LQlr_1e3}
    \end{subfigure}
    \hspace{-4.00mm}
~
  \begin{subfigure}[t]{0.25\columnwidth}
  	\centering
  	\includegraphics[width=\linewidth]{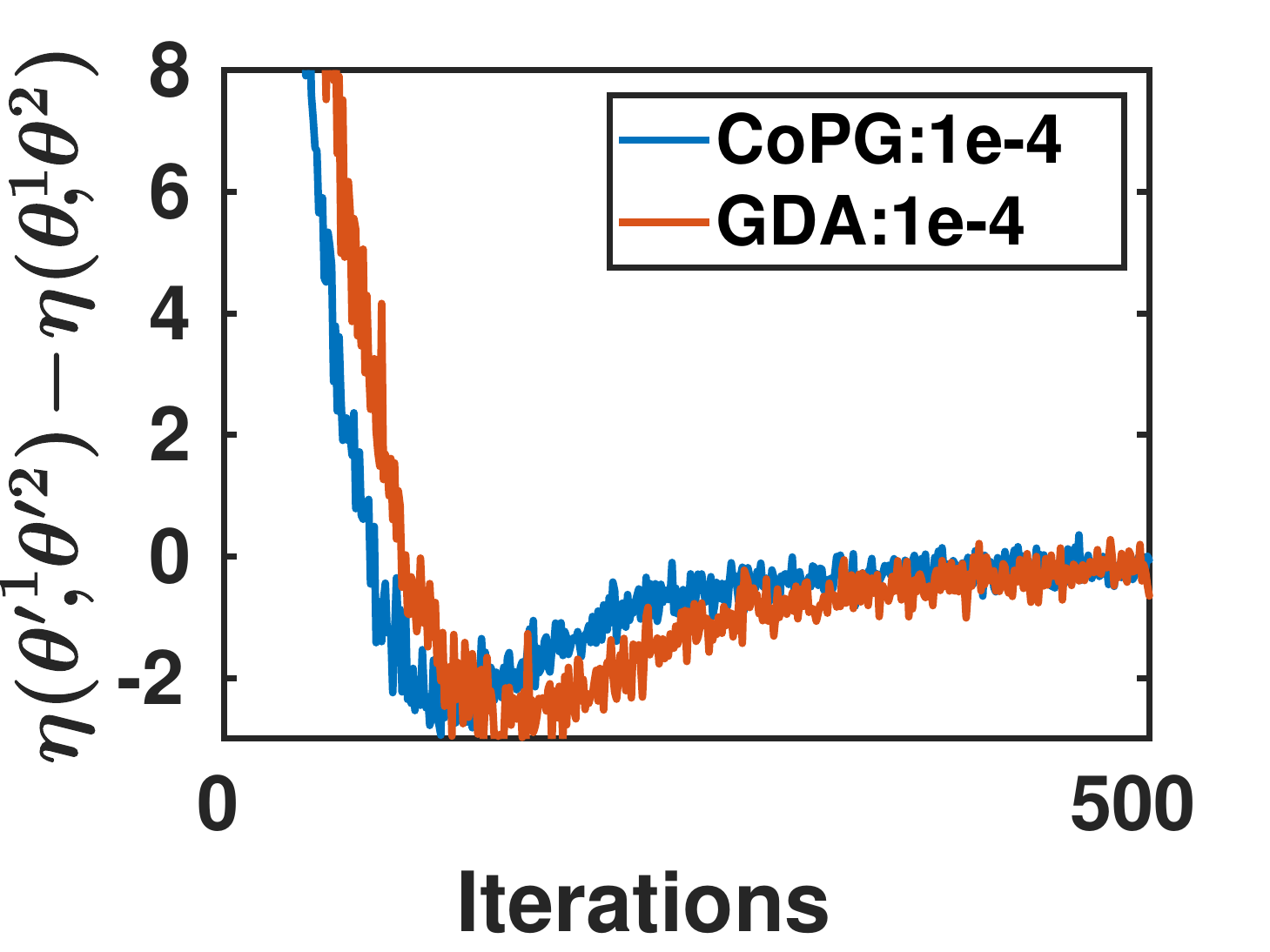}  
    \caption{}
    \label{Fig: LQlr_1e4}
  \end{subfigure}
  \caption{Difference in game objective due to policy updates for LQ game at different learning rates}
  \label{fig: AdvLQgame}
\end{figure}

The zero sum LQ game is defined by: 
\begin{gather} \label{appxeqn: lq_game}
    J(K^1,K^2) = \max_{K^1} \min_{K^2} \mathbb{E} \biggl[ \sum_{k=0}^{|\tau|} \gamma^k \bigl({s_k}^T Q s_k + {a^1_k}^T R_{11} a^1_k   + \ {a^2_k}^T R_{22} a^2_k \bigr) \biggr] , \\
    \text{where:} \ s_{k+1} = As_k + B_{1}a^1_k + B_{2}a^2_k,
     \qquad \qquad a^1_{k+1} = -K^1s_k, \ a^2_{k+1} = -K^2s_k, \nonumber
\end{gather}
where, $s_k \in R^n$ is the system state vector, $a^1 \in R^{d_1}$ and $a^2 \in R^{d_2}$ are the control inputs of player 1 and 2 respectively. The matrix $A \in R^{n\times n}$, $B_1 \in R^{n \times d_1}$ and $ B_1 \in R^{n \times d_2}$ describe the system dynamics. Here, $K^1$ and $K^2$ denotes the control policies. We considered a simple game where, $A = 0.9,\ B_1 = 0.8,\ B_2 = 1.5,\ R_{11} = 1,\ R_{22} = 1$. These environment parameters are not known to the players. The policy of each player is linear in the state and actions are sampled from Gaussian distribution $a^1\sim\mathcal{N}(\mu^1,\sigma^1)$, $a^2\sim\mathcal{N}(\mu^2,\sigma^2)$, with initial values $\mu^1_0 = 0.1$ and $\mu^2_0=-0.1$ and $\log(\sigma^1_0)=\log(\sigma^2_0)=0.1$. The game dynamics and rewards follows \eq~\ref{appxeqn: lq_game}. We collected a batch of 1000 trajectories, each 5 time steps long. The optimal control policy $(K^{1*}, K^{2*} = -0.5735,-0.3059)$, obtained using coupled riccatti equations ~\citep{zhang2019policyZerosum}. The experiment is performed for 5 different random seeds. As per discussion in \appx~\ref{secappx: advantagebilinear}, we used \GAE~for advantage estimation with $\lambda=0.95$ and $\gamma=0.99$. \fig~\ref{fig: AdvLQgame} shows the difference in game objective due to policy update from $(\theta^1,\theta^2)$ to $(\theta'^1,\theta'^2)$ given by $\eta(\theta^1,\theta^2) - \eta(\theta^1,\theta^2) = \mathbb{E}_{\tau \sim  f(\cdot;\theta'^1,\theta'^2)}\sum\nolimits_{k=0}^{|\tau|-1} \gamma^k A(s,a^1,a^2;\theta_{}^1,\theta_{}^2)$ of \gda and \cpg for different learning rates. 
The number of iterations it takes to converge for different learning rates is given in Table~\ref{tab: LQgameLR}.
  \begin{table}[ht]
\centering 
\begin{tabular}{|c||c|c|c|c|c|c|}
\hline
 learning rate & 0.1 & 0.05 & 0.01 & 0.005 & 0.001 & 0.0001\\
 \hline \hline
 \gda & $\oslash$ & $\oslash$ & 113 & 139 & 223 & 611\\
 \hline
 \cpg & 75 & 87 & 105 & 127 & 195 & 590\\
 \hline
\end{tabular}
\caption{Number of iterations to converge to optimal policies}
\label{tab: LQgameLR}
\end{table}

\subsection{Bilinear game}
\label{appxsec: CPGBilinear}
\begin{figure}[ht]
	\centering
    \begin{subfigure}[t]{0.32\columnwidth}
  	\centering
  	\includegraphics[width=\linewidth]{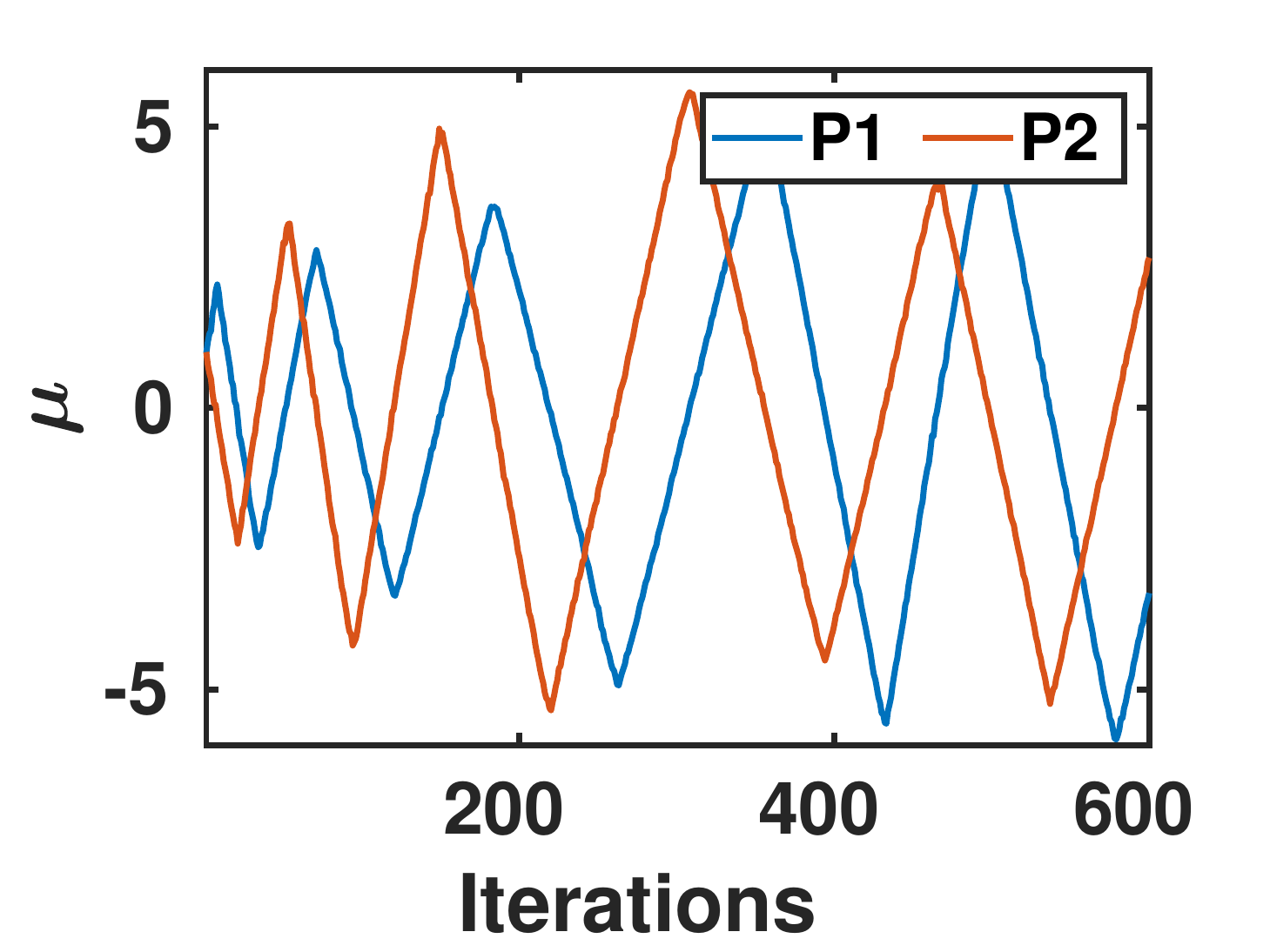}
    \caption{}
    \label{fig: TRCPOBilinear_policyvsstep}
    \end{subfigure}
    ~
    \begin{subfigure}[t]{0.32\columnwidth}
  	\centering
  	\includegraphics[width=\linewidth]{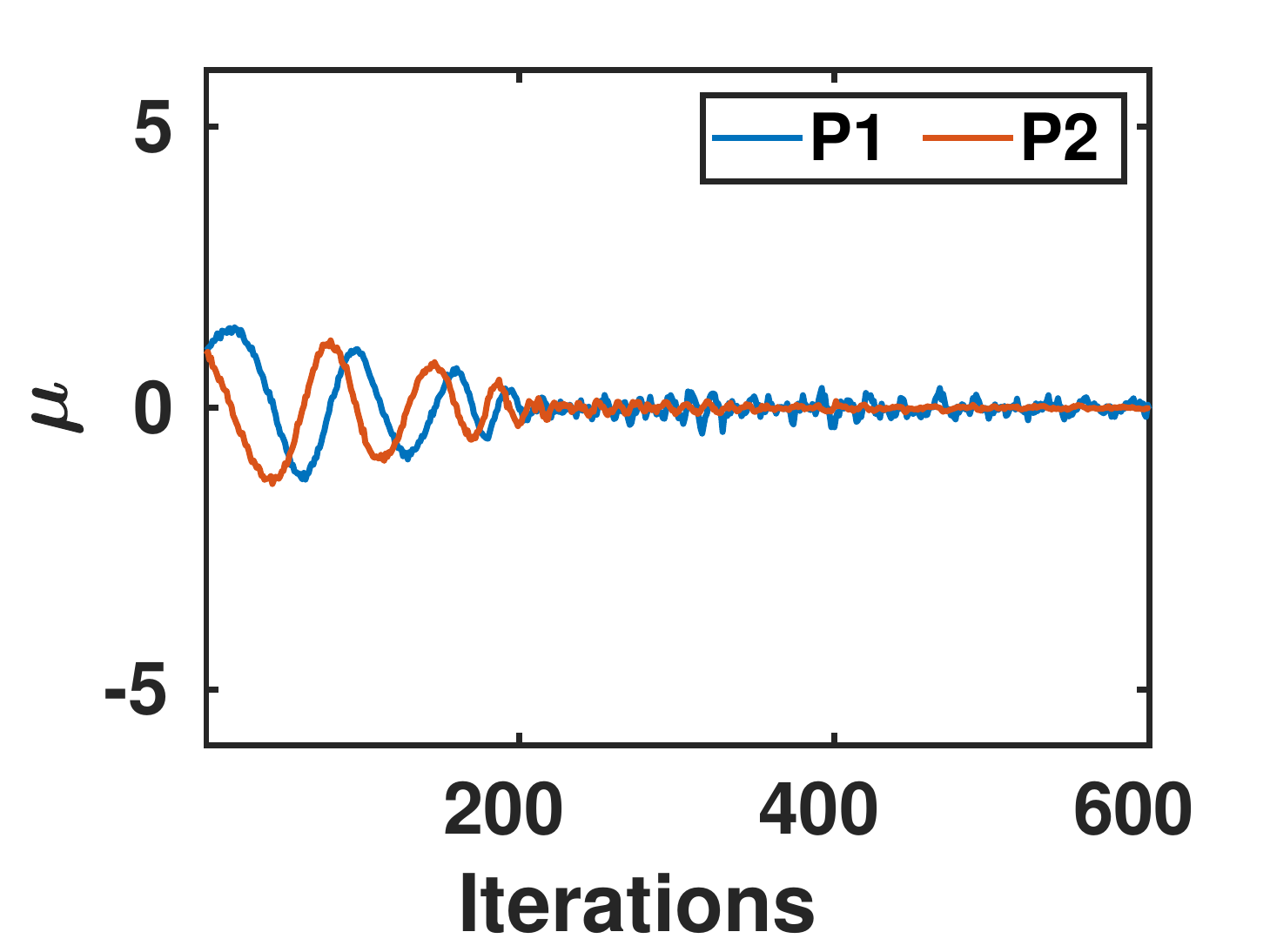}
    \caption{}
    \label{Fig: trcp2gda_bilinear_steps}
    \end{subfigure}
\caption{Policy trajectories of player 1 in a bilinear game. (a) \trpogda (b) \trcpo}
  \label{fig: TRCPOBilinear}
\end{figure}

In a bilinear game both players play actions simultaneously and receive rewards based on the product of their actions, 
\begin{align}
    r^1(a^1,a^2) = a^1a^2 ,\;\;\;\;\;\;\;\;\; r^2(a^1,a^2) = -a^1a^2,
\end{align}
where $a^1\sim\mathcal{N}(\mu_1,\sigma_1)$ and $a^2\sim\mathcal{N}(\mu_2,\sigma_2)$ are actions of player 1 and player 2 respectively. In our experimental setup, a player's policy is modelled as a Gaussian distribution with  $\mu_i, \sigma_i, i\in\{1,2\}$ as mean and variance respectively. The policy is randomly initialised. We collected batch of 1000 trajectories. The experiment was performed for learning rates between 0.5 and 0.05 for 8 random seeds. 
The main paper features results of \cpg vs \gda, but even for the trust region approach we observe a similar contrast between \trcpo and \trpogda. Starting from some random policy initialization, \trcpo converges to the unique Nash equilibrium of the game whereas \trpogda diverges for all $\delta$. \fig~\ref{fig: TRCPOBilinear} shows such behavior for $\delta$ of 0.001.

\subsection{Matching pennies}
\label{appxsec: CPGMP}
The matching pennies game is played between two players, each player holds a coin and hence has two possible actions \{Head, Tail\}. Both the players secretly turn their coin to either head or tail. If the pennies of both the players matches then player 1 wins and player 2 looses. Otherwise, if pennies do not match then player 2 wins and player 1 loses. The game matrix is given in \tab~\ref{appxtab: matching_pennies} captures win and lose of players for every possible action pair. 
\begin{table}[ht]
\centering 
\begin{tabular}{|c||c|c|}
\hline
 & H & T  \\
 \hline \hline
 H & 1,-1 & -1,1   \\
 \hline
 T & -1,1 & 1,-1   \\
\hline
\end{tabular}
\caption{Game matrix for matching pennies game}
\label{appxtab: matching_pennies}
\end{table}

\begin{figure}[ht]
	\centering
    \begin{subfigure}[t]{0.32\columnwidth}
  	\centering
  	\includegraphics[width=\linewidth]{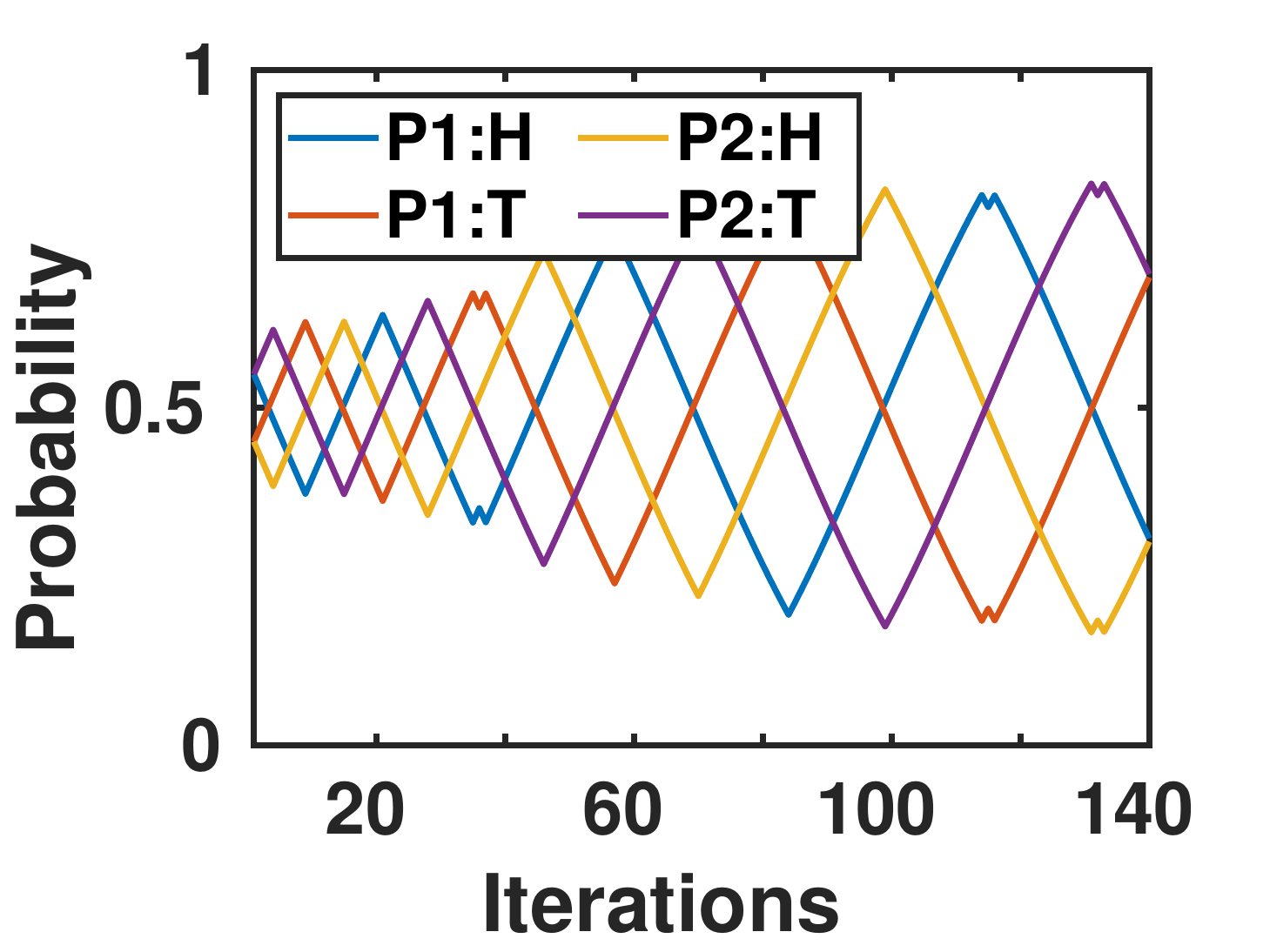}
    \caption{\trpogda}
    \label{Fig: trpo2gda_matching_pennies}
    \end{subfigure}
    ~
        \begin{subfigure}[t]{0.32\columnwidth}
  	\centering
  	\includegraphics[width=\linewidth]{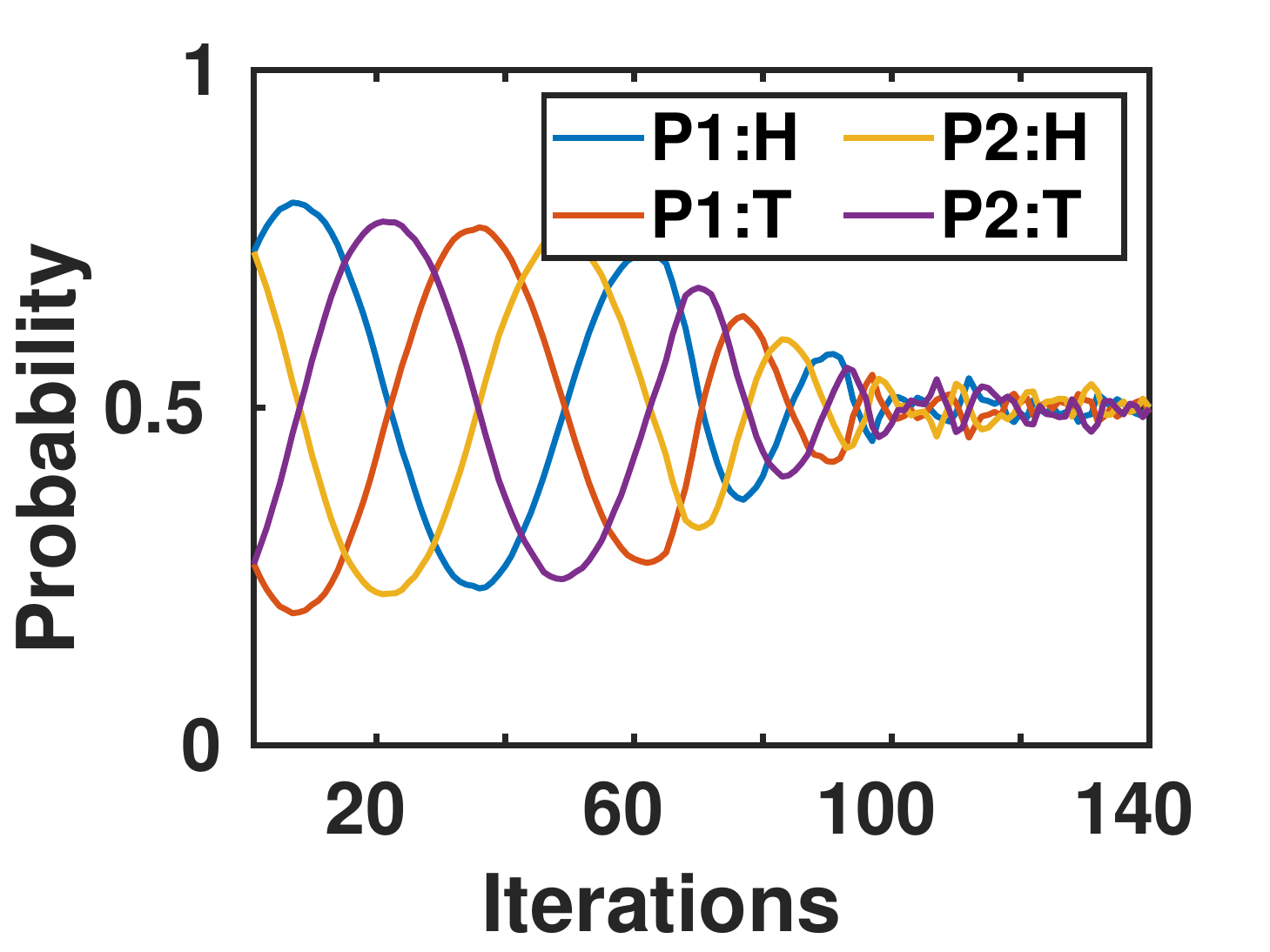}
    \caption{\trcpo}
    \label{Fig: trcpo_matching_pennies}
    \end{subfigure}
\caption{Policy trajectory in matching pennies game}
  \label{fig: TRCPOMatchingPennies}
\end{figure}

Players’ policy is modeled with a two-class categorical distribution, which is randomly initialized. They sample actions from a softmax probability function over the categorical distribution and receive rewards for their actions according to the game matrix. We collect a batch of 1000 trajectories, to estimate gradient and bilinear terms in every epoch. \cpg converges to the Nash equilibrium of the game $(H,T)$ = $\bigl(\frac{1}{2}, \frac{1}{2} \bigr)$. The experiment was performed for learning rates between 0.5 and 0.05 for 8 random seeds.
The main paper features plot of \cpg vs \gda, but even for trust region approach we observe a similar contrast between \trcpo and \trpogda. Starting from random policy initialization, \trcpo converges to the unique Nash equilibrium of the game $(H,T)$ = $\bigl(\frac{1}{2}, \frac{1}{2} \bigr)$, whereas \trpogda diverges for all KL-divergence bounds $\delta$. \fig~\ref{fig: TRCPOMatchingPennies} shows this behaviour for a $\delta$ of 0.01. This experiment was performed also performed for 8 different random seeds and the results where consistent.

\subsection{Rock paper scissors}
\label{appxsec: CPGRPS}
\begin{figure}[h!]
\hspace{-6.00mm}
    \centering
    \begin{subfigure}[t]{0.26\columnwidth}
  	\centering
  	\includegraphics[width=\linewidth]{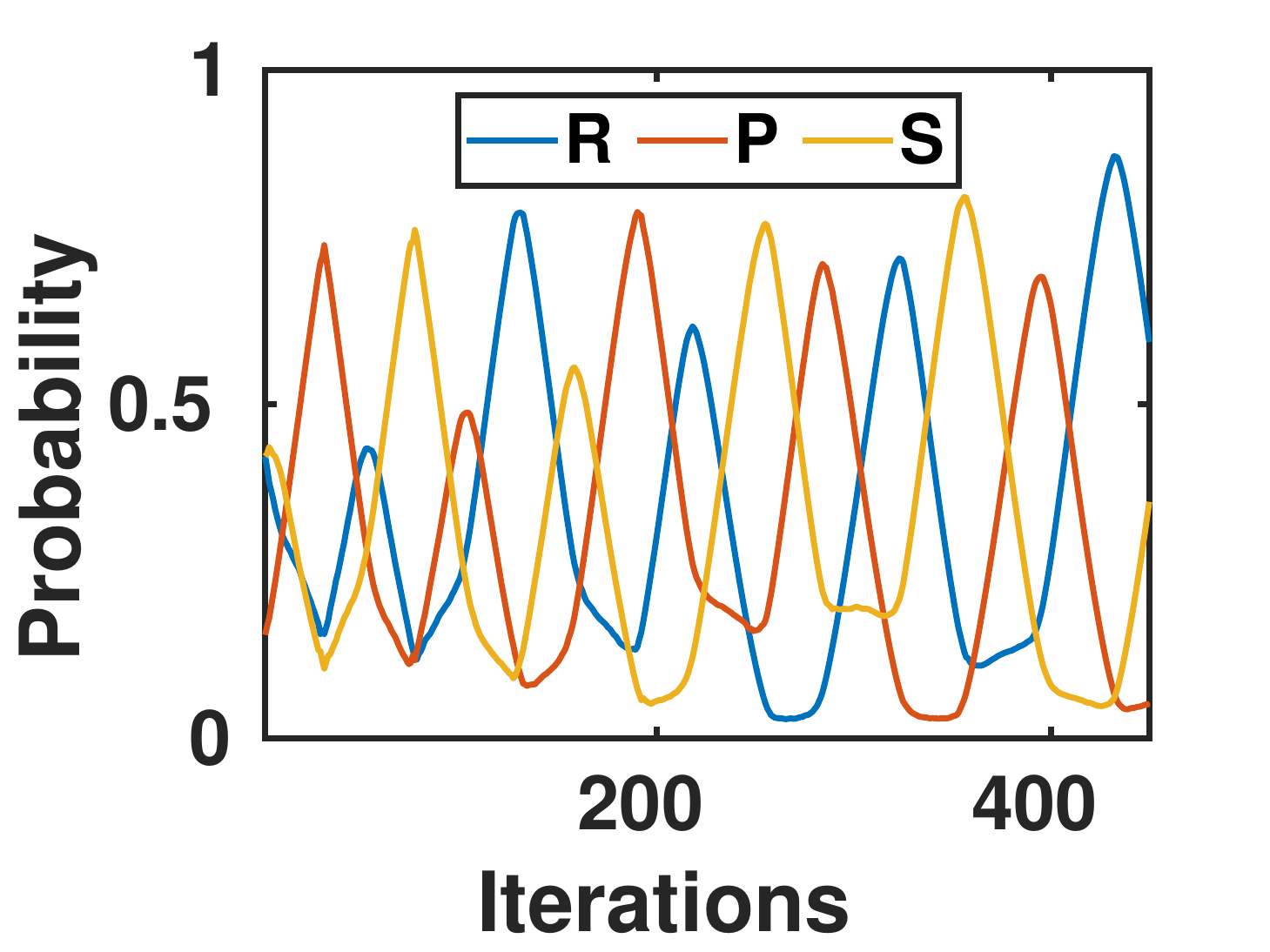}
    \caption{Player 1 \trpogda}
    \label{Fig: trpo2gda_rps_p1}
    \end{subfigure}
    \hspace{-5.00mm}
    ~
    \centering
    \begin{subfigure}[t]{0.26\columnwidth}
  	\centering
  	\includegraphics[width=\linewidth]{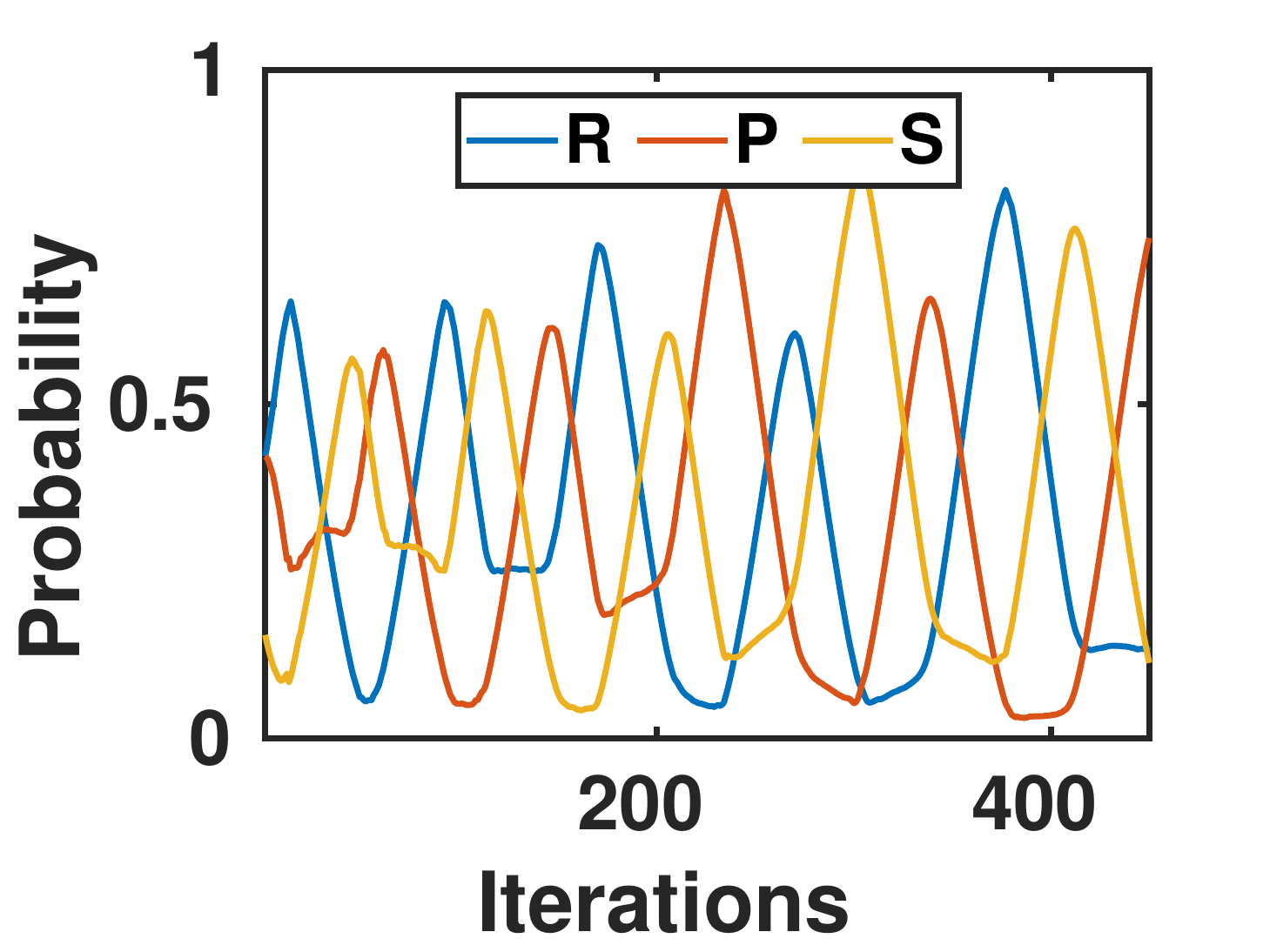}
    \caption{Player 2 \trpogda}
    \label{Fig: trpo2gda_rps_p2}
    \end{subfigure}
    \hspace{-5.00mm}
    ~
	\centering
    \begin{subfigure}[t]{0.26\columnwidth}
  	\centering
  	\includegraphics[width=\linewidth]{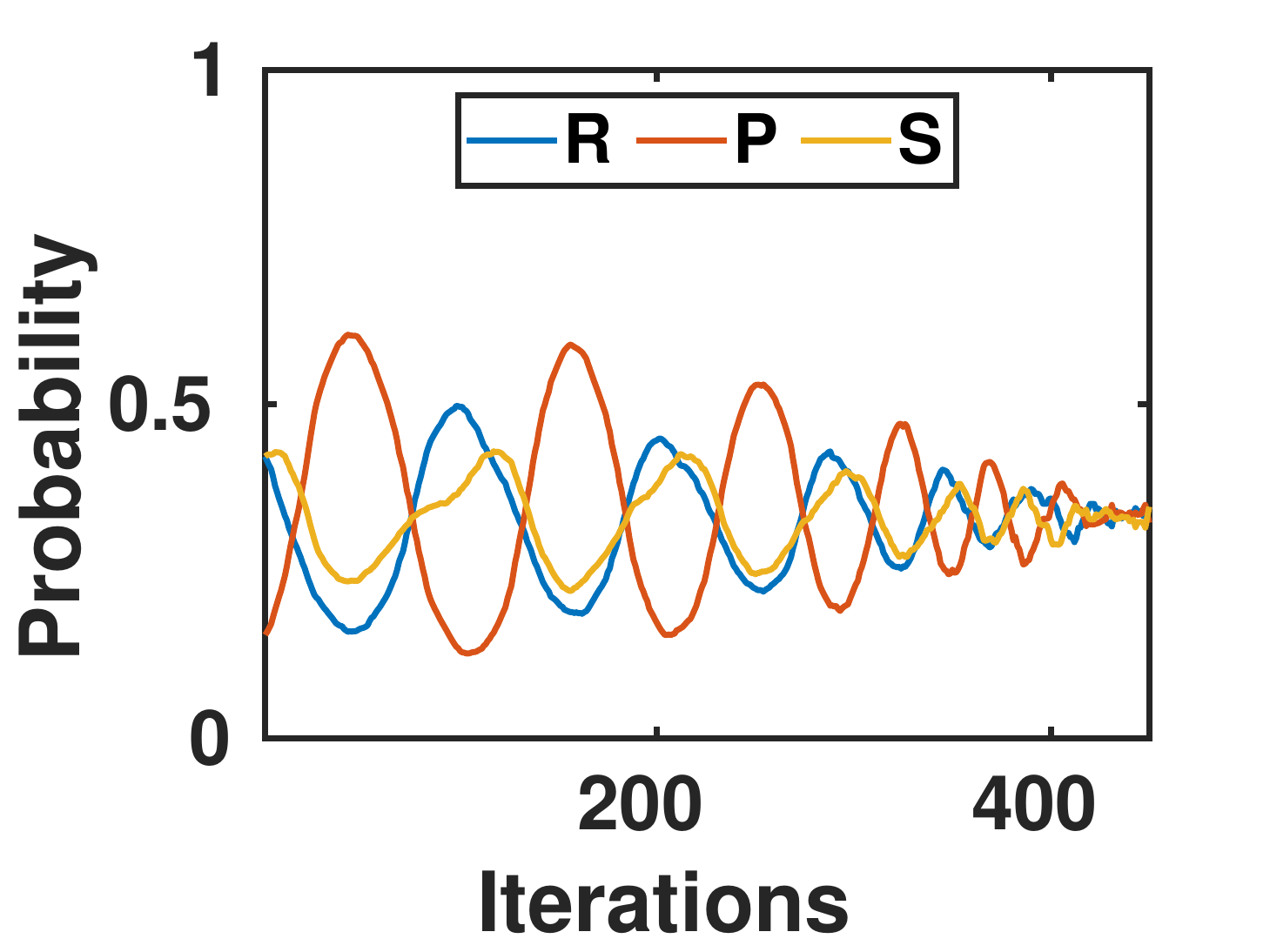}
    \caption{Player 1 \trcpo}
    \label{Fig: trcpo_rps2_p1}
    \end{subfigure}
    \hspace{-5.00mm}
    ~
    \centering
    \begin{subfigure}[t]{0.26\columnwidth}
  	\centering
  	\includegraphics[width=\linewidth]{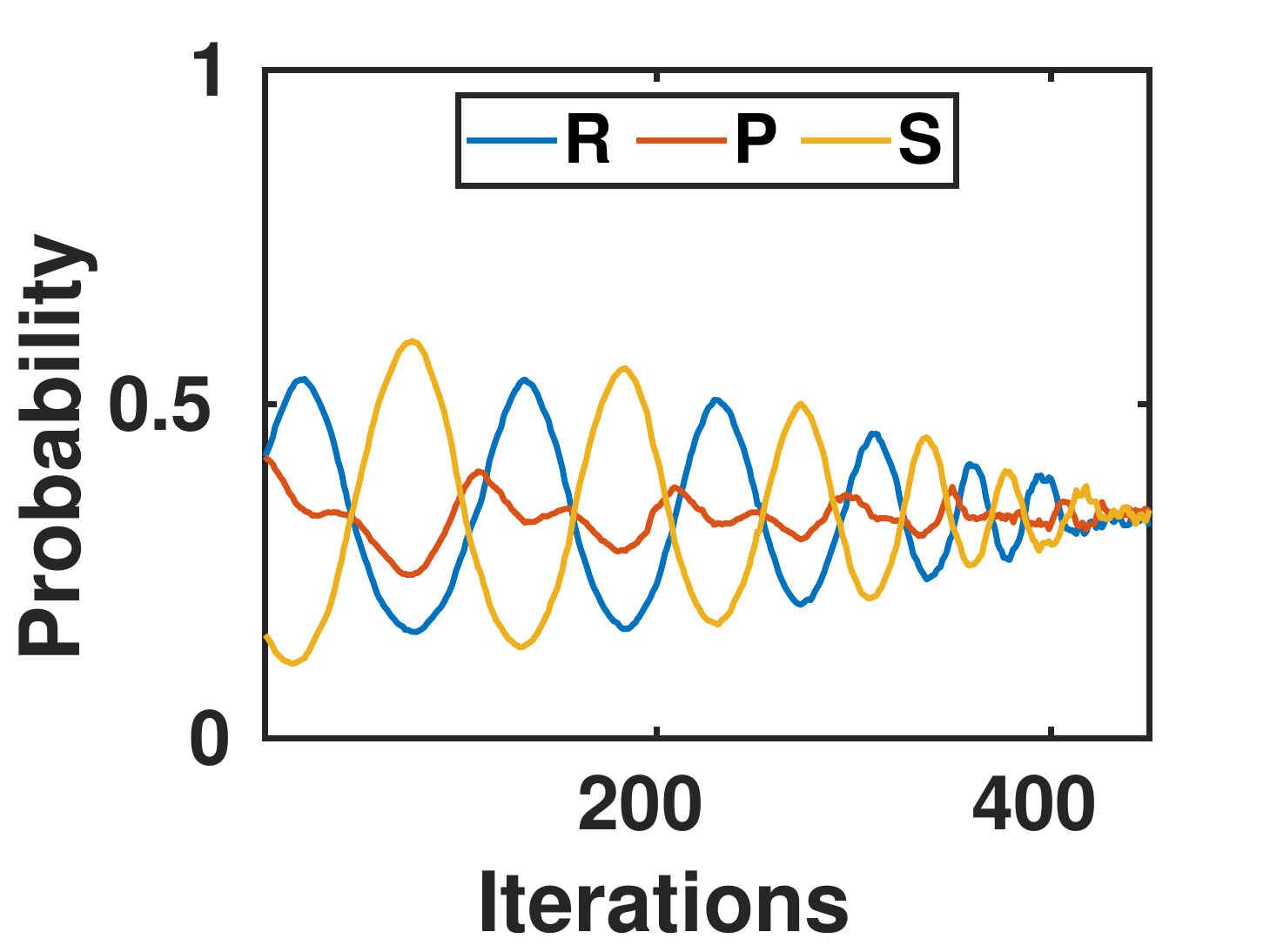}
    \caption{Player 2 \trcpo}
    \label{Fig: trcpo_rps2_p2}
    \end{subfigure}
    \hspace{-6.00mm}
\caption{Policy trajectory in a rock paper scissors game}
  \label{fig: TRCPORPS}
\end{figure}

The rock paper scissors game is played between two players, each player has three possible actions \{Rock, Paper, Scissors\}. Both players will simultaneously display one of three actions: a rock, a paper, or a scissors. Rock beats scissors, scissors beat paper by cutting it, and paper beats rock by covering it. This is captured in the game matrix given by \tab~\ref{appxtab: rock_paper_scissors}.
\begin{table}[ht]
\centering 
\begin{tabular}{|c||c|c|c|}
\hline
 & Rock & Paper & Scissors  \\
 \hline \hline
 Rock & 0,0 & -1,1 & 1,-1   \\
 \hline
 Paper & 1,-1 & 0,0 & -1,1   \\
 \hline
 Scissors & -1,1 & 1,-1 & 0,0   \\
\hline
\end{tabular}
\caption{Game matrix for rock paper scissors game}
\label{appxtab: rock_paper_scissors}
\end{table}
Players’ policies are modeled with 3 class categorical distribution which is randomly initialized. Players sample action from a softmax probability function over categorical distribution. They receive a reward for their actions according to the game matrix. We collect a batch of 1000 trajectories. \cpg converges to the unique Nash equilibrium of the game, where probability of playing $(R, P, S) = (\frac{1}{3},\frac{1}{3},\frac{1}{3})$. The experiment was performed for learning rates between 0.5 and 0.05 for 8 random seeds where \cpg converged in all cases and \gda diverged.
Similarly for the trust region approach, the \trcpo agent converges to a unique Nash equilibrium of the game $(R,P,S)$= $\bigl(\frac{1}{3}, \frac{1}{3}, \frac{1}{3} \bigr)$, where as \trpogda diverges. \fig~\ref{fig: TRCPORPS} shows this behavior for $\delta$ of 0.001. The experiment was performed for 8 different random seeds.

\subsection{Markov Soccer Game}
\label{appxsec: soccer}
The soccer game setup is shown in \fig~\ref{fig: Soccer}. It is played between two players A and B, both are randomly initialized in one of the 4x5 grid cells. The ball is also randomly initialised in one of the 4x5 grid cells, this contrast to the previous studies, where one of the players was randomly initialised with the ball. Players are supposed to pick up the ball and place it in the opponent's goal. The players are allowed to move \textit{up, down, left, right} or stand still. The player without the ball seizes the ball, if both players move simultaneously to the same grid cell. The game ends if a player scores a goal or none of them scores within 1000 steps. The winner receives a +1 reward and -1 is awarded to the losing player, hence the game is formulated as a zero sum game. This game does not have an optimal deterministic policy. This game depicts heavy interactions, where strategy depends on what the other player plays. Potentially, a good player can learn defending, dodging and scoring. 

The state vector of a player P with respect to opponent O is $s^P = [x_{G_O},y_{G_O},x_{ball},y_{ball},x_{O},y_{O}]$, where $P, O \in \{A,B\}$. $G_O$ is goal of the opponent, $x,y$ refers to the relative position from the player to the subscript, e.g., $x_{O},y_{O}$ is the relative position to the opponent $O$. The state vector of the game used during training is $s = [s^{A},s^{B}]$, $s \in \mathcal{R}^{12}$, it captures the position of each player relative to the goal, the ball and relative to the opponent. The players' policy maps state vector of the game to a categorical distribution with 5 categories using a network with two hidden layers one with 64 and other with 32 neurons. Players sample actions from a softmax probability function over the categorical distribution. The players were trained using \gda and \cpg. In both the algorithms, players were trained for roughly 30,000 episodes until the number of wins against any good player saturates. In each epoch we collected a batch which consists of 10 trajectories. All the parameters were the same throughout the training for \cpg and \gda. The experiment was tested with 6 different random seeds. We used learning rate of 0.01 and \GAE~for advantage estimation with $\lambda$ of 0.95 and $\gamma$ of 0.99. 

\begin{figure}[!h]
\hspace{-3.00mm}
	\centering
    \begin{subfigure}[t]{0.26\columnwidth}
  	\centering
  	\includegraphics[width=\linewidth]{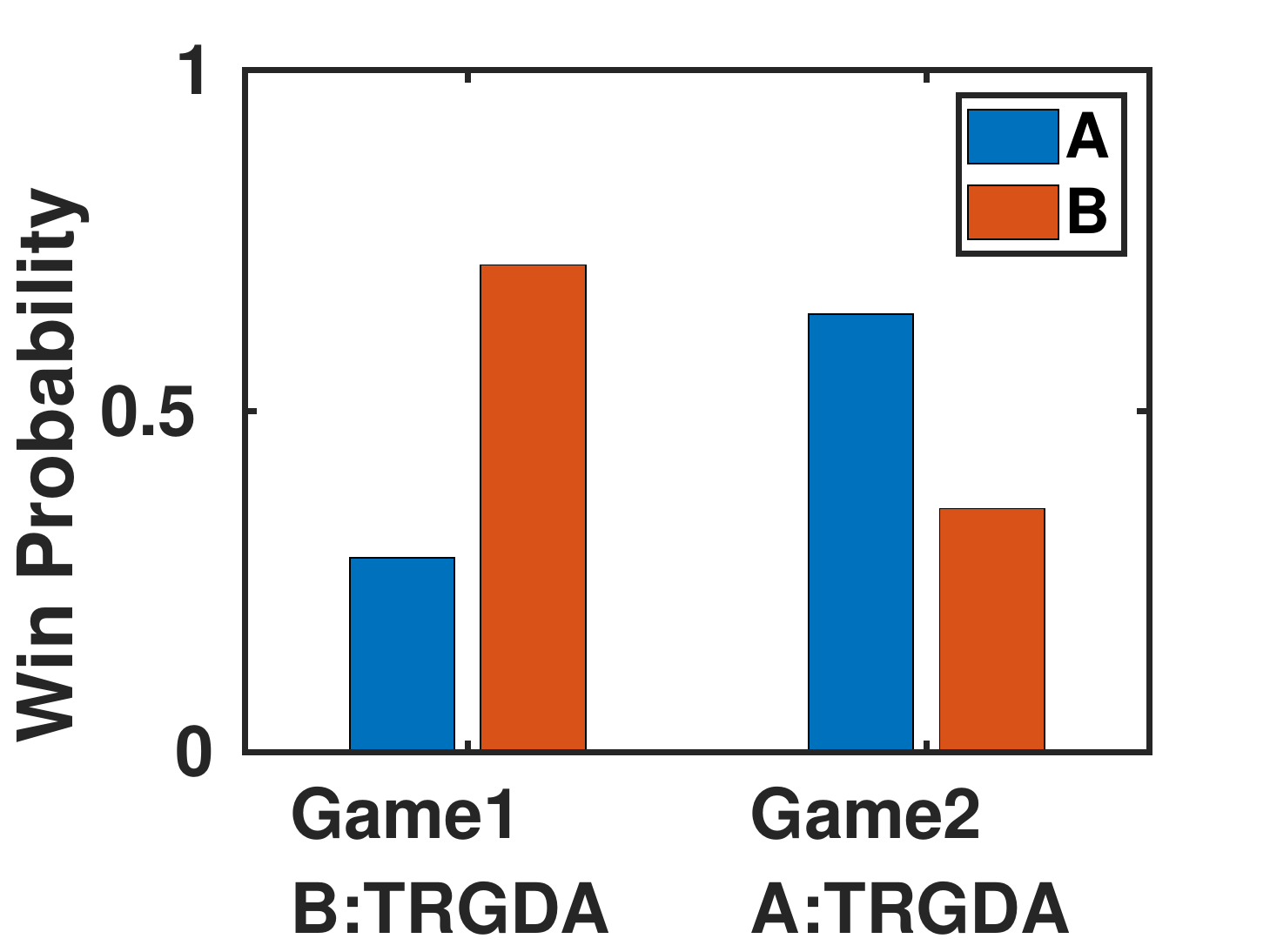}
    \caption{\gda vs \trpogda}
    \label{Fig: EX_sgd_trpo_winning_plot}
    \end{subfigure}
    \hspace{-6.00mm}
    ~
    \begin{subfigure}[t]{0.26\columnwidth}
  	\centering
  	\includegraphics[width=\linewidth]{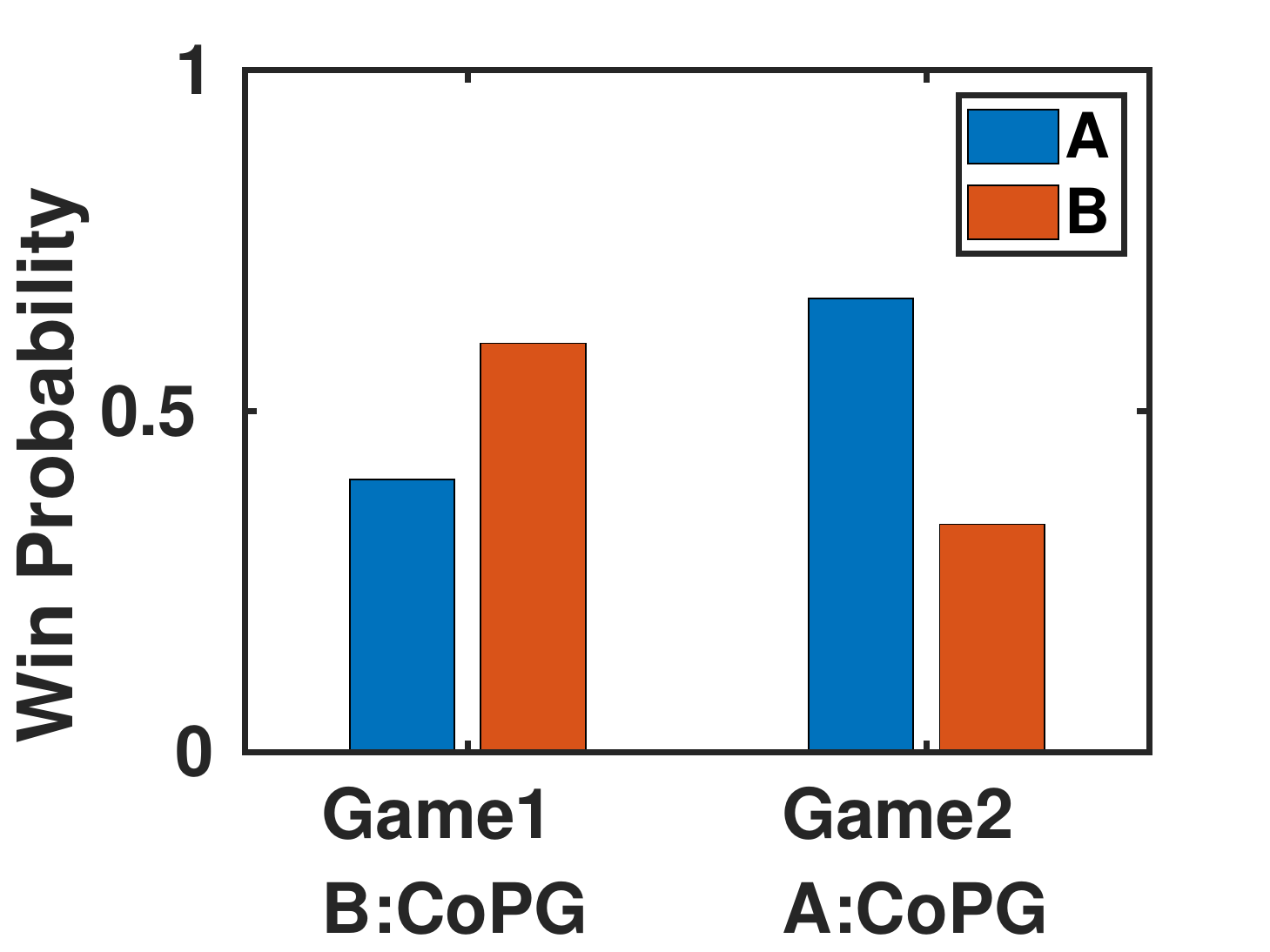}
    \caption{\trpogda vs \cpg }
    \label{Fig: EX_trpo_cpg_winning_plot}
    \end{subfigure}
    \hspace{-6.00mm}
    ~
    \begin{subfigure}[t]{0.26\columnwidth}
  	\centering
  	\includegraphics[width=\linewidth]{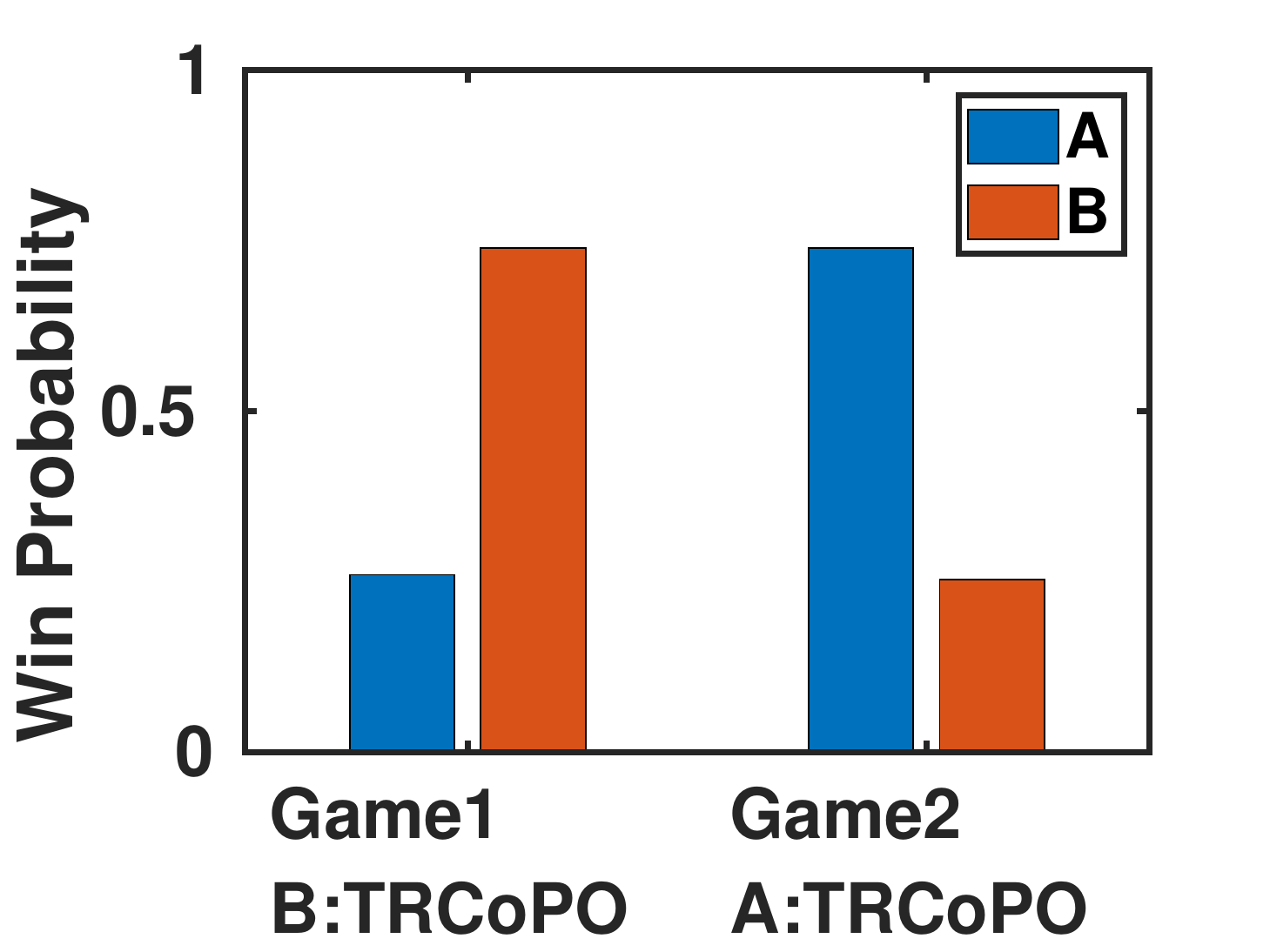}
    \caption{\cpg vs \trcpo}
    \label{Fig: EX_cpg_trcpo_winning_plot}
    \end{subfigure}
    \hspace{-6.00mm}
~
  \begin{subfigure}[t]{0.26\columnwidth}
  	\centering
  	\includegraphics[width=\linewidth]{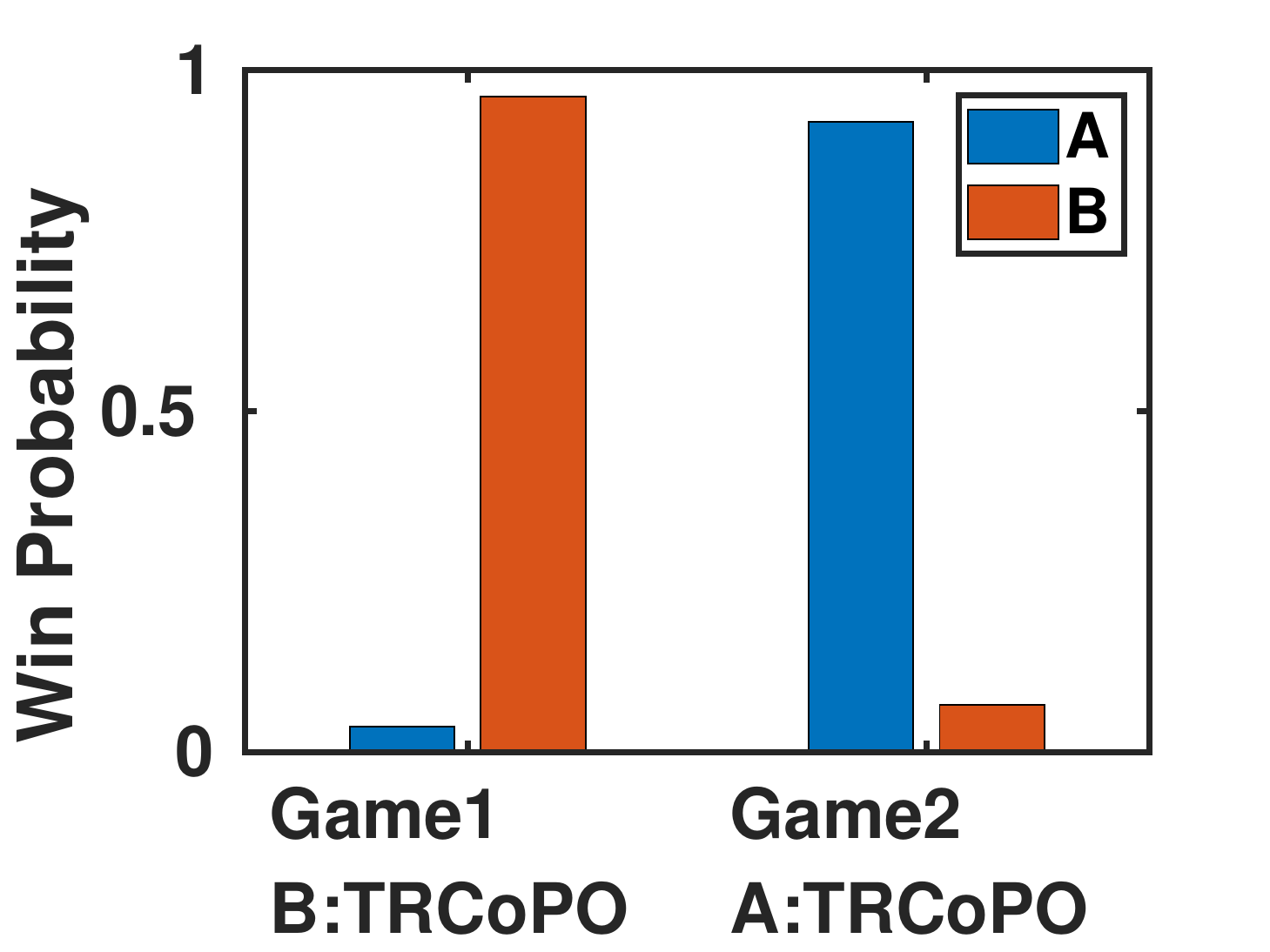}
    \caption{\gda vs \trcpo}
    \label{Fig: EX_sgd_trcpo_winning_plot}
  \end{subfigure}
  \hspace{-4.00mm}
  \caption{ Win probability in soccer game played between \gda, \cpg, \trpogda and \trcpo }
  \label{fig: CompMarkovScoccer}
  \vspace{-1.0em}
\end{figure}

Similarly, for the trust region approach the same game setting was used, with a maximum KL divergence $\delta$ of 0.0001. The players were trained with \trcpo or \trpogda. Both the agents were trained for 5,000 episodes, each episode denotes a batch update where each batch consists of 10 trajectories.  For comparison, we played 10000 games between agents trained using \trcpo and with that using \trpogda, in which for 50\% of the games \trcpo was player A and for remaining \trpogda was player A. \trpogda generates unequal players, on competing with the stronger \trpogda agent, the \trcpo agent still won more than 80\% games against \trpogda as shown in \fig~\ref{Fig: trpo_trcpo_winning_plot}. \trcpo was able to learn better tactics for snatching the ball and dodging and defending its goal post. 
The histogram in \fig~\ref{fig: CompMarkovScoccer}, depicts a comparison of \gda, \cpg, \trpogda, and \trcpo in the soccer game playing against each other.

\subsection{The Car Racing game}
\label{appx: ORCADescription}
\begin{wrapfigure}{r}{0.53\textwidth}
\vspace*{-1em}
\hspace{-6em}
    \begin{subfigure}[h]{0.9\columnwidth}
  	\includegraphics[width=\linewidth]{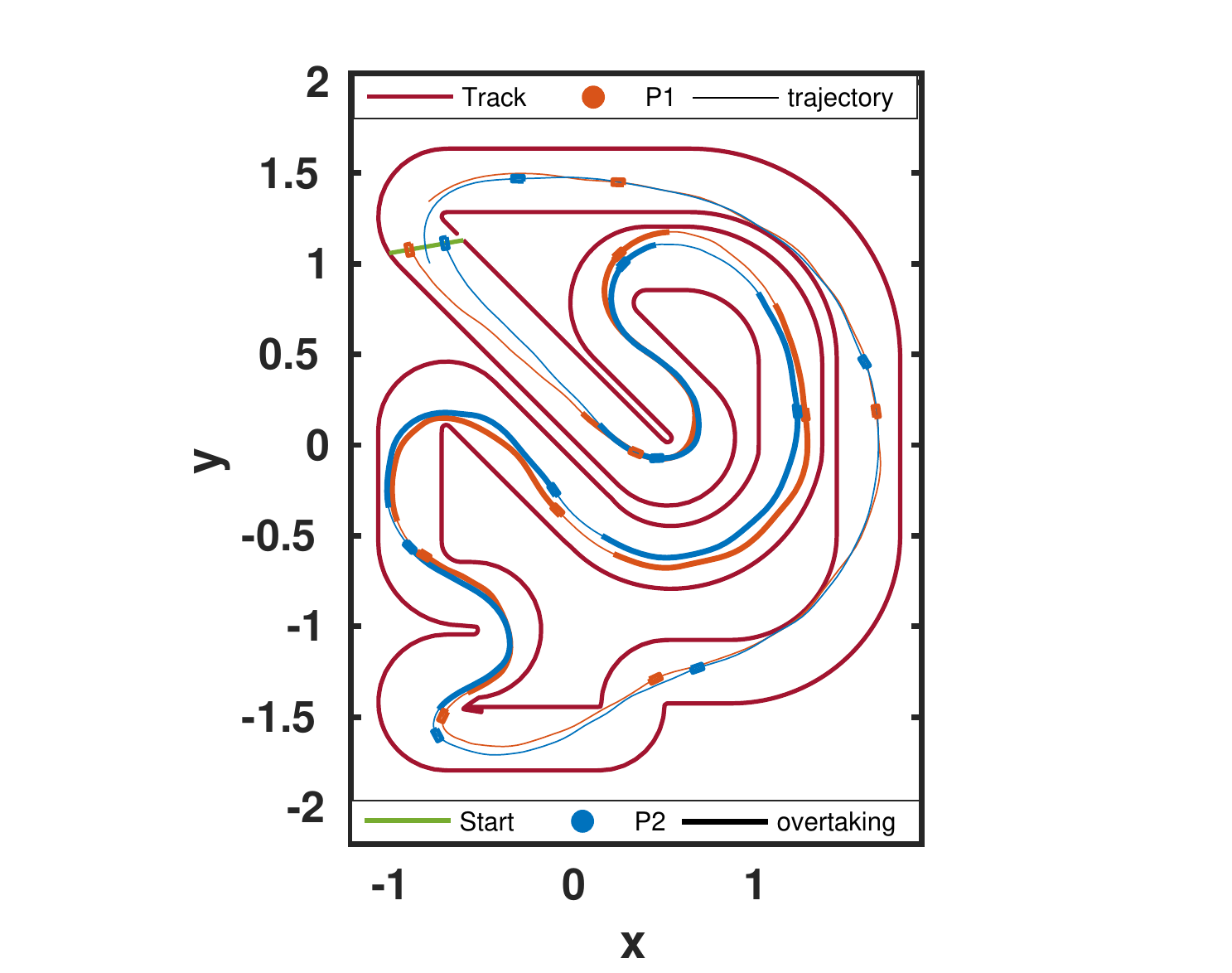}
    \end{subfigure}
  \caption{Overtaking maneuvers of \cpg agents in the Car Racing Game. The thick line shows trajectory when the trailing agent overtook.}
  \label{fig: overtaking_manuever_northsouth}
  \vspace{-1em}
\end{wrapfigure}

Our final experimental setup is a racing game between two miniature race cars, where the goal is to finish the one lap race first. This involves both, learning a policy that can control a car at the limit of handling, as well as strategic interactions with the opposing car. Only if a player has the correct relative position to the opponent, this player is able to overtake or to block an overtaking. Our simulation study mimics the experimental platform located at ETH Zurich, which use miniature autonomous race cars.
Following ~\citep{Liniger2017OptimizationBasedAR} the dynamics of one car is modeled as a dynamic bicycle model with Pacejka tire models ~\citep{mf}. However, compared to ~\citep{Liniger2017OptimizationBasedAR} we formulate the dynamics in curvilinear coordinates~\cite{Vzquez2020OptimizationBasedHM} where the position and orientation are represented relative to a reference path. This change in coordinates significantly simplifies the definition of our reward, and simplified the policy learning. The resulting state of a single car is given as $z =  [\rho, d, \mu, V_x, V_y, \psi]^T$, where $\rho$ is the progress along a reference path, $d$ is the deviation from a reference path, $\mu$ is the local heading with respect to the reference path, $V_x$ and $V_y$ are the longitudinal and the lateral velocity respectively in car frame and $\psi$ is the yawrate of the car. The inputs to the car is $[D, \delta]^T$, where $D \in [-1, 1]$ is duty cycle input to the electric motor varying from full braking at $-1$ to full acceleration at $1$ and $\delta \in [-1, 1]$ is the steering angle. The test track which consists of 13 turns with different curvature can be seen in Figure \ref{fig: overtaking_manuever_northsouth}. For the reference path we used the X-Y path obtained by a laptime optimization tool~\cite{Vzquez2020OptimizationBasedHM}, note that it is not necessary to use a pre-optimize reference path, but we saw that it helped the convergence of the algorithm. Finally, to get a discrete time MDP we discretize the continuous time dynamics using an RK4 integrator with a sampling time of 0.03s. 
To formulate the racing game between two cars, we first define the state of the game as the concatenated state of the two players, $s = [z^1,z^2]$. Second we convert the objective of finishing the race first without an accident into a zero sum game. Therefore, we define the following reward function using reward shaping. First to model our no accident constraints we use a freezing mechanism: (i) If the car is leaving the track we freeze it until the episode ends, (ii) if the cars collide, the rear car (car with the lower progress $\rho$) is stopped for 0.1s, which corresponds to a penalty of about two car lengths. Note that this gives an advantage to the car ahead, but in car racing the following car has to avoid collisions. Furthermore, an episode ends if both cars are either frozen due to leaving the track or the first car finished the lap. Finally, to reward the players receive at every time step is $r(s_k,a^1_{k+1},a^2_{k+1}) = \Delta \rho_{car_1} - \Delta \rho_{car_2}$, where $\Delta \rho = \rho_{k+1} - \rho_k$. This reward encourages the player to get as far ahead of the of the opponent player as possible. Note that the reward is zero sum, and the collision constraints do only indirectly influence the reward.

For the training we started the players on the start finish line ($\rho = 0$) and randomly  assigned $d \in \{0.1, -0.1\}$. We also limited one episode to 700 time steps, which is about twice as long as it takes an expert player to finish a lap. For each policy gradient step we generated a batch of 8 game trajectories, and we run the training for roughly 20000 epochs until the player consistently finish the lap. This takes roughly 62 hours of training for a single core CPU. To increase the performance and robustness of learning we adapted the learning rate with RMS prop for both \gda and \cpg and used a slow learning rate of $5.10^{-5}$ for both players. For our experiments we run 8 different random seeds and report the numbers of the best seed. As a policy we use a multi-layer perceptron with two hidden layers, each with 128 neurons, we used ReLU activation functions and a Tanh output layer to enforce the input constraints. \GAE~ is used for advantage estimation with $\lambda$ of $0.95$ and $\gamma$ of $0.99$. The same setting was also used to compare performance of \trcpo and \trpogda. The maximum KL divergence $\delta$ was set between 1e-4 to 1e-5 and the training was performed for 8 different random seeds. The statistics of the experiments are presented in the main part of the paper~\sect~\ref{sec:exp_cpg_trcpo}.

\subsection{Comparison with \maddpg and \lola}
\label{appx: maddpg_comp}

\begin{figure}[h]
{\centering
  	\hspace{-3.00mm}
    \begin{subfigure}[t]{0.26\columnwidth}
  	\centering
  	\includegraphics[width=\linewidth]{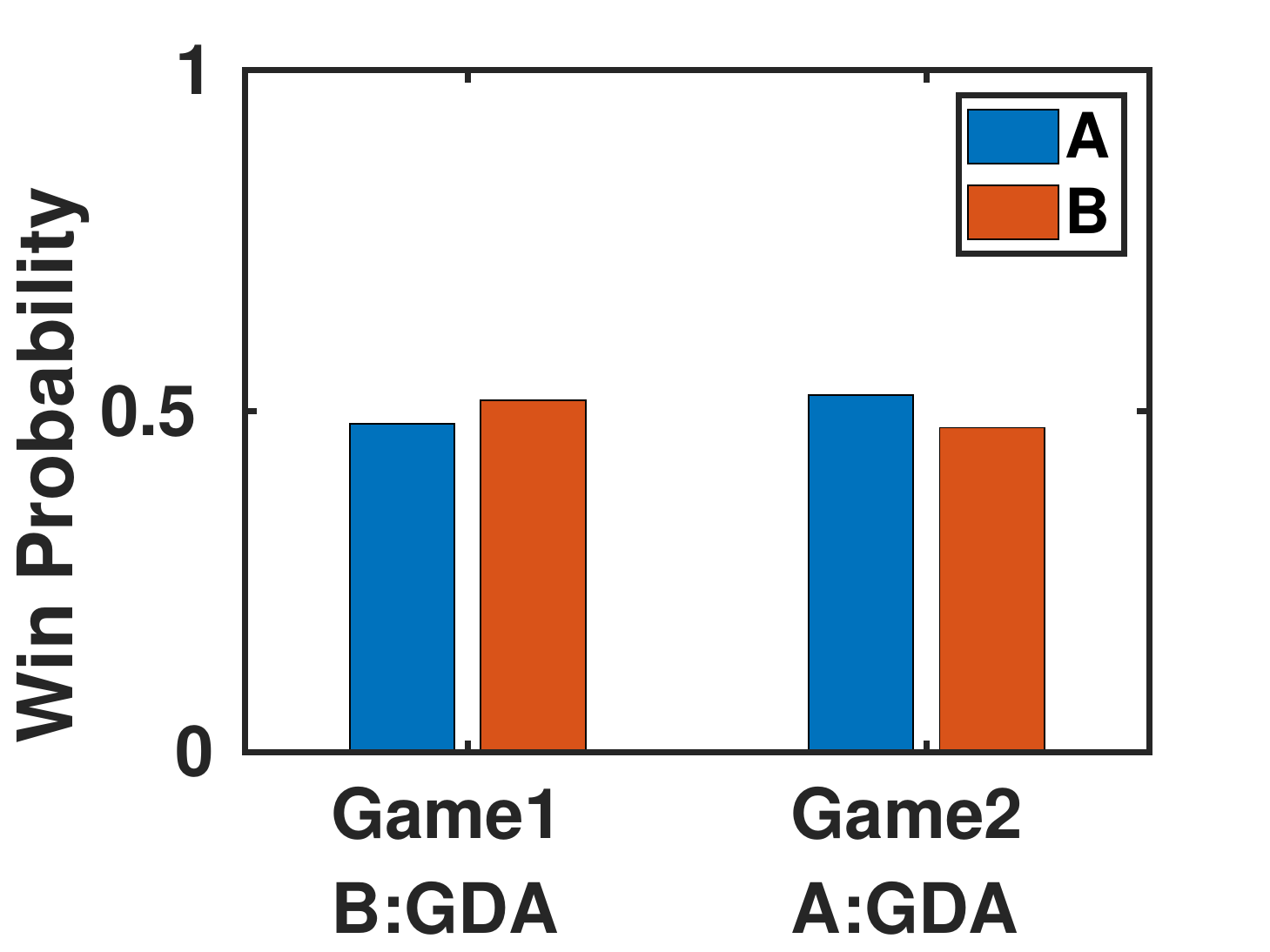}
    \caption{\gda vs \maddpg}
    \label{fig: EX_sgd_maddpg_winning_plot}
    \end{subfigure}
    \hspace{-6.0mm}
~
  \begin{subfigure}[t]{0.26\columnwidth}
  	\centering
  	\includegraphics[width=\linewidth]{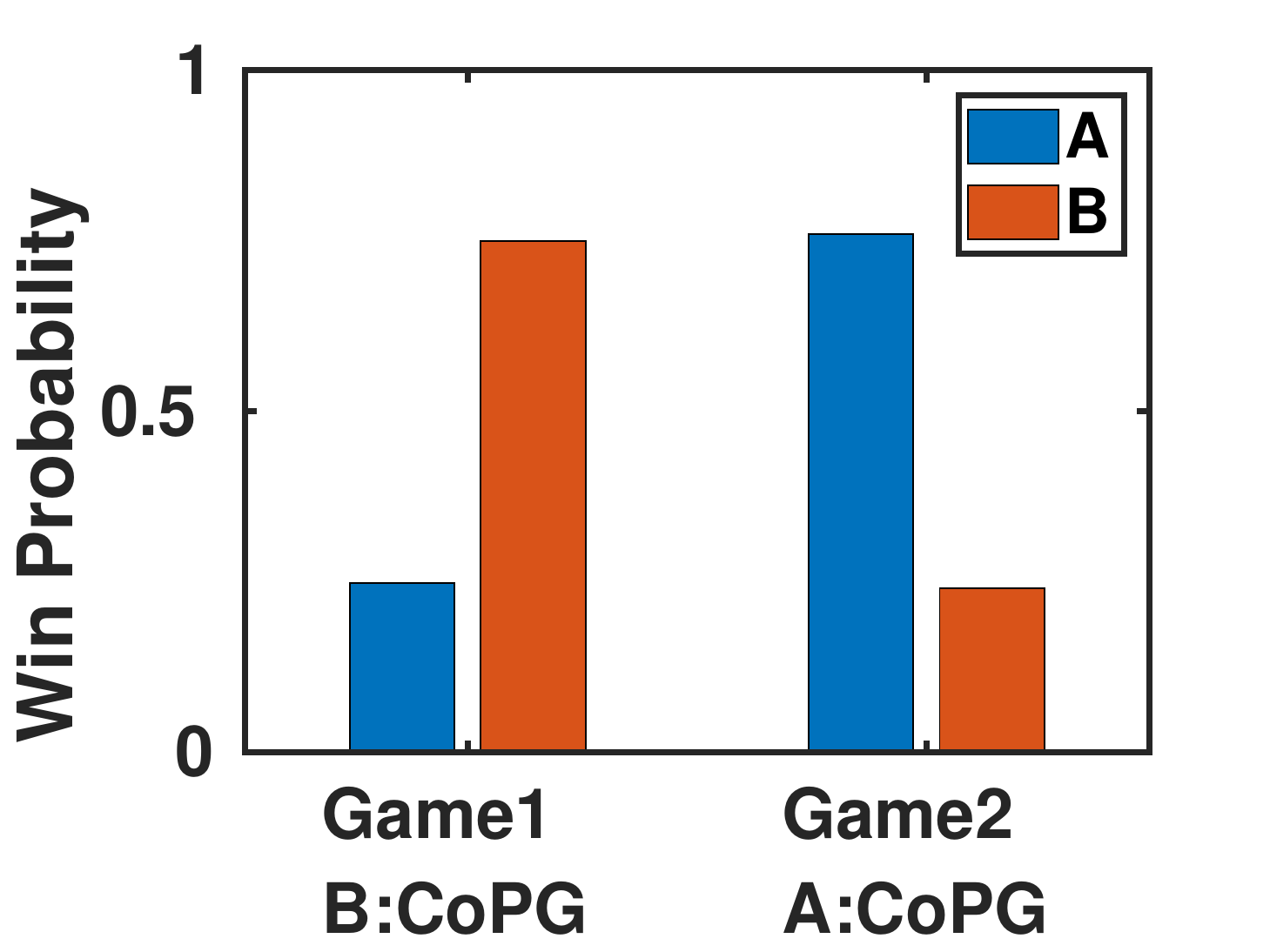}
    \caption{\cpg vs \maddpg}
    \label{fig: EX_cpg_maddpg_winning_plot}
  \end{subfigure}
   \hspace{-6.00mm}
   ~
  \begin{subfigure}[t]{0.26\columnwidth}
  	\centering
  	\includegraphics[width=\linewidth]{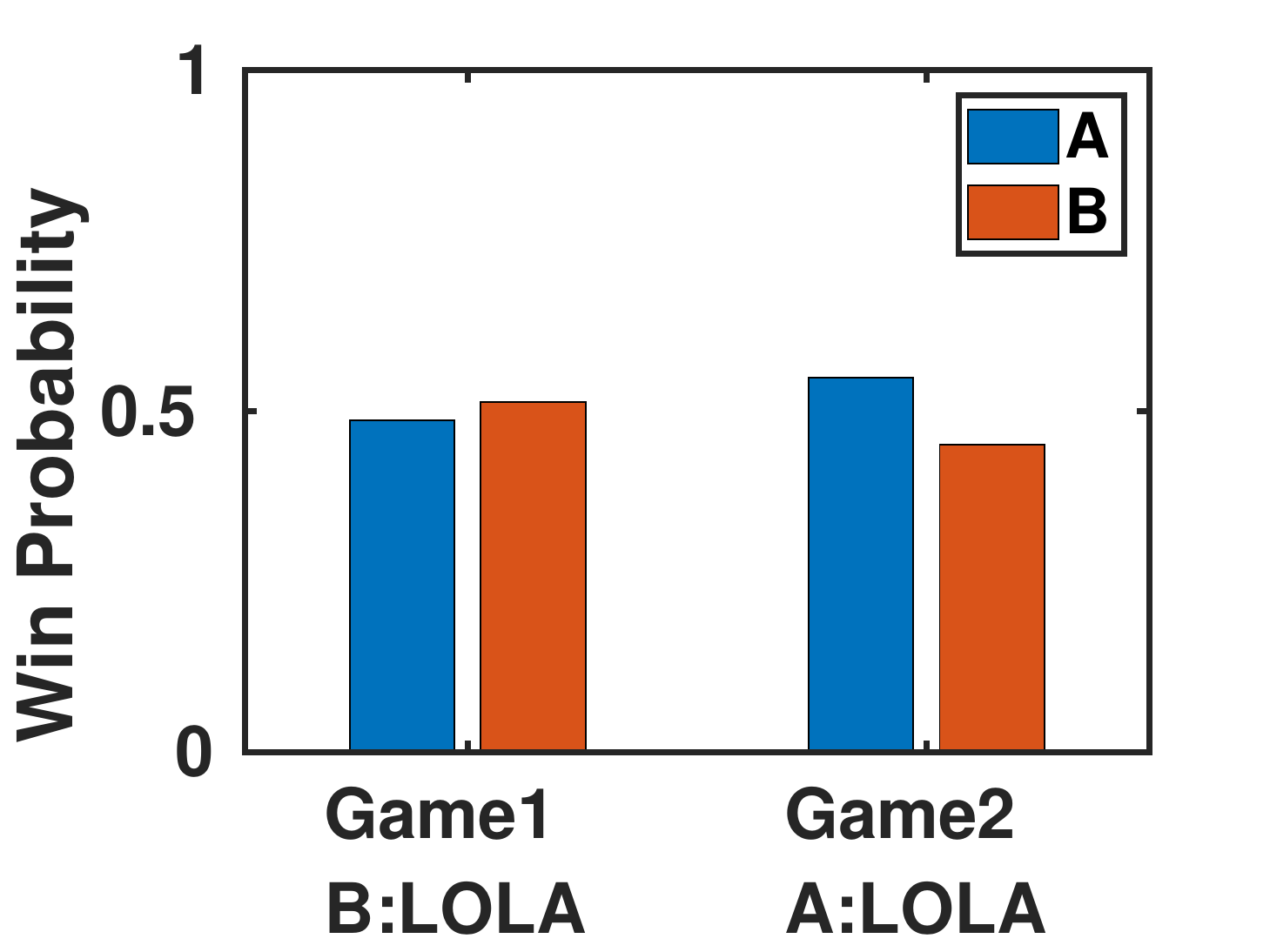}
    \caption{\lola vs \gda}
    \label{fig: EX_sgd_lola_winning_plot}
  \end{subfigure}
   \hspace{-6.0mm}
 ~
  \begin{subfigure}[t]{0.26\columnwidth}
  	\centering
  	\includegraphics[width=\linewidth]{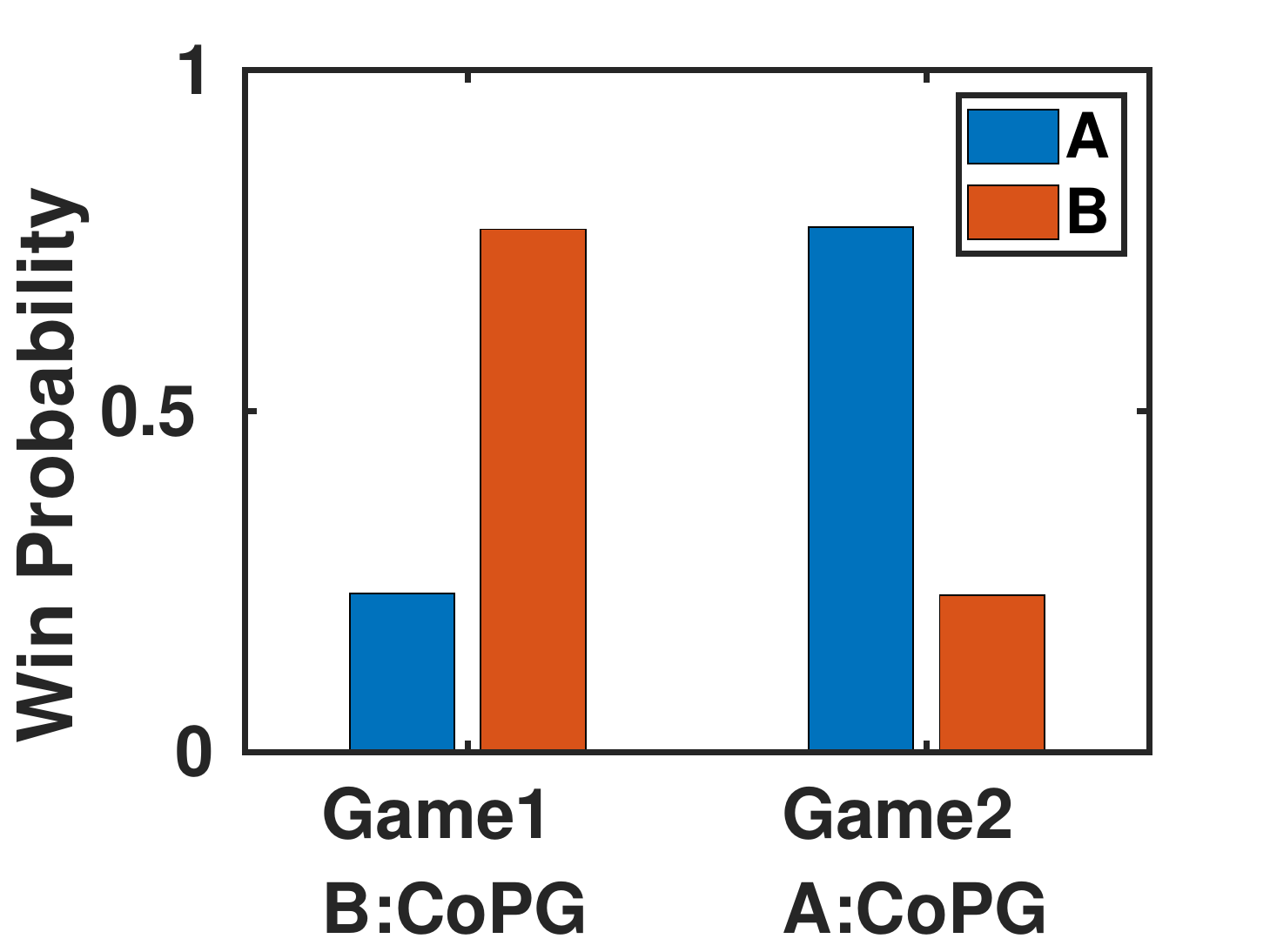}
    \caption{\cpg vs \lola}
    \label{fig: EX_lola_cpg_winning_plot}
  \end{subfigure}
   \hspace{-5.00mm}
}
    \caption{Comparison with \lola and \maddpg on the Markov soccer game}
    \label{fig: maddpg_lola_soccer}
\end{figure}

Finally, we give more details on our comparison of \cpg with \maddpg and \lola. We performed this comparison on the Markov soccer and the racing game. We start with the comparison for the Markov soccer game. To compare the performance of \maddpg on discrete action spaces (such as in the Markov soccer game), we used policies that can produce differentiable samples through a Gumbel-Softmax distribution, similar to ~\citep{iqbal19a}. The rest of the experimental setup was kept as similar as explained in \ref{appxsec: soccer}, and we trained both \maddpg and \cpg. \fig~\ref{fig: EX_sgd_maddpg_winning_plot}  shows that a policy learned by \maddpg performs similar to that of \gda. But when the \maddpg player competes against \cpg, the \cpg agent wins 80\% of the games. Second we studied \lola, we observed that \lola learns better policies than \gda potentially due to the second level reasoning, but \lola still loses 72\% of the games when playing against \cpg. \fig~\ref{fig: maddpg_lola_soccer} show the number of games won by \cpg when competing against \maddpg and \lola.

Next we evaluated \maddpg and \lola in the game of Car Racing, where we kept all the setting identical to \ref{appx: ORCADescription}. \fig~\ref{fig: progress_plot_orca_maddpg} show the progress of an \maddpg agent during learning, and we can see that it performs similar to \gda. The car learns to drive up to 30\% of the track, but after that the agents show oscillatory behaviour in training. We also performed this experiment with \lola, but due to the variance in estimating the bilinear term and more importantly the non adaptive learning rate of \lola it does not learn to drive around the track. 

\end{document}